\documentclass[11pt]{article}

\usepackage[final]{acl}
\usepackage{multirow}
\usepackage{float}
\usepackage{times}
\usepackage{latexsym}
\usepackage{amsmath}
\usepackage[most]{tcolorbox}
\usepackage[T1]{fontenc}
\usepackage[utf8]{inputenc}
\usepackage{amssymb}

\usepackage{microtype}

\usepackage{inconsolata}

\usepackage{graphicx}
\usepackage{subcaption} % 导言区加上
\usepackage{svg}      % 专门插入 SVG

\usepackage{array}
\usepackage{xcolor}
\usepackage{colortbl}
\usepackage{longtable}
\usepackage{booktabs}
\usepackage{arydshln}
\usepackage{hyperref}

\usepackage{appendix}
\newcommand{\ourMethod}{GlobalRAG}

\title{GlobalRAG: Enhancing Global Reasoning in Multi-hop Question Answering via Reinforcement Learning}

\author{
  \textbf{Jinchang Luo\textsuperscript{1}},
  \textbf{Mingquan Cheng\textsuperscript{1}},
  \textbf{Fan Wan\textsuperscript{1}},
  \textbf{Ni Li\textsuperscript{1}},
  \textbf{Xiaoling Xia\textsuperscript{1}},
   \\
  \textbf{Shuangshuang Tian\textsuperscript{2}},
  \textbf{Tingcheng Bian\textsuperscript{1,4}},
  \textbf{Haiwei Wang \textsuperscript{1}}\thanks{Corresponding authors. Email: \href{mailto:wanghaiwei@baidu.com}{wanghaiwei@baidu.com} and \href{mailto:haohuan@tsinghua.edu.cn}{haohuan@tsinghua.edu.cn}},
  \textbf{Haohuan Fu\textsuperscript{2,3}}\footnotemark[1],
  \textbf{Yan Tao\textsuperscript{1}}
\\
\\
  \textsuperscript{1}Baidu Inc.,
  \textsuperscript{2}Tsinghua Shenzhen International Graduate School, Tsinghua University
   \\
  \textsuperscript{3}National Supercomputing Center in Shenzhen \\
  \textsuperscript{4}Shenzhen University
  \\
}

\begin{document}
\maketitle

\begin{abstract}
Reinforcement learning has recently shown promise in improving retrieval-augmented generation (RAG). Despite these advances, its effectiveness in multi-hop question answering (QA) remains constrained by two fundamental issues: (i) global planning absence to structure multi-step reasoning, and (ii) unfaithful execution, which hinders effective query formulation and consistent use of retrieved evidence. We propose \textbf{\ourMethod{}}, a reinforcement learning framework designed to enhance global reasoning in multi-hop QA. \ourMethod{} decomposes questions into subgoals, coordinates retrieval with reasoning, and refines evidence iteratively. To guide this process, we introduce \textit{Planning Quality Reward} and \textit{SubGoal Completion Reward}, which encourage coherent planning and reliable subgoal execution. In addition, a progressive weight annealing strategy balances process-oriented and outcome-based objectives. Extensive experiments on both in-domain and out-of-domain benchmarks demonstrate that \ourMethod{} significantly outperforms strong baselines while using only 8k training data (42\% of the training data used by strong baselines), achieving average improvements of 14.2\% in both EM and F1. 
% The code and model checkpoints are available at \href{https://github.com/CarnegieBin/GlobalRAG}{https://github.com/CarnegieBin/GlobalRAG}.
\end{abstract}

% 放在 \maketitle 之后（或正文第一页的最前面）

\section{Introduction}
{

\begin{figure}[t]
  \centering
  % 上面大图
  \includegraphics[width=\columnwidth]{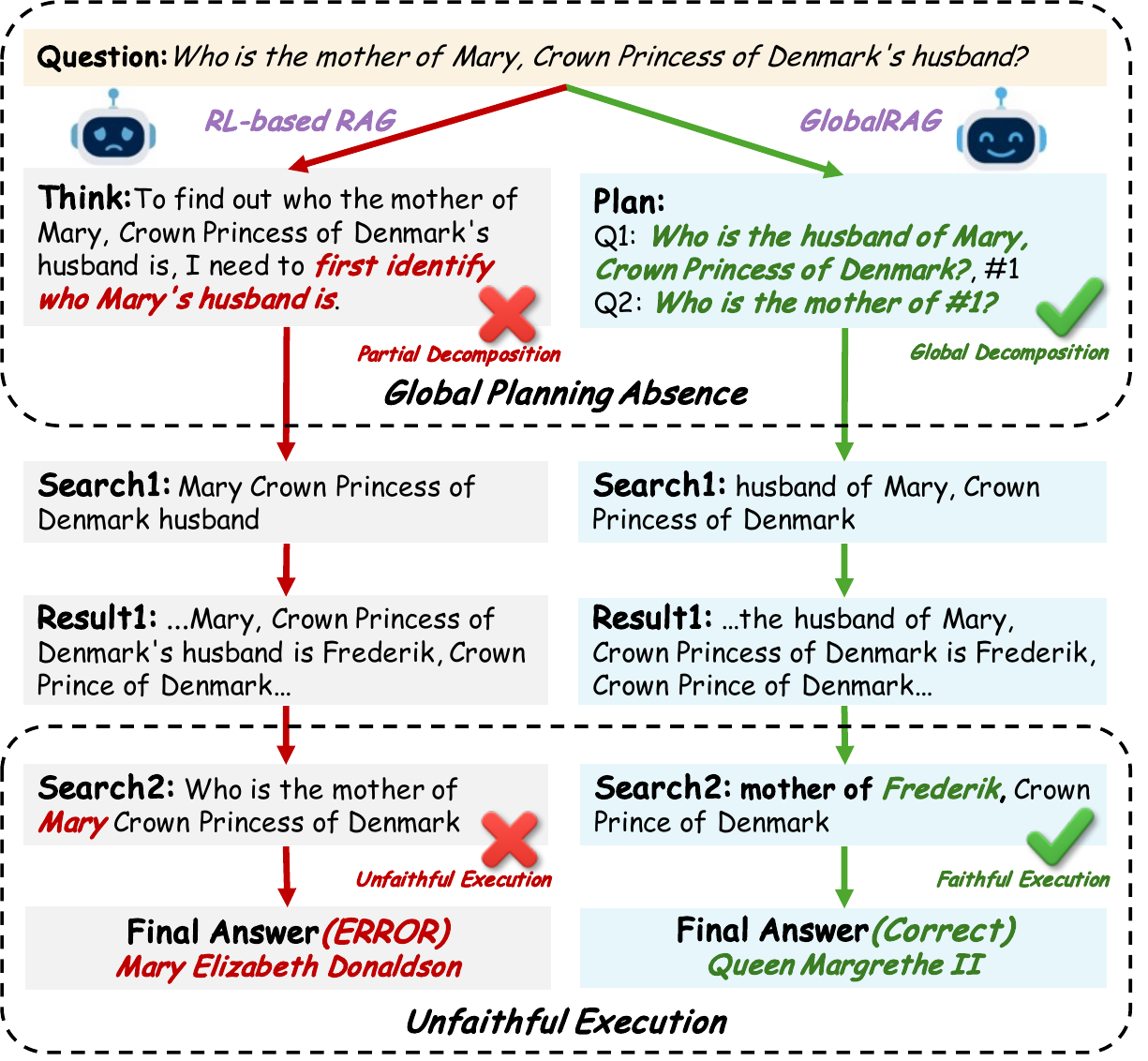}
  %\caption*{(a) Problem case description} % 给大图标题，不影响整体caption
  \\[3pt]

  % 下面两张子图并列
  \begin{subfigure}{0.53\linewidth}
      \centering
      \includegraphics[width=\linewidth]{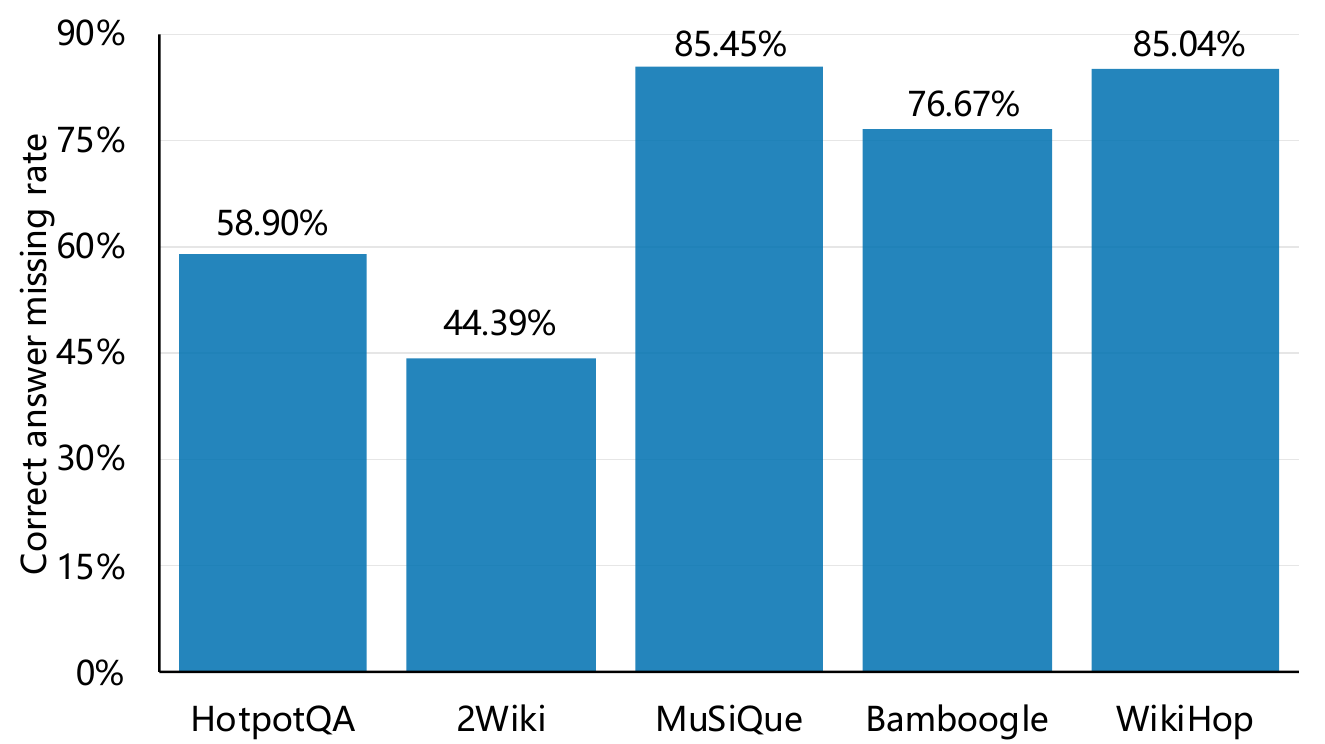}
   %   \caption{Correct Answer Missing Ratio}
  \end{subfigure}
  \hfill
  \begin{subfigure}{0.45\linewidth}
      \centering
      \includegraphics[width=\linewidth]{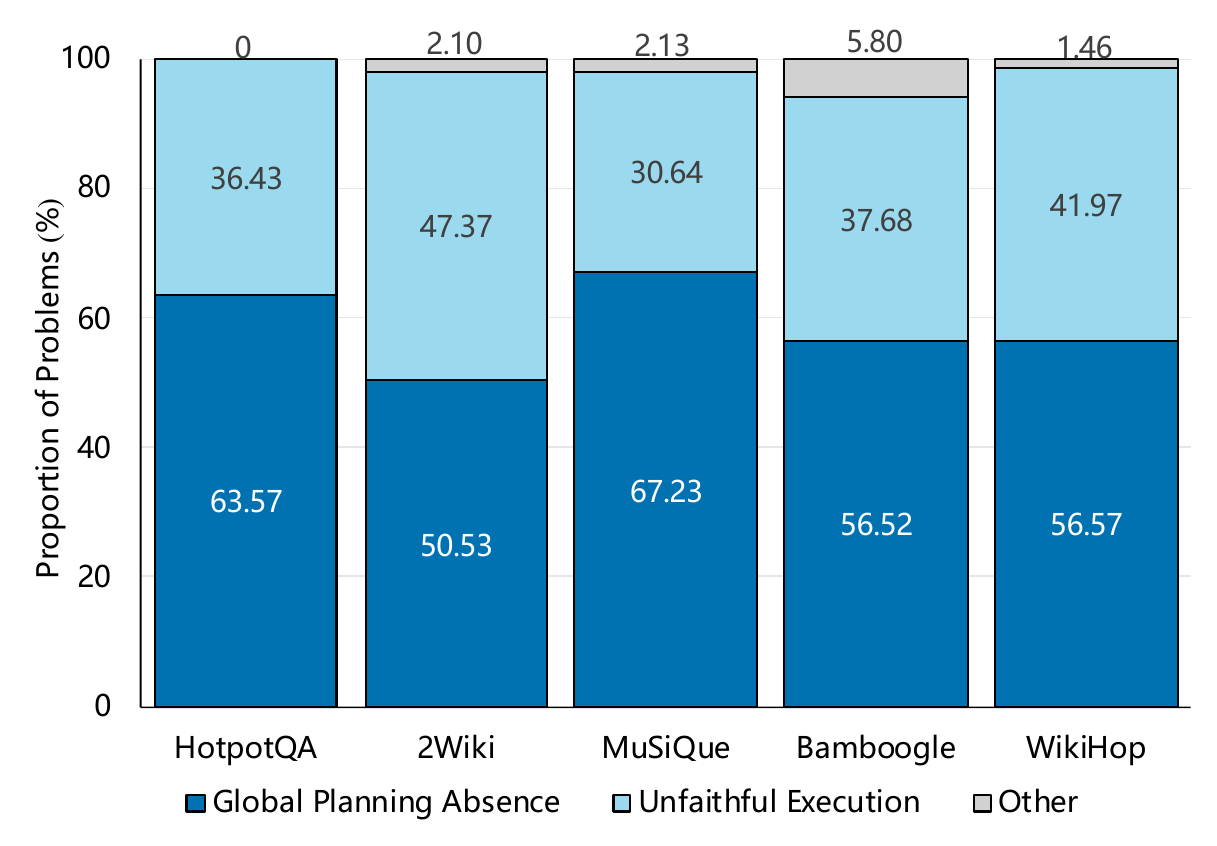}
  %    \caption{Error Type of Missing Answers}
  \end{subfigure}

  \caption{(\textit{top}) Example of global planning absence and unfaithful execution. 
(\textit{bottom left}) The most frequent failure of Search-R1 is missing the correct document.
(\textit{bottom right}) Error distribution where these two issues account for over 94\% of failures. }
  \label{fig:example}
\end{figure}

Reinforcement learning with verifiable rewards (RLVR) has recently demonstrated strong empirical gains \cite{guo-2025-deepseek-r1-nature,dong2025agentic,lai2025computerrl}, particularly in retrieval-augmented generation (RAG), where RL-based methods achieve competitive performance by explicitly optimizing reasoning and retrieval behaviors \cite{jin-2025-searchr1,chen2025researchlearningreasonsearch,DBLP:journals/corr/abs-2503-05592}. 
Despite this progress, existing approaches still struggle with multi-hop and compositional questions, where reasoning requires long-horizon planning and consistent execution across multiple retrieval steps.

Recent studies attempt to mitigate this limitation by encouraging more frequent or structured retrieval, such as \textit{StepSearch}~\cite{anonymous2025stepsearch}, \textit{LeTS}~\cite{zhang2025lets}, \textit{ParallelSearch}~\cite{zhao2025parallelsearch}, and \textit{GraphRAG-R1}~\cite{DBLP:journals/corr/abs-2507-23581}. While these methods alleviate shallow retrieval to some extent, they do not fundamentally address the dominant failure modes in multi-hop QA.

Through an empirical analysis on five representative multi-hop benchmarks—HotpotQA, 2WikiMultiHopQA, MuSiQue, Bamboogle, and WikiHop—we find that the majority of errors arise from the model’s inability to retrieve documents containing the correct answer after multiple retrieval steps. A finer-grained inspection of reasoning trajectories further reveals two dominant causes: \textbf{global planning absence}, where the model fails to construct a coherent decomposition of the question, and \textbf{unfaithful execution}, where retrieval and reasoning drift away from the intended subgoals. Together, these two issues account for over 94\% of failures in cases where the correct answer is missing (details in Appendix~\ref{appendix:B}).

Motivated by these observations, we propose \textbf{\ourMethod{}}, a reinforcement learning framework designed to enhance global reasoning in multi-hop QA. To mitigate global planning absence, \ourMethod{} introduces planning-aware optimization that explicitly decomposes questions into structured subgoals, guided by a \textit{Planning Quality Reward} measuring both structural and semantic consistency. To address unfaithful execution, we further design a \textit{SubGoal Completion Reward} that encourages faithful alignment between planned subgoals and executed retrieval–reasoning steps. These objectives are jointly optimized using GRPO~\cite{shao2024deepseekmath}, enabling stable and coherent long-horizon reasoning. Additionally, inspired by TIRESRAG-R1~\cite{he-2025-tiresragr1}, we adopt a progressive weight annealing strategy to balance process-level supervision and final-answer optimization during training.

Experiments across diverse in-domain and out-of-domain benchmarks demonstrate that \ourMethod{} consistently outperforms strong baselines, yielding substantial gains in multi-hop reasoning accuracy.

% Our contributions are summarized as follows:
% \begin{itemize}
%     \item We identify global planning absence and unfaithful execution as the two primary failure modes of RL-based multi-hop QA.
%     \item We propose \ourMethod{}, a planning-aware reinforcement learning framework that strengthens global reasoning through graph-guided dense supervision.
%     \item Extensive experiments show that \ourMethod{} achieves consistent improvements over prior methods, with average gains of 14.2\% in both EM and F1.
% \end{itemize}

Our contributions can be summarized as follows:
\begin{itemize}
  \item We identify two fundamental limitations of existing RL-based approaches for multi-hop QA: global planning absence and unfaithful execution.
    \item We propose \ourMethod{}, a reinforcement learning framework that introduces planning-aware optimization to strengthen global reasoning in multi-hop QA.
    \item Extensive experiments on both in-domain and out-of-domain datasets demonstrate that \ourMethod{} consistently outperforms strong baselines, yielding average gains of 14.2\% in both EM and F1.
\end{itemize}

}

\section{Related Work}

\noindent{\textbf{Retrieval-Augmented Generation for Multi-hop QA.}} 
Multi-hop QA requires multi-step reasoning over evidence from multiple sources~\cite{li-du-2023-leveraging, zhang-etal-2024-end, cheng-2025-dualrag}. Retrieval-augmented generation (RAG) addresses this challenge by coupling retrieval with generation~\cite{survey_of_natural_language_generation, 10448015, shi-etal-2024-generate}. Existing RAG methods for multi-hop QA mainly fall into two categories. 
Graph-structured approaches~\cite{HippoRAG-2024, hippo2-gutierrez2025, fang-etal-2025-kirag} organize textual knowledge into entity–relation graphs and aggregate evidence along multi-hop paths, but require explicit graph construction and maintenance~\cite{HippoRAG-2024, hippo2-gutierrez2025}. 
Iterative RAG approaches~\cite{trivedi-etal-2023-interleaving, shao-etal-2023-enhancing, liu-etal-2024-ra, retrieve_summarize_plan, li2025search} perform retrieval and reasoning in multiple steps (e.g., IRCOT~\cite{trivedi-etal-2023-interleaving}, Search-o1~\cite{li2025search}), yet heavily rely on LLM-generated intermediate thoughts and are thus prone to hallucination~\cite{liu-etal-2024-ra, retrieve_summarize_plan}. 
In contrast, \ourMethod{} leverages reinforcement learning to enhance global planning and coordinated reasoning, enabling more accurate evidence localization and integration for multi-hop QA.

\noindent{\textbf{Large Language Models and Reinforcement Learning.}} 
Reinforcement learning has substantially improved complex reasoning and self-verification in LLM post-training~\cite{openai2024openaio1card, guo-2025-deepseek-r1-nature, dong2025agentic, lai2025computerrl}, and has recently been extended to RAG~\cite{jin-2025-searchr1, chen2025researchlearningreasonsearch, DBLP:journals/corr/abs-2503-05592, anonymous2025stepsearch}. 
Search-R1~\cite{jin-2025-searchr1} adopts outcome-based rewards to encourage retriever usage, while StepSearch~\cite{anonymous2025stepsearch} scores per-step retrieval quality via information gain and redundancy penalties. However, lacking explicit global planning and suffering from unfaithful execution, these methods remain suboptimal for multi-hop QA~\cite{jin-2025-searchr1, anonymous2025stepsearch}. 
By contrast, we introduce a Planning Quality Reward and a Subgoal Completion Reward to explicitly reinforce global reasoning and structured progress in multi-hop QA.

\lstset{
  basicstyle=\ttfamily\footnotesize,
  columns=fullflexible,     % 允许等宽字母压缩以便换行
  breaklines=true,          % 自动换行
  breakatwhitespace=false,  % 逗号/冒号等也可断
  keepspaces=true,
}
\section{Method}
\begin{figure*}[htbp] 
    \centering
    \includegraphics[width=0.9\textwidth]{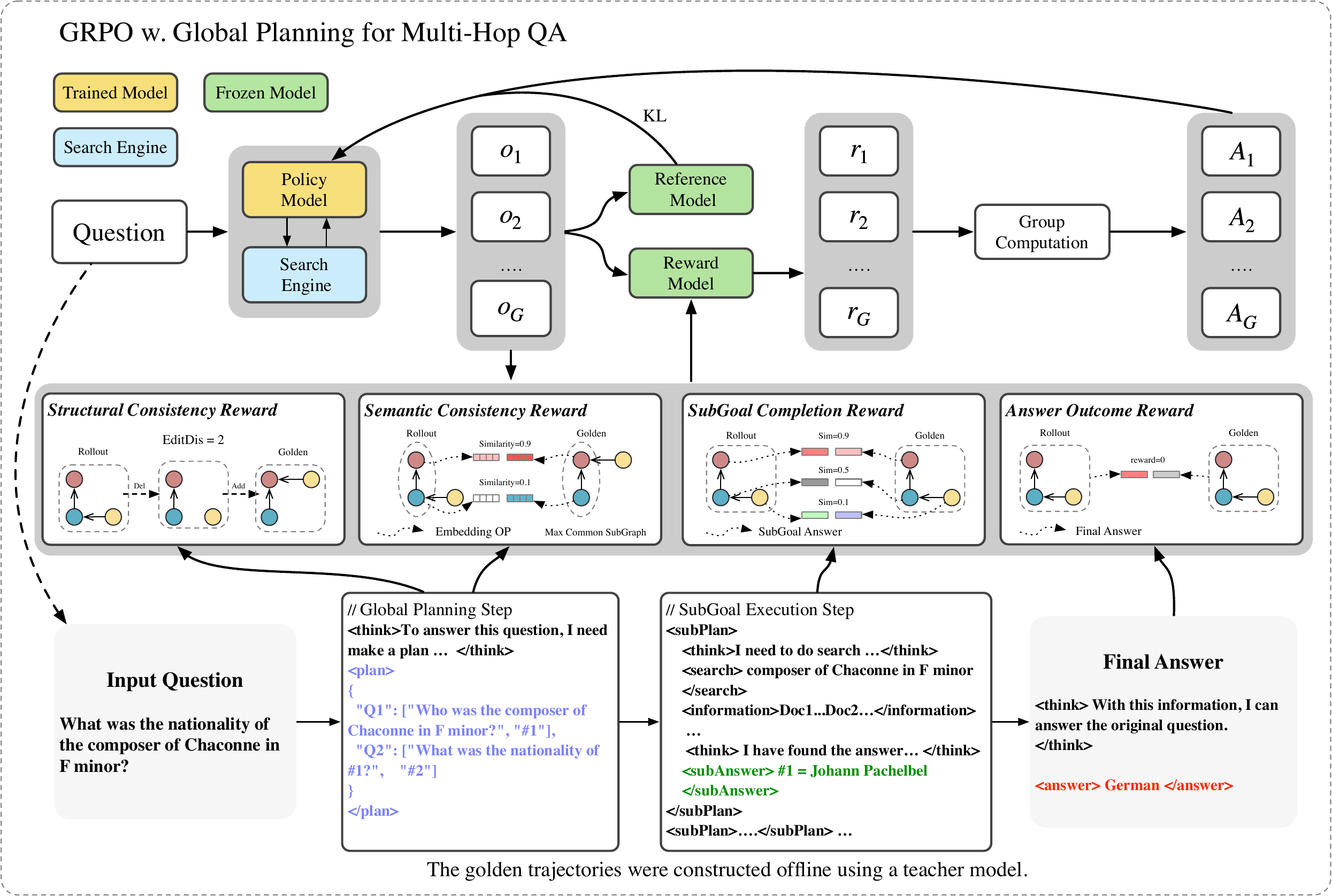} 
    % \includesvg[width=0.95\textwidth]{resource/figures/pipeline/Canvas 1}
    \caption{Overview of \ourMethod{}. The top illustrates rollout retrieval–enhanced GRPO with a search engine; the bottom shows golden trajectories constructed by a teacher model; and the middle depicts the global planning–aware reward design used to update the policy model.}
    \label{fig:pipeline}
\end{figure*}

{
In this section, we present \ourMethod{}, a framework for multi-hop reasoning. First, we introduce the rollout retrieval–enhanced GRPO algorithm. Second, we describe the generation of golden trajectories for reward supervision. Third, we detail the design of global planning–aware rewards. Finally, we present the progressive weight annealing strategy that integrates all rewards. An overview of the framework is illustrated in Figure~\ref{fig:pipeline}, while detailed formulations and implementation specifics of the GRPO training procedure are deferred to Appendix~\ref{appendix:C}.

% \subsection{Rollout Retrieval Enhanced GRPO for Multi-hop QA}
% {
% % An overview of the GRPO training scheme is shown in Figure~\ref{fig:pipeline}(top). Formally, given an actor model $\pi_\theta$ that interacts with the retrieval engine to generate $n$ rollouts for multi-hop QA, GRPO computes rewards from these rollouts and updates the actor model $\pi_\theta$ with reference to a frozen model $\pi_{\text{ref}}$. More Details can be found in \ref{appendix:C}

% An overview of the GRPO training scheme is shown in Figure~\ref{fig:pipeline}(top). The actor model $\pi_\theta$ interacts with the retrieval engine to generate $n$ rollouts for multi-hop QA, from which GRPO rewards are computed to update $\pi_\theta$ against a frozen reference model $\pi_{\text{ref}}$. Detailed formulations and implementation specifics are provided in Appendix~\ref{appendix:C}.

% }

% 3.1 
\subsection{Golden Trajectory Generation with Global Planning}
{

In this section, we describe the offline generation of golden trajectories, as illustrated in Figure~\ref{fig:pipeline}(bottom). The generation process proceeds through three stages: \textit{global planning}, \textit{subgoal execution}, and \textit{final answer generation}. For a detailed description of the trajectory generation procedure, including implementation specifics, see Appendix~\ref{appendix:D}.

% First, a teacher model performs global planning by analyzing the problem, decomposing it into sub-problems, and constructing a task-dependency graph with placeholder identifiers that define the execution paths.  
% Second, the model solves sub-problems iteratively in topological order, conducting reasoning, retrieval, document acquisition, and sub-answer generation, while propagating intermediate results through placeholder substitution.  
% Finally, the model produces the final answer by integrating the sub-answers according to the global plan and reasoning process.
}
\subsection{Global Planning--Aware Reward Design}
% 结合GPT重新描述，增加防御点：放置审稿人认为我们的方法是单一分解路径依赖，缺乏分解多样性

While reinforcement learning has been shown effective for RAG, existing approaches that rely primarily on simple rule-based rewards (such as final-answer or format rewards) tend to encourage shallow retrieval and short-horizon optimization, particularly in multi-hop QA settings where explicit global planning signals are absent. This often leads the model to deviate from the original reasoning intent during intermediate steps. To mitigate this issue, we propose \emph{Global Planning--Aware Rewards}, which provide process-level supervision to better align long-horizon reasoning with the overall task objective. Rather than enforcing strict imitation of a single reference plan, our design adopts a soft, similarity-based alignment mechanism that assigns partial credit to diverse yet valid planning structures, \textbf{which we empirically observe to vary in subgoal ordering and granularity while remaining task-consistent}. This formulation promotes flexible but faithful reasoning and reduces overfitting to specific planning patterns. Concretely, our approach integrates three complementary components: \textit{Structural Consistency}, \textit{Semantic Consistency}, and \textit{Subgoal Completion} rewards.

\subsubsection{Planning Quality Reward}

Our reward design evaluates the consistency between the rollout plan and the golden plan, both represented as directed acyclic graphs (DAGs) where nodes are subgoals and edges denote dependencies (Figure~\ref{fig:pipeline}). We introduce two rewards: \textit{Structural Consistency Reward}, measuring graph-level alignment, and \textit{Semantic Consistency Reward}, assessing subgoal-level similarity. These rewards encourage more faithful and coherent reasoning.

\textbf{Structural Consistency Reward} 
enforces global planning awareness by evaluating the full dependency structure of complex tasks. In contrast to step-wise RL methods that optimize only the current step \cite{song-2025-r1searcherpp,he-2025-tiresragr1,shi-2025-autorefine}, we assess the entire graph topology to capture long-range dependencies and inter-task relationships. To quantify structural differences, we employ graph edit distance—the minimum number of operations required to transform $G$ into $\tilde{G}$—and map it into a similarity score in $[0,1]$ using an exponential function, with perfect alignment yielding the maximum reward.  

Formally, given the rollout plan graph $G$ and golden plan graph $\tilde{G}$, the structural consistency reward is defined as:
\begin{equation}
R_{\text{str}} = \exp\big(-\frac{d_{\text{edit}}(G, \tilde{G})}{N_{\tilde{G}} + E_{\tilde{G}}}\big),
\end{equation}
where $d_{\text{edit}}(G, \tilde{G})$ denotes the graph edit distance. $N_{\tilde{G}}$ and $E_{\tilde{G}}$ denote the node and edge counts of $\tilde{G}$, respectively.

\textbf{Semantic Consistency Reward} evaluates the semantic alignment between the predicted plan and the golden plan. First, we employ a depth-first search (DFS) algorithm to identify the maximum common subgraph between the generated plan and the reference plan, thereby establishing valid and reliable structural correspondences. Second, we use the E5-base~\cite{wang-2022-e5} model to embed all subproblem nodes into vectors, and compute semantic similarity only for matched subproblems within the maximum common subgraph. This design avoids spurious comparisons between structurally incompatible components and ensures evaluation focuses on semantically meaningful correspondences.  

Formally, let $\mathcal{P}_{\text{gold}}$ denote the reference subproblems and $\mathcal{M}_P$ the set of matched subproblem pairs within the maximum common subgraph. For each matched pair $i$, $\rho_{\text{pred}}^{(i)}$ and $\rho_{\text{gold}}^{(i)}$ denote the embeddings of predicted and reference subproblems, respectively. The semantic consistency reward is defined as:
\begin{equation}
R_{\text{sem}} = \frac{1}{|\mathcal{P}_{\text{gold}}|} \sum_{i=1}^{|\mathcal{M}_P|} 
\cos\!\left(\rho_{\text{pred}}^{(i)}, \rho_{\text{gold}}^{(i)}\right),
\end{equation}
where $\cos(\cdot,\cdot)$ denotes the cosine similarity.

\subsubsection{Subgoal Completion Reward}
{
This reward addresses the problem of unfaithful execution by encouraging the model to follow the global plan and produce reliable outputs for each subgoal. Unlike StepSearch~\cite{anonymous2025stepsearch}, which applies rewards directly to the token sequence of generated search queries to stabilize training and increase signal density, we argue that this strategy overly constrains exploration and discourages the discovery of novel reasoning paths. Instead, our reward design focuses solely on the correctness of subgoal answers, without penalizing the search process, thereby leaving room for more exploratory and heuristic solution strategies.  

Formally, for each matched subgoal $i \in \mathcal{M}_P$, let $o_{\text{pred}}^{(i)}$ and $o_{\text{gold}}^{(i)}$ denote predicted and gold answers, and $\phi(\cdot,\cdot)$ a similarity function (e.g., the F1 score between the predicted and gold answers). The reward is normalized by $|\mathcal{P}_{\text{gold}}|$ to ensure coverage, with unmatched nodes contributing zero:
\begin{equation}
R_{\text{step}} = \frac{1}{|\mathcal{P}_{\text{gold}}|} 
\sum_{i=1}^{|\mathcal{M}_P|} \phi\!\left(o_{\text{pred}}^{(i)}, o_{\text{gold}}^{(i)}\right).
\end{equation}

}

\subsubsection{Outcome Reward}
{

A global plan and process rewards guide how the model reasons, but training must be closed by a terminal constraint requiring the correct final answer. The outcome reward enforces this by validating that execution yields the target solution, preventing credit on well-formed but unfaithful traces, and aligning optimization with end accuracy. We adopt a strict exact-match (EM) signal, applied after standard normalization, as a binary reward:
\begin{equation}
R_{\text{answ}} =
\begin{cases}
1, & \text{if the answer exactly matches,} \\
0, & \text{otherwise.}
\end{cases}
\end{equation}

}

\subsubsection{Format Reward}
{
We treat the format reward as a front-door constraint that enforces a machine-parseable reasoning path aligned with our planning-centric RL pipeline, enabling the policy to learn \emph{how} to plan, \emph{what} to retrieve, and \emph{where} to decide. This parseable protocol supports plan rewards, step supervision, stable policy-gradient training, and scalable evaluation.  

The format reward is binary:
\begin{equation}
R_{\mathrm{form}}=
\begin{cases}
1,& \text{if the format is fully compliant},\\
0,& \text{otherwise}.
\end{cases}
\end{equation}

}

\subsubsection{Progressive weight annealing strategy }
{
{
\newcommand{\best}[1]{\textbf{#1}}

\begin{table*}[!t]
\centering
\footnotesize
\setlength{\tabcolsep}{4pt}
\renewcommand{\arraystretch}{1.12}

\resizebox{\textwidth}{!}{
\begin{tabular}{l *{12}{c}}
\toprule
\multirow{2}{*}{Methods} &
\multicolumn{6}{c}{\textbf{In-Domain QA}} &
\multicolumn{4}{c}{\textbf{Out-of-Domain QA}} &
\multicolumn{2}{c}{\textbf{Average}} \\
\cmidrule(lr){2-7}\cmidrule(lr){8-11}\cmidrule(lr){12-13}
& \multicolumn{2}{c}{\textbf{HotpotQA}} 
& \multicolumn{2}{c}{\textbf{2Wiki}}
& \multicolumn{2}{c}{\textbf{MuSiQue}}
& \multicolumn{2}{c}{\textbf{Bamboogle}}
& \multicolumn{2}{c}{\textbf{WikiHop}}
&  \\
& EM & F1 & EM & F1 & EM & F1 & EM & F1 & EM & F1 & EM & F1 \\
\midrule
\textbf{Qwen2.5-3b-Base/Instruct} \\

Zero-Shot
& 1.2 & 3.3 & 1.6 & 2.7 & 0.0 & 1.2 & 0.0 & 1.9 & 0.1 & 0.7 & 0.6 & 2.0 \\
NaiveRAG
& 6.5& 13.8 & 2.2 & 4.1 & 0.5 & 1.9 & 4.0 & 8.5 & 2.8 & 7.6 & 3.2 & 7.2 \\
%\midrule
IRCOT
& 4.8 & 11.2 & 3.0 & 8.6 & 0.9 & 3.2 & 5.6 & 10.1 & 2.8 & 6.3 & 3.4 & 7.9 \\
Search-o1
& 14.6 & 22.1 & 14.0 & 17.0 & 3.6 & 7.7 & 26.4 & 34.9 & 5.3 & 10.3 & 12.8 & 18.4 \\
%\midrule
Search-R1-base
& 26.8 & 36.7 & 25.6 & 29.2 & 2.5 & 7.3 & 11.2 & 20.4 & 9.3 & 16.5 & 15.1 & 22.0 \\
Search-R1-instruct
& {31.3} & {41.5} & 31.3 & 36.5 & 7.7 & 13.2 & 28.0 & 34.7 & 8.2 & 13.8 & 21.3 & 27.9 \\
StepSearch-base
&29.1  &40.3  & 33.5 & 38.7 & 10.4 & {17.7} & 32.8  & 41.8 &\best{13.5} &19.6  &23.9  & 31.6 \\
StepSearch-instruct
& 28.9 & 39.9 & 31.9 & 38.3 & 9.5 & 16.6 & 32.0 & 43.8 & 11.2 & 18.1 & 22.7 & 31.3 \\
%\midrule
\hdashline
\ourMethod-base
& 30.4 & 41.0 & \best{43.1} & \best{48.3} & \best{11.3} & 17.4 & \best{40.0} & \best{49.5} & 12.7 & {19.7} & \best{27.5} & {35.2} \\
\ourMethod-instruct
& \best{32.9} & \best{44.2} & {42.3} & {47.8} & {10.8} & \best{18.6} & {37.6} & {49.3} & {12.9} & \best{20.7} & {27.3} & \best{36.1} \\

\midrule

\textbf{Qwen2.5-7b-Base/Instruct} \\
Search-R1-base
& 28.5 & 38.3 & 26.7 & 31.3 & 7.1 & 12.9 & 35.2 & 43.1 & 11.4 & 17.4 & 21.8 & 28.6 \\
Search-R1-instruct
&  35.9 & 47.5 & 33.4 & 39.5 & 9.7 & 16.7 & 43.4 & 52.7 & 15.5 & 23.0 & 27.6 & 35.9 \\
StepSearch-base
&  33.5 & 43.5 & 39.1 & 44.8 & 13.1 & 22.1 & 42.4 & 52.4 & 16.9 & 24.6 & 29.0 & 37.5 \\
StepSearch-instruct
&  34.8 & 46.9 & 36.3 & 42.5 & 13.4 & 21.4 & 41.6 & 53.1 & 14.9 & 22.3 & 28.2 & 37.2 \\
%\midrule
\hdashline
\ourMethod-base
& 33.0 & 44.4 & 47.0 & 52.4 & 14.2 & \best{22.7} & 46.4 & 51.0 & 16.0 & 22.7 & 31.3 & 38.6 \\
\ourMethod-instruct
& \best{38.2} & \best{50.7} & \best{47.8} & \best{53.2} & \best{14.9} & 21.2 & \best{48.8} & \best{61.6} & \best{19.0} & \best{28.0} & \best{33.7} & \best{42.9} \\

\bottomrule
\end{tabular}}
\caption{
Multi-hop QA results (first three in-domain; last two out-of-domain). Best are \best{bold}. Methods without “base/instruct” default to instruct.
}
\label{tab:main-results}
\end{table*}
}

Inspired by prior work, we adopt a progressive annealing scheme of \emph{structure first, accuracy later}. Early training emphasizes format, planning structure, semantics, and step execution, guiding the policy to construct and follow a coherent graph. Later training shifts weight toward the outcome signal to refine exactness without losing this discipline.  
The scheduling weight $w_t$ balances process and outcome rewards:
\begin{equation}
    w_t = \frac{1}{1+\exp\left(\frac{t-0.9T}{10}\right)},
\end{equation}
where $t$ and $T$ denote the current and total training steps.  

The overall reward integrates all components with dynamic weighting:
\begin{multline}
R(t) = w_t\big(\alpha R_{\text{form}} + \lambda R_{\text{str}} + \gamma R_{\text{sem}} \\
+ \delta R_{\text{step}}\big) + R_{\text{answ}},
\end{multline}
This curriculum smooths optimization, reduces variance, coordinates retrieval under the global plan, and transitions from structural learning to end-to-end reasoning accuracy.

}

\section{Experiments}

\subsection{Experimental Setup}
{
\noindent{\textbf{Datasets.}}
We evaluate on five multi-hop QA benchmarks: HotpotQA~\cite{yang-etal-2018-hotpotqa}, 2WikiMultiHopQA~\cite{ho-etal-2020-constructing-data}, MuSiQue~\cite{trivedi-etal-2022-musique}, Bamboogle ~\cite{press-etal-2023-measuring-data}, and WikiHop~\cite{welbl-etal-2018-constructing}, which require multi-step reasoning over evidence from multiple documents. Detailed statistics and corpus information are provided in Appendix~\ref{appendix:E.1}.

% We evaluate on five multi-hop QA benchmarks: HotpotQA~\cite{yang-etal-2018-hotpotqa}, 2WikiMultiHopQA~\cite{ho-etal-2020-constructing-data}, MuSiQue~\cite{trivedi-etal-2022-musique}, Bamboogle ~\cite{press-etal-2023-measuring-data}, and WikiHop~\cite{welbl-etal-2018-constructing}, which require multi-step reasoning over evidence from multiple documents. We report results on the full test sets for Bamboogle and WikiHop, and on sampled subsets for HotpotQA, 2WikiMultiHopQA, and MuSiQue. For datasets with nonpublic test sets (e.g., HotpotQA, 2WikiMultiHopQA, MuSiQue, WikiHop), we use their development sets as test sets and report the corresponding results.  Detailed statistics and corpus information are provided in Appendix~\ref{appendix:C.1}.
% 12 25

\noindent{\textbf{Baselines}}
To evaluate the effectiveness of \ourMethod{}, we compare against representative baselines spanning prompting- and reinforcement-learning paradigms as well as static and dynamic retrieval strategies, organized into three categories: (1) \textbf{Inference without Retrieval}: Zero-Shot, where the model answers solely from parametric knowledge without external context. (2) \textbf{Retrieval-Augmented Inference}: NaiveRAG~\cite{10.5555/3495724.3496517}, IRCOT~\cite{trivedi-etal-2023-interleaving}, and Search-o1~\cite{li2025search}. (3) \textbf{Retrieval with RL}: Search-R1~\cite{jin-2025-searchr1}, StepSearch~\cite{anonymous2025stepsearch}. We rely on authors’ official implementations and released checkpoints where available to ensure comparability. 

% Graph-RAG is not included as it optimizes the retriever, which is not directly comparable to our planning-based approach.

\noindent{\textbf{Evaluation Metrics.}}
We adopt Exact Match (EM) and word-level F1 as the evaluation metrics. Details are provided in Appendix~\ref{appendix:E.2}.

\noindent{\textbf{Training and Implementation Details.}}
We conduct experiments on Qwen-2.5(3B/7B) models (Base/Instruct) and use a dense retriever retrieving three passages for all retrieval-based methods. Training data are merged from HotpotQA, 2WikiMultiHopQA, and MuSiQue, generating 8,394 supervised trajectories for \ourMethod{}. See Appendix~\ref{appendix:E.3} for more details.
% We conduct experiments on two Qwen-2.5-3B (Base/Instruct)~\cite{qwen2025qwen25technicalreport}. For retrieval, we build a \texttt{faiss-gpu} index over the 2018 Wikipedia snapshot~\cite{karpukhin-etal-2020-dense} and use \textsc{E5} embeddings~\cite{wang-2022-e5} as the dense retriever. To ensure fair comparison, we follow Search-R1\citep{jin-2025-searchr1} and set the number of retrieved passages to 3 across all retrieval-based methods. 

% For training, we merge the training splits of HotpotQA, 2WikiMultiHopQA, and MuSiQue into a unified corpus and derive global-plan supervision for \ourMethod{} by inducing a subgoal–dependency plan and verifying intermediate steps, yielding 8,394 supervised trajectories. Our data generation is based on PER-PSE \cite{liu2025beyond}.
% 12 25

}

\subsection{Main Results}
          {
Table~\ref{tab:main-results} compares \ourMethod{} with competitive RL-trained and non-RL baselines across five datasets, with the following observations:

\noindent{\textbf{(1) Overall SOTA with strong gains:}} \ourMethod{} consistently outperforms all baselines, achieving the highest F1 on all datasets and four best/one second-best EM scores. Compared with StepSearch, average F1 improves by 14.2\% (HotpotQA +9.7\%, 2Wiki +23.5\%, MuSiQue +5.1\%, Bamboogle +17.9\%, WikiHop +5.6\%), demonstrating the effectiveness of our planning-aware RL framework in multi-hop QA.

\noindent{\textbf{(2) Gains correlate with dataset structure and multi-hop necessity:}}
Performance improvements vary with dataset design and multi-hop requirements. WikiHop and HotpotQA include many questions solvable with single-hop evidence, where extra retrieval can add noise~\cite{chen2019understanding}. In contrast, 2Wiki’s cross-document dependencies make iterative planning more effective. MuSiQue requires complex multi-hop reasoning; \ourMethod{} outperforms Search-R1, though gains over StepSearch are smaller since it already excels on its training distribution. Overall, the method shows the largest advantage when multi-hop reasoning and explicit cross-document aggregation are essential.

\noindent{\textbf{(3) Robust in-domain and out-of-domain performance:}} \ourMethod{} leads on both in-domain (HotpotQA, 2Wiki, MuSiQue) and out-of-domain datasets (Bamboogle, WikiHop), indicating strong generalization. By contrast, Search-R1 performs well in-domain but degrades notably out-of-domain, highlighting generalization challenges of existing methods.

% \noindent{\textbf{(4) Efficiency in Data, Training, and Inference-Time Trade-offs:}}Using only 8k examples (42\% of StepSearch's 19k; 4.7\% of Search-R1's 170k), \ourMethod{} achieves both 14.2\% EM/F1 gains over StepSearch and 28.2\% EM, 29.4\% F1 gains over Search-R1. This efficiency in data utilization and inference-time performance--cost trade-offs is mainly attributed to planning-aware supervision, including the Planning Quality and SubGoal Completion Rewards. Appendix~\ref{appendix:F} provides a detailed analysis of the performance--reasoning cost trade-off, showing that efficiency gains arise from informative rewards rather than reduced computational overhead.

\begin{figure}[!t]
    \centering
    \includegraphics[width=0.8\linewidth]{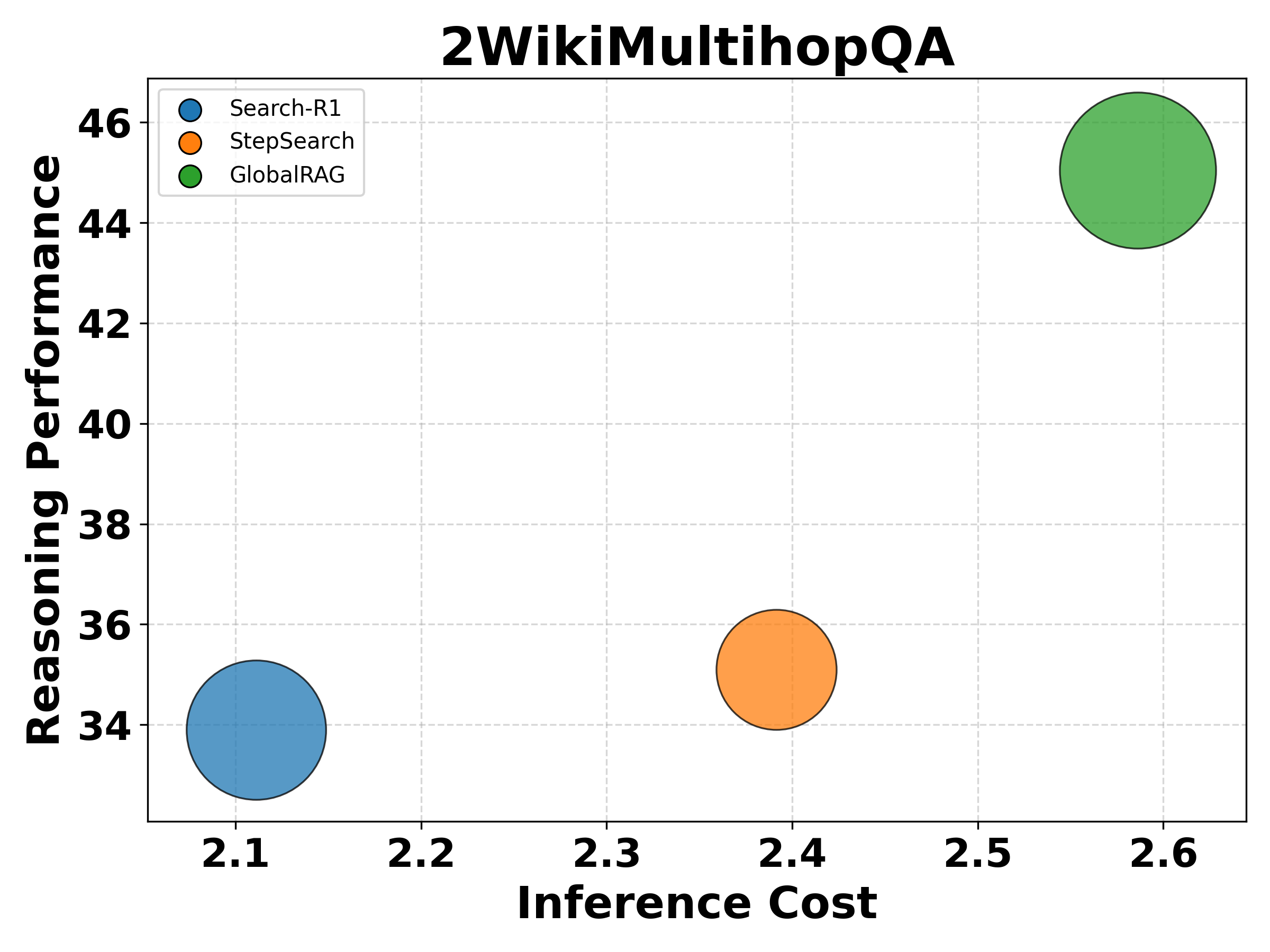}
    \caption{Inference-time performance versus token cost for three reinforcement learning based methods on \textbf{2WikiMultihopQA}.}
    \label{fig:2wikimultihopqa}
\end{figure}

\noindent{\textbf{(4) Efficiency in Data, Training, and Inference-Time Trade-offs:}} Using only 8k examples (42\% of StepSearch's 19k; 4.7\% of Search-R1's 170k), \ourMethod{} achieves both 14.2\% EM/F1 gains over StepSearch and 28.2\% EM, 29.4\% F1 gains over Search-R1. This efficiency in data utilization and inference-time performance--cost trade-offs is mainly attributed to planning-aware supervision, including the Planning Quality and SubGoal Completion Rewards. As illustrated in Figure~\ref{fig:2wikimultihopqa} (where bubble size indicates the efficiency ratio), \ourMethod{} establishes a superior balance between reasoning performance and token cost on 2WikiMultihopQA, with detailed analyses for the remaining datasets deferred to Appendix~\ref{appendix:F}.

% \noindent{\textbf{(5) Generalization across backbones and sizes:}} Across Qwen2.5 3B/7B Base and Instruct models, \ourMethod{} consistently tops all baselines, and performance margins are preserved or widened with larger models. This shows our method is backbone-agnostic and capacity-compatible, demonstrating strong plug-and-play generalization.
\noindent{\textbf{(5) Generalization across backbones and scales:}} 
Across Qwen2.5 3B/7B Base and Instruct models, \ourMethod{} consistently tops all baselines, and performance margins are preserved or widened with larger models. 
Moreover, as shown in Appendix~\ref{appendix:L}, consistent gains are also observed on the LLaMA-3.1-8B backbone, indicating that our method generalizes beyond the Qwen family to different model architectures and pretraining regimes.

}  

\subsection{Model Scale Study}
{
{\begin{figure}[!t] % 单栏
    \centering
    \begin{subfigure}[b]{0.48\linewidth} % 左上
        \centering
        \includegraphics[width=\linewidth]{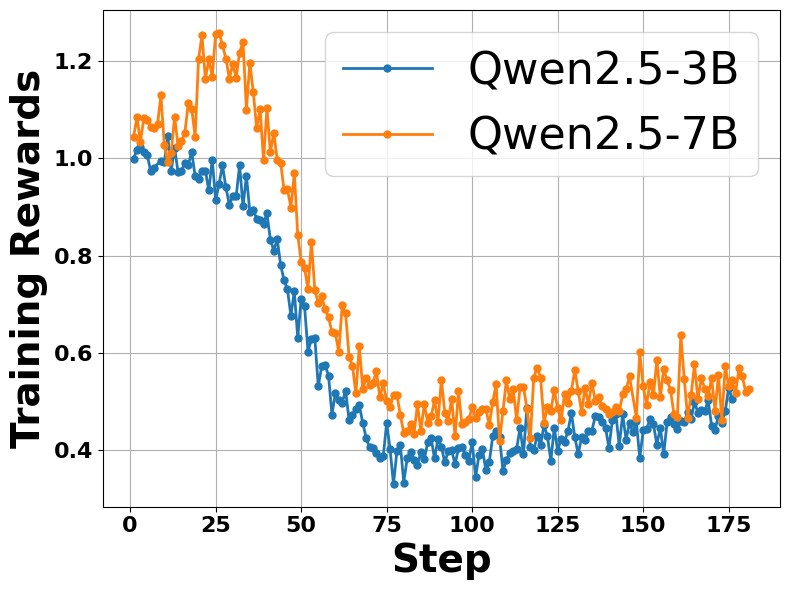}
        \caption{Rewards}
    \end{subfigure}
    \hfill
    \begin{subfigure}[b]{0.48\linewidth} % 右上
        \centering
        \includegraphics[width=\linewidth]{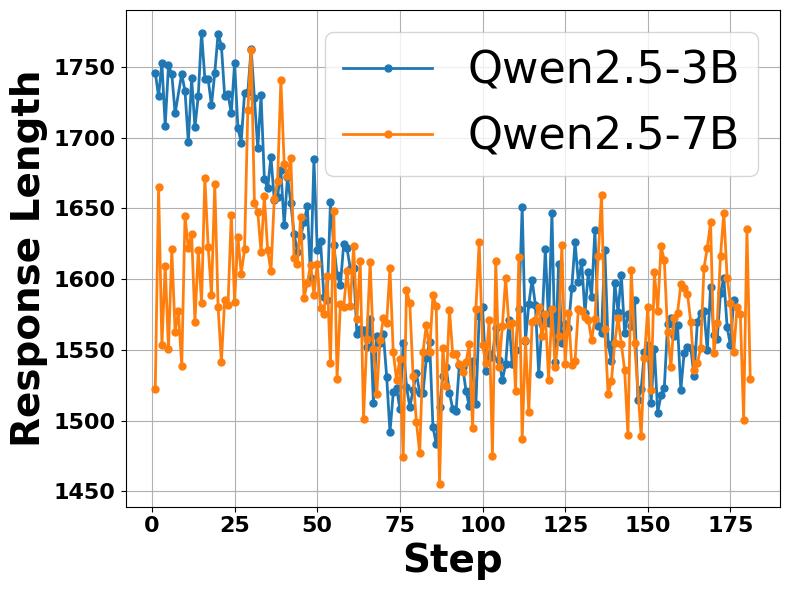}
        \caption{Response Length}
    \end{subfigure}

    \vskip 0.3cm % 上下图之间间距

    \begin{subfigure}[b]{0.48\linewidth} % 左下
        \centering
        \includegraphics[width=\linewidth]{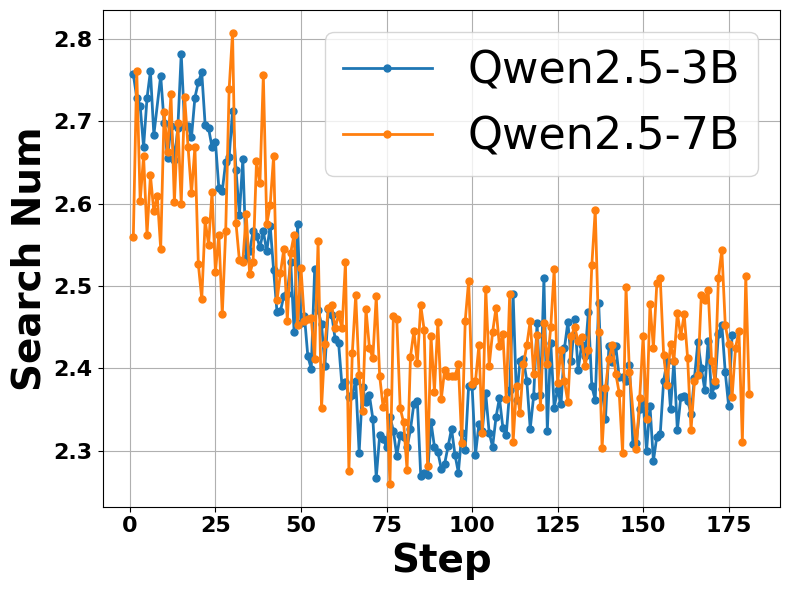}
        \caption{Search Num}
    \end{subfigure}
    \hfill
    \begin{subfigure}[b]{0.48\linewidth} % 右下
        \centering
        \includegraphics[width=\linewidth]{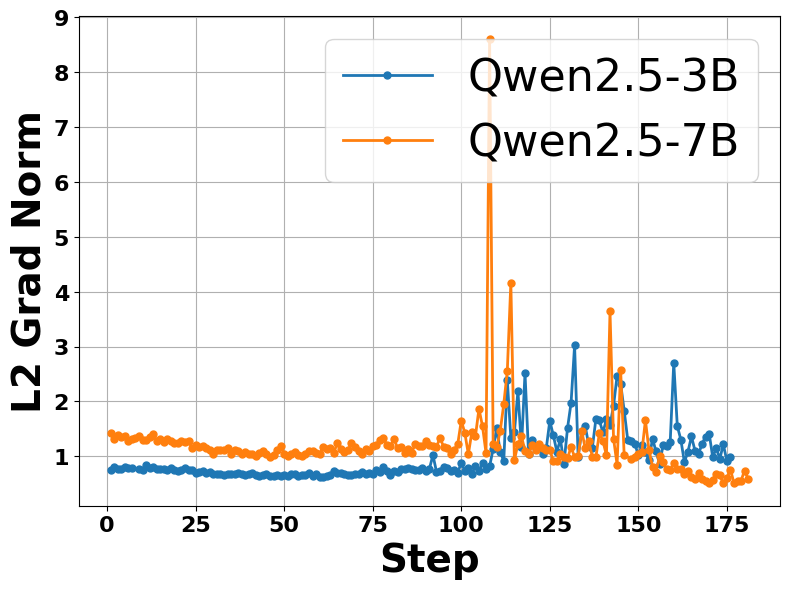}
        \caption{Grad Norm}
    \end{subfigure}

    \caption{Comparison of Qwen2.5-3B-Instruct and Qwen2.5-7B-Instruct models across four metrics. From left to right and top to bottom: (a) Training Rewards; (b) Response Length; (c) Search Num; (d) L2 Gradient Norm.}
    \label{fig:scale}
\end{figure}}
We use Qwen2.5 as the base model and conduct experiments at the 3B and 7B scales. As shown in Figure \ref{fig:scale}, the 7B model achieves higher reward values overall than the 3B model at the same training stage, and it also performs more searches with larger gradient norms. In the initial stage, the 7B model has shorter response lengths, but these gradually converge toward those of the 3B model during training. These trends indicate that, for larger-scale models, our method encourages the agent to retrieve information and optimize strategies more actively, while exhibiting more dynamic training behavior.

}

\subsection{Ablation Study}

% {
% \begin{table}[!t]
%   \centering
%   \resizebox{\columnwidth}{!}{%
%     \begin{tabular}{cccc}
%         \toprule
%         Methods & \textbf{Search-R1} & \textbf{StepSearch} & \textbf{\ourMethod{}} \\
%         \midrule
%         Train Size & 170k & 19k & 8k \\
%         \bottomrule
%     \end{tabular}
%     }
%   \caption{Comparison of training data sizes for different methods.}
%   \label{tab:train_data_num}
% \end{table}
% }

\begin{table}[h]
\centering
\footnotesize
\setlength{\tabcolsep}{3.8pt}
\renewcommand{\arraystretch}{1.12}
\label{tab:ablation}
\resizebox{\columnwidth}{!}{%
\begin{tabular}{lcccccccccc}
\toprule
\multirow{2}{*}{Methods} &
\multicolumn{2}{c}{\textbf{HotpotQA}} &
\multicolumn{2}{c}{\textbf{2Wiki}} &
\multicolumn{2}{c}{\textbf{MuSiQue}} &
\multicolumn{2}{c}{\textbf{Bamboogle}} &
\multicolumn{2}{c}{\textbf{WikiHop}} \\
\cmidrule(lr){2-3}\cmidrule(lr){4-5}\cmidrule(lr){6-7}\cmidrule(lr){8-9}\cmidrule(lr){10-11}
& EM & F1 & EM & F1 & EM & F1 & EM & F1 & EM & F1 \\
\midrule
w/o $R_{\mathrm{sem}}$  & 30.0 & 41.5 & 38.1 & 44.1 & 10.1 & 17.2 & 37.6 & 51.4 & 10.4 & 18.1 \\
w/o $R_{\mathrm{str}}$  & 30.6 & 42.2 & 39.6 & 46.0 &  9.5 & 18.3 & 39.2 & 47.9 & 11.2 & 18.7 \\
w/o $R_{\mathrm{step}}$ & 27.6 & 39.2 & 34.7 & 41.4 &  8.8 & 15.3 & 36.0 & 47.9 &  9.0 & 16.8 \\
w/o $w_t$               & 30.6 & 42.6 & 39.7 & 45.9 & 10.4 & 18.1 & 33.6 & 46.5 & 12.4 & 20.1 \\
\midrule
\ourMethod        & \textbf{32.9} & \textbf{44.2} & \textbf{42.3} & \textbf{47.8} & \textbf{10.8} & \textbf{18.6} & \textbf{37.6} & \textbf{49.3} & \textbf{12.9} & \textbf{20.7}
 \\
\bottomrule
\end{tabular}%
}
\caption{Ablation study on reward components.}
\label{tb:ablation}
\end{table}

We conduct a systematic ablation on \ourMethod{}, and report the results in Table~\ref{tb:ablation}.

To validate global planning, we ablate two plan-level rewards. Removing structural (\emph{w/o $R_{\mathrm{str}}$}) or semantic (\emph{w/o $R_{\mathrm{sem}}$}) supervision reduces EM/F1 across all datasets, with $R_{\mathrm{sem}}$ showing a larger degradation. This highlights the necessity of plan-level guidance: $R_{\mathrm{str}}$ controls long-range dependencies, while $R_{\mathrm{sem}}$ ensures faithful decomposition for accurate retrieval and reasoning.

Removing intermediate subgoal supervision (\emph{w/o $R_{\mathrm{step}}$}) yields the largest performance declines (averaging 17.7\% EM and 12.5\% F1 drops). This proves that validating intermediate execution is essential to ensure faithful plan execution and prevent partial answers from becoming final predictions.

Disabling progressive reward-weight annealing (\emph{w/o $w_t$}) also consistently degrades performance. This result suggests
that the $w_t$-based curriculum stabilizes optimization by progressively shifting focus from early planning guidance to outcome-driven rewards, mitigating noise from imperfect teacher trajectories (see Appendix~\ref{appendix:G} for plan diversity analysis).

Finally, extended analyses (Appendices~\ref{appendix:H}, \ref{appendix:I}, \ref{appendix:J}, \ref{appendix:K}) demonstrate \ourMethod{}'s robustness to hyperparameters, identifying optimal values at retrieval top-$k=3$, GRPO group size $5$, $\lambda=0.5$, $\gamma=0.5$, and transition point $T=50$.

\subsection{Algorithmic Attribution Study}
{
\begin{figure}[!t]
    \centering
    \includegraphics[width=0.95\linewidth]{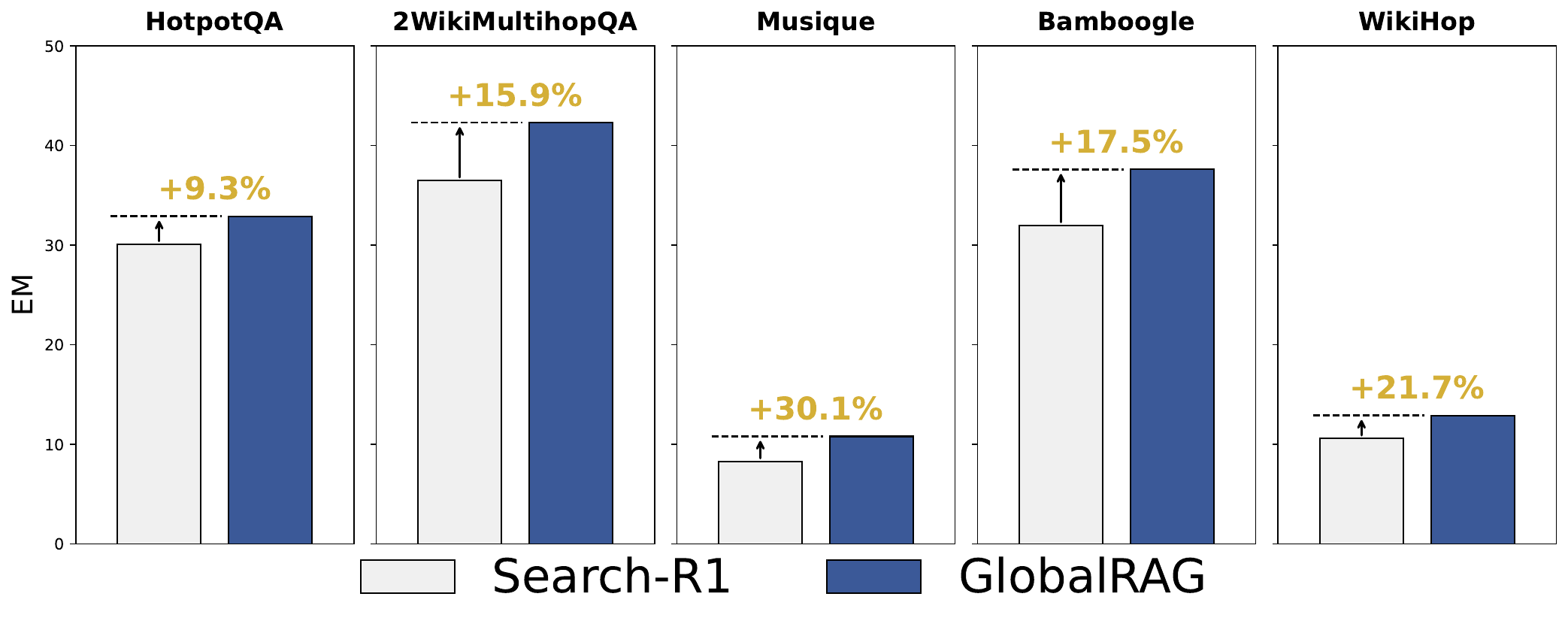}
    \caption{Comparison of the performance of our method GlobalRAG with the baseline Search-R1 on five benchmark datasets under identical datasets and unified experimental settings. (LLM: Qwen2.5-3B-Instruct)}
    \label{fig:algorithm}
\end{figure}

\begin{table}[!h]
\centering
\footnotesize
\setlength{\tabcolsep}{3.8pt}
\renewcommand{\arraystretch}{1.12}
\resizebox{\columnwidth}{!}{%
\begin{tabular}{lcccccccccc}
\toprule
\multirow{2}{*}{Methods} &
\multicolumn{2}{c}{\textbf{HotpotQA}} &
\multicolumn{2}{c}{\textbf{2Wiki}} &
\multicolumn{2}{c}{\textbf{MuSiQue}} &
\multicolumn{2}{c}{\textbf{Bamboogle}} &
\multicolumn{2}{c}{\textbf{WikiHop}} \\
\cmidrule(lr){2-3}\cmidrule(lr){4-5}\cmidrule(lr){6-7}\cmidrule(lr){8-9}\cmidrule(lr){10-11}
& EM & F1 & EM & F1 & EM & F1 & EM & F1 & EM & F1 \\
\midrule
\textbf{Qwen2.5-3b-Instruct} \\
\text{Search-R1} &30.1 &36.1 &36.5 &40.8 &8.3 &12.7 &32.0 &33.5 &10.6 &13.3 \\
\text{GlobalRAG} &32.9 &44.2 &42.3 &47.8 &10.8 &18.6 &37.6 &49.3 &12.9 &20.7 \\

\midrule
\textbf{Qwen2.5-7b-Instruct} \\
\text{Search-R1} &34.4 &45.6 &42.4 &47.6 &9.0 &16.7 &41.1 &49.3 &15.1 &21.9 \\

\text{GlobalRAG} &38.2 &50.7 &47.8 &53.2 &14.9 &21.2 &48.8 &61.6 &19.0 &28.0 \\

\midrule
\textbf{Llama3.1-8b-Instruct} \\
Search-R1 & 35.7 & 46.9 & 44.6 & 50.1 & 12.9 & 20.0 & 44.8 & 56.5 & 16.5 & 22.0 \\
GlobalRAG & 36.9 & 48.6 & 46.6 & 51.9 & 17.0 & 25.7 & 49.6 & 61.9 & 17.8 & 25.1 \\

\bottomrule
\end{tabular}%
}
\caption{Algorithmic attribution (controlled data): GlobalRAG vs. Search-R1.}
\label{tb:algorithmic}
\end{table}

To evaluate the intrinsic effectiveness of our proposed framework, we conducted an algorithmic attribution study comparing GlobalRAG against the Search-R1 baseline under strictly controlled experimental settings. As illustrated in Figure \ref{fig:algorithm}, GlobalRAG achieves a significant performance leap, specifically yielding a 9.3\% relative improvement in EM score when utilizing the Qwen2.5-3B-Instruct backbone. This performance gain, achieved using identical training datasets and the GRPO reinforcement learning algorithm, underscores that the advancements are primarily driven by our architectural innovations in global reasoning and evidence aggregation rather than data scaling. 

Notably, similar and consistent improvements are also observed when switching to the LLaMA3.1-8B-Instruct backbone, where GlobalRAG consistently outperforms Search-R1 across all five benchmarks, indicating that the proposed framework generalizes well across different model families. Specifically, GlobalRAG demonstrates superior proficiency in handling complex multi-hop queries where structured knowledge integration is paramount. Detailed analyses are provided in Appendix~\ref{appendix:L}.
% Detailed quantitative results across various model scales (3B, 7B, and 8B parameters) and all five benchmark datasets are provided in Appendix \ref{appendix:L}, which further confirm the consistency of these algorithmic advantages.
}

\subsection{Different RL Methods: PPO vs. GRPO}
{
\begin{figure}[h]
    \centering
    \includegraphics[width=0.95\linewidth]{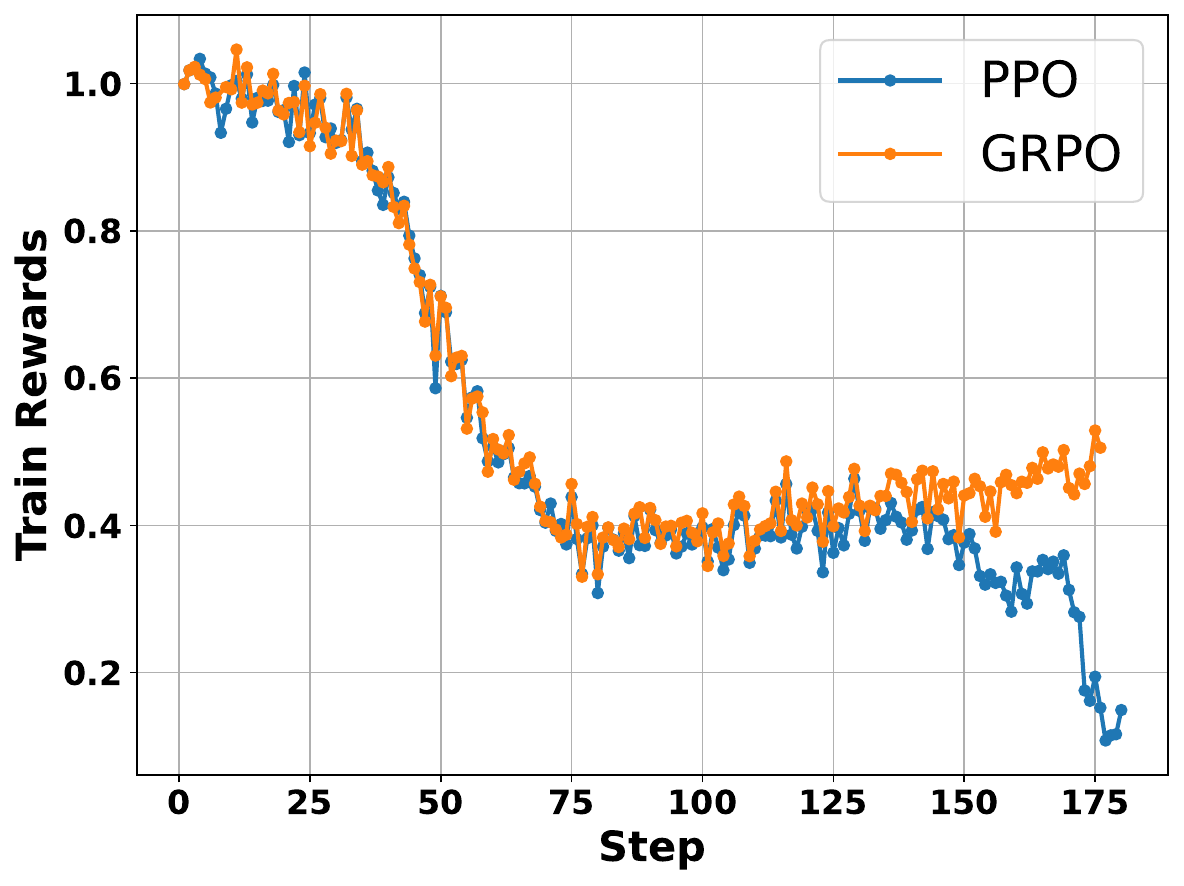}
    \caption{Training reward curves for PPO and GRPO. GRPO achieves higher rewards with faster convergence and more stable training.}
    \label{fig:RL-Methods}
\end{figure}

\begin{table}[h]
\centering
\footnotesize
\setlength{\tabcolsep}{3.8pt}
\renewcommand{\arraystretch}{1.12}
\resizebox{\columnwidth}{!}{%
\begin{tabular}{lcccccccccc}
\toprule
\multirow{2}{*}{Methods} &
\multicolumn{2}{c}{\textbf{HotpotQA}} &
\multicolumn{2}{c}{\textbf{2Wiki}} &
\multicolumn{2}{c}{\textbf{MuSiQue}} &
\multicolumn{2}{c}{\textbf{Bamboogle}} &
\multicolumn{2}{c}{\textbf{WikiHop}} \\
\cmidrule(lr){2-3}\cmidrule(lr){4-5}\cmidrule(lr){6-7}\cmidrule(lr){8-9}\cmidrule(lr){10-11}
 & EM & F1 & EM & F1 & EM & F1 & EM & F1 & EM & F1 \\
\midrule
PPO  & 29.5 & 40.6 & 37.5 & 43.6 & 9.1 & 15.1 & 32.0 & 45.2 & 10.3 & 17.9 \\
GRPO & 32.9 & 44.2 & 42.3 & 47.8 & 10.8 & 18.6 & 37.6 & 49.3 & 12.9 & 20.7 \\
\bottomrule
\end{tabular}%
}
\caption{Performance comparison of PPO and GRPO within the GlobalRAG framework on five multi-hop QA benchmarks, measured by EM and F1 scores.}
\label{tb:rl-methods}
\end{table}

We compare PPO and GRPO for training GlobalRAG. As shown in Figure \ref{fig:RL-Methods}, GRPO attains higher training rewards and converges more steadily than PPO, indicating a more effective optimization process.

Table \ref{tb:rl-methods} reports the downstream QA results. GRPO consistently outperforms PPO across all benchmarks on both EM and F1, with more pronounced gains on HotpotQA, 2WikiMultihopQA, and Bamboogle, which require multi-hop evidence aggregation and complex reasoning. Improvements on MuSiQue and WikiHop, though smaller in magnitude, remain consistent.

These results indicate that GRPO exhibits more stable training dynamics and superior generalization compared to PPO.
}

\subsection{Retrieval Performance for Multi-hop QA}
{
To explain why \ourMethod{} achieves stronger multi-hop QA performance, we analyze its retrieval behavior. As shown in Figure~\ref{fig:search_num}, both \ourMethod{} and StepSearch perform roughly twice as many retrievals as Search-R1 (about $1.2\times$ on WikiHop), indicating that incorporating process-oriented rewards, rather than relying solely on outcome-based rewards, more effectively encourages iterative retrieval and supports correct answer identification.

Notably, \ourMethod{} uses a comparable but slightly smaller number of retrievals than StepSearch (Figure~\ref{fig:search_num}), while achieving higher EM and F1 scores (Table~\ref{tab:main-results}). This suggests that the planning quality and subgoal completion rewards improve global planning and faithful execution, leading to higher accuracy without increasing retrieval steps.

\begin{figure}[!t]
    \centering
    \includegraphics[width=\linewidth]{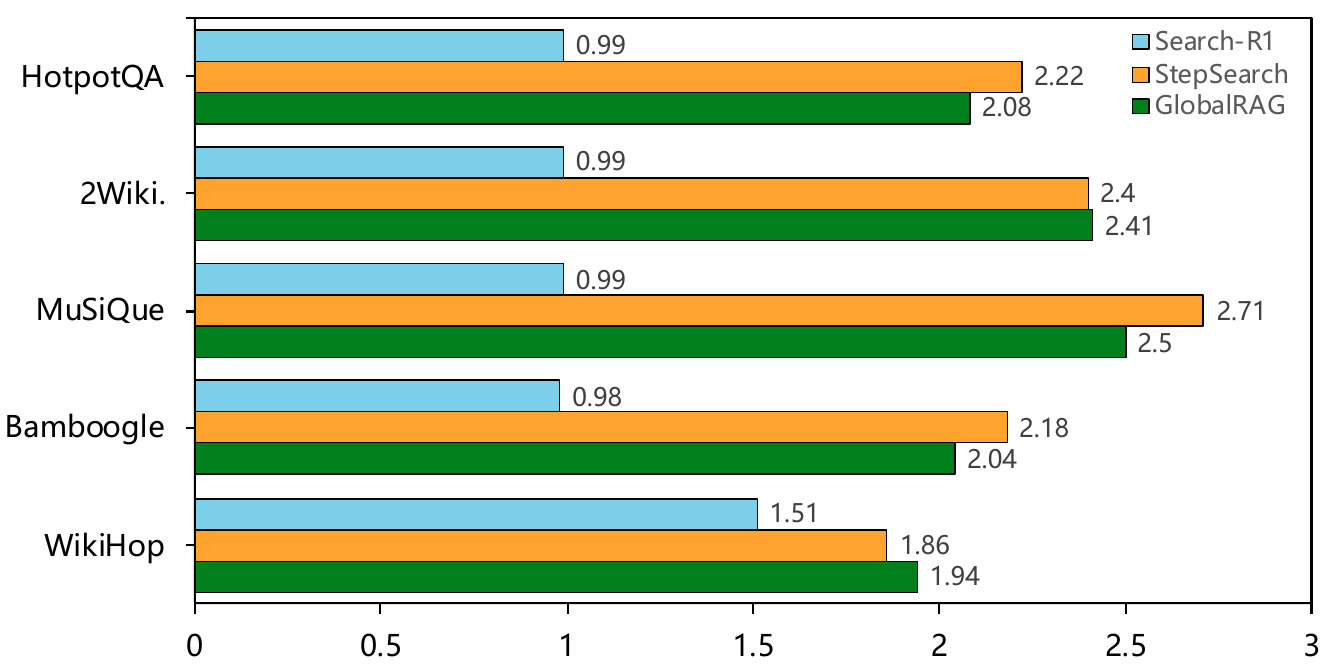}
    \caption{Comparison of average retrieval counts across datasets for Search-R1, StepSearch and GlobalRAG.}
    \label{fig:search_num}
\end{figure}
}

\section{Conclusions}

{
In this work, we presented GlobalRAG, a reinforcement learning framework for multi-hop question answering that explicitly enforces global planning and faithful execution. By modeling question decomposition as a structured plan and optimizing planning-quality and subgoal-completion rewards under a progressive annealing schedule, GlobalRAG aligns high-level reasoning intent with retrieval and execution behaviors. Extensive experiments across in-domain and out-of-domain benchmarks show that this plan-centric optimization consistently improves multi-hop reasoning accuracy without increasing retrieval cost. These results highlight the importance of supervising how reasoning unfolds, rather than optimizing final answers alone, and point to future work on more flexible planning and reasoning trajectories.
}
\section*{Limitations}

Our study has four main limitations. First, due to computational and cost constraints, we do not conduct reinforcement learning training on very large-scale models (e.g., 14B or 32B parameters). Whether the proposed framework can be effectively scaled to such model sizes while maintaining stable training dynamics and planning performance remains an open question. Second, our experiments focus on multi-hop question answering. We do not include a detailed analysis or dedicated training data generation for single-hop tasks, which may limit the applicability of our findings to other reasoning settings. Third, for complex multi-hop queries, multiple valid decomposition paths may exist. Although the similarity-based reward design allows partial credit for semantically aligned plans, the current training setup still optimizes toward a single reference trajectory. Explicitly modeling and optimizing over multiple valid planning paths is left for future work.   Finally, the framework improves planning accuracy primarily through reinforcement learning during training, but does not incorporate an explicit mechanism for real-time error correction at inference. If the model produces an incorrect initial decomposition, it cannot revise the global plan based on intermediate retrieval failures or reasoning inconsistencies, which remains a challenge for achieving more robust and adaptive multi-hop reasoning.

% \subsection{References}

% \nocite{Ando2005,andrew2007scalable,rasooli-tetrault-2015}

% The \LaTeX{} and Bib\TeX{} style files provided roughly follow the American Psychological Association format.
% If your own bib file is named \texttt{custom.bib}, then placing the following before any appendices in your \LaTeX{} file will generate the references section for you:
% \begin{quote}
% \begin{verbatim}
% \bibliography{custom}
% \end{verbatim}
% \end{quote}

% You can obtain the complete ACL Anthology as a Bib\TeX{} file from \url{https://aclweb.org/anthology/anthology.bib.gz}.
% To include both the Anthology and your own .bib file, use the following instead of the above.
% \begin{quote}
% \begin{verbatim}
% \end{verbatim}
% \end{quote}

% Please see Section~\ref{sec:bibtex} for information on preparing Bib\TeX{} files.

% \subsection{Equations}

% An example equation is shown below:
% \begin{equation}
%   \label{eq:example}
%   A = \pi r^2
% \end{equation}

% Labels for equation numbers, sections, subsections, figures and tables
% are all defined with the \verb|\label{label}| command and cross references
% to them are made with the \verb|\ref{label}| command.

% This an example cross-reference to Equation~\ref{eq:example}.

% \subsection{Appendices}

% Use \verb|\appendix| before any appendix section to switch the section numbering over to letters. See Appendix~\ref{sec:appendix} for an example.


\bibliography{custom}

@inproceedings{
jin-2025-searchr1,
title={Search-R1: Training {LLM}s to Reason and Leverage Search Engines with Reinforcement Learning},
author={Bowen Jin and Hansi Zeng and Zhenrui Yue and Jinsung Yoon and Sercan O Arik and Dong Wang and Hamed Zamani and Jiawei Han},
booktitle={Second Conference on Language Modeling},
year={2025},
url={https://openreview.net/forum?id=Rwhi91ideu}
}

@article{DBLP:journals/corr/abs-2503-05592,
  publtype={informal},
  author={Huatong Song and Jinhao Jiang and Yingqian Min and Jie Chen and Zhipeng Chen and Wayne Xin Zhao and Lei Fang and Ji-Rong Wen},
  title={R1-Searcher: Incentivizing the Search Capability in LLMs via Reinforcement Learning},
  year={2025},
  month={March},
  cdate={1740787200000},
  journal={CoRR},
  volume={abs/2503.05592},
  url={https://doi.org/10.48550/arXiv.2503.05592}
}

@article{song-2025-r1searcherpp,
  title   = {R1-Searcher++: Scaling Up LLMs as Effective Web Searchers via Reinforcement Learning},
  author  = {Huatong Song and others},
  journal = {arXiv preprint arXiv:2505.17005},
  year    = {2025},
  month   = may,
  url     = {https://arxiv.org/abs/2505.17005},
  doi     = {10.48550/arXiv.2505.17005}
}

@misc{fang-etal-2025-kirag,
      title={KiRAG: Knowledge-Driven Iterative Retriever for Enhancing Retrieval-Augmented Generation}, 
      author={Jinyuan Fang and Zaiqiao Meng and Craig Macdonald},
      year={2025},
      eprint={2502.18397},
      archivePrefix={arXiv},
      primaryClass={cs.CL},
      url={https://arxiv.org/abs/2502.18397}, 
}

@article{he-2025-tiresragr1,
  title   = {From Sufficiency to Reflection: Reinforcement-Guided Thinking Quality in Retrieval-Augmented Reasoning for LLMs},
  author  = {Jie He and others},
  journal = {arXiv preprint arXiv:2507.22716},
  year    = {2025},
  month   = jul,
  url     = {https://arxiv.org/abs/2507.22716},
  doi     = {10.48550/arXiv.2507.22716}
}

@misc{shi-2025-autorefine,
      title={Search and Refine During Think: Facilitating Knowledge Refinement for Improved Retrieval-Augmented Reasoning}, 
      author={Yaorui Shi and Sihang Li and Chang Wu and Zhiyuan Liu and Junfeng Fang and Hengxing Cai and An Zhang and Xiang Wang},
      year={2025},
      eprint={2505.11277},
      archivePrefix={arXiv},
      primaryClass={cs.CL},
      url={https://arxiv.org/abs/2505.11277}, 
}

@article{DBLP:journals/corr/abs-2507-23581,
  publtype={informal},
  author={Chuanyue Yu and Kuo Zhao and Yuhan Li and Heng Chang and Mingjian Feng and Xiangzhe Jiang and Yufei Sun and Jia Li and Yuzhi Zhang and Jianxin Li and Ziwei Zhang},
  title={GraphRAG-R1: Graph Retrieval-Augmented Generation with Process-Constrained Reinforcement Learning},
  year={2025},
  month={July},
  cdate={1751328000000},
  journal={CoRR},
  volume={abs/2507.23581},
  url={https://doi.org/10.48550/arXiv.2507.23581}
}

@article{guo-2025-deepseek-r1-nature,
  title={DeepSeek-R1 incentivizes reasoning in LLMs through reinforcement learning},
  author={Guo, Daya and Yang, Dejian and Zhang, Haowei and Song, Junxiao and Wang, Peiyi and Zhu, Qihao and Xu, Runxin and Zhang, Ruoyu and Ma, Shirong and Bi, Xiao and others},
  journal={Nature},
  volume={645},
  number={8081},
  pages={633--638},
  year={2025},
  publisher={Nature Publishing Group UK London}
}

@misc{openai2024openaio1card,
      title={OpenAI o1 System Card}, 
      author={OpenAI and : and Aaron Jaech and Adam Kalai and Adam Lerer and Adam Richardson and Ahmed El-Kishky and Aiden Low and Alec Helyar and Aleksander Madry and Alex Beutel and Alex Carney and Alex Iftimie and Alex Karpenko and Alex Tachard Passos and Alexander Neitz and Alexander Prokofiev and Alexander Wei and Allison Tam and Ally Bennett and Ananya Kumar and Andre Saraiva and Andrea Vallone and Andrew Duberstein and Andrew Kondrich and Andrey Mishchenko and Andy Applebaum and Angela Jiang and Ashvin Nair and Barret Zoph and Behrooz Ghorbani and Ben Rossen and Benjamin Sokolowsky and Boaz Barak and Bob McGrew and Borys Minaiev and Botao Hao and Bowen Baker and Brandon Houghton and Brandon McKinzie and Brydon Eastman and Camillo Lugaresi and Cary Bassin and Cary Hudson and Chak Ming Li and Charles de Bourcy and Chelsea Voss and Chen Shen and Chong Zhang and Chris Koch and Chris Orsinger and Christopher Hesse and Claudia Fischer and Clive Chan and Dan Roberts and Daniel Kappler and Daniel Levy and Daniel Selsam and David Dohan and David Farhi and David Mely and David Robinson and Dimitris Tsipras and Doug Li and Dragos Oprica and Eben Freeman and Eddie Zhang and Edmund Wong and Elizabeth Proehl and Enoch Cheung and Eric Mitchell and Eric Wallace and Erik Ritter and Evan Mays and Fan Wang and Felipe Petroski Such and Filippo Raso and Florencia Leoni and Foivos Tsimpourlas and Francis Song and Fred von Lohmann and Freddie Sulit and Geoff Salmon and Giambattista Parascandolo and Gildas Chabot and Grace Zhao and Greg Brockman and Guillaume Leclerc and Hadi Salman and Haiming Bao and Hao Sheng and Hart Andrin and Hessam Bagherinezhad and Hongyu Ren and Hunter Lightman and Hyung Won Chung and Ian Kivlichan and Ian O'Connell and Ian Osband and Ignasi Clavera Gilaberte and Ilge Akkaya and Ilya Kostrikov and Ilya Sutskever and Irina Kofman and Jakub Pachocki and James Lennon and Jason Wei and Jean Harb and Jerry Twore and Jiacheng Feng and Jiahui Yu and Jiayi Weng and Jie Tang and Jieqi Yu and Joaquin Quiñonero Candela and Joe Palermo and Joel Parish and Johannes Heidecke and John Hallman and John Rizzo and Jonathan Gordon and Jonathan Uesato and Jonathan Ward and Joost Huizinga and Julie Wang and Kai Chen and Kai Xiao and Karan Singhal and Karina Nguyen and Karl Cobbe and Katy Shi and Kayla Wood and Kendra Rimbach and Keren Gu-Lemberg and Kevin Liu and Kevin Lu and Kevin Stone and Kevin Yu and Lama Ahmad and Lauren Yang and Leo Liu and Leon Maksin and Leyton Ho and Liam Fedus and Lilian Weng and Linden Li and Lindsay McCallum and Lindsey Held and Lorenz Kuhn and Lukas Kondraciuk and Lukasz Kaiser and Luke Metz and Madelaine Boyd and Maja Trebacz and Manas Joglekar and Mark Chen and Marko Tintor and Mason Meyer and Matt Jones and Matt Kaufer and Max Schwarzer and Meghan Shah and Mehmet Yatbaz and Melody Y. Guan and Mengyuan Xu and Mengyuan Yan and Mia Glaese and Mianna Chen and Michael Lampe and Michael Malek and Michele Wang and Michelle Fradin and Mike McClay and Mikhail Pavlov and Miles Wang and Mingxuan Wang and Mira Murati and Mo Bavarian and Mostafa Rohaninejad and Nat McAleese and Neil Chowdhury and Neil Chowdhury and Nick Ryder and Nikolas Tezak and Noam Brown and Ofir Nachum and Oleg Boiko and Oleg Murk and Olivia Watkins and Patrick Chao and Paul Ashbourne and Pavel Izmailov and Peter Zhokhov and Rachel Dias and Rahul Arora and Randall Lin and Rapha Gontijo Lopes and Raz Gaon and Reah Miyara and Reimar Leike and Renny Hwang and Rhythm Garg and Robin Brown and Roshan James and Rui Shu and Ryan Cheu and Ryan Greene and Saachi Jain and Sam Altman and Sam Toizer and Sam Toyer and Samuel Miserendino and Sandhini Agarwal and Santiago Hernandez and Sasha Baker and Scott McKinney and Scottie Yan and Shengjia Zhao and Shengli Hu and Shibani Santurkar and Shraman Ray Chaudhuri and Shuyuan Zhang and Siyuan Fu and Spencer Papay and Steph Lin and Suchir Balaji and Suvansh Sanjeev and Szymon Sidor and Tal Broda and Aidan Clark and Tao Wang and Taylor Gordon and Ted Sanders and Tejal Patwardhan and Thibault Sottiaux and Thomas Degry and Thomas Dimson and Tianhao Zheng and Timur Garipov and Tom Stasi and Trapit Bansal and Trevor Creech and Troy Peterson and Tyna Eloundou and Valerie Qi and Vineet Kosaraju and Vinnie Monaco and Vitchyr Pong and Vlad Fomenko and Weiyi Zheng and Wenda Zhou and Wes McCabe and Wojciech Zaremba and Yann Dubois and Yinghai Lu and Yining Chen and Young Cha and Yu Bai and Yuchen He and Yuchen Zhang and Yunyun Wang and Zheng Shao and Zhuohan Li},
      year={2024},
      eprint={2412.16720},
      archivePrefix={arXiv},
      primaryClass={cs.AI},
      url={https://arxiv.org/abs/2412.16720}, 
}

@misc{chen2025researchlearningreasonsearch,
      title={ReSearch: Learning to Reason with Search for LLMs via Reinforcement Learning}, 
      author={Mingyang Chen and Tianpeng Li and Haoze Sun and Yijie Zhou and Chenzheng Zhu and Haofen Wang and Jeff Z. Pan and Wen Zhang and Huajun Chen and Fan Yang and Zenan Zhou and Weipeng Chen},
      year={2025},
      eprint={2503.19470},
      archivePrefix={arXiv},
      primaryClass={cs.AI},
      url={https://arxiv.org/abs/2503.19470}, 
}

@article{wang-2022-e5,
  title   = {Text Embeddings by Weakly-Supervised Contrastive Pre-training},
  author  = {Wang, Hongxin and He, Zheng and Chen, Jing and others},
  journal = {arXiv preprint arXiv:2212.03533},
  year    = {2022},
  url     = {https://arxiv.org/abs/2212.03533}
}

@inproceedings{cheng-2025-dualrag,
  title   = {DualRAG: A Dual-Process Approach to Integrate Reasoning and Retrieval for Multi-Hop Question Answering},
  author  = {Cheng, Rong and Liu, Jinyi and Zheng, Yan and Ni, Fei and Du, Jiazhen and Mao, Hangyu and Zhang, Fuzheng and Wang, Bo and Hao, Jianye},
  booktitle = {Proceedings of the 63rd Annual Meeting of the Association for Computational Linguistics (ACL)},
  year    = {2025},
  url     = {https://aclanthology.org/2025.acl-long.1539/}
}

@inproceedings{liu2025beyond,
  title={Beyond the Answer: Advancing Multi-Hop QA with Fine-Grained Graph Reasoning and Evaluation},
  author={Liu, Qichuan and Zhang, Chentao and Zheng, Chenfeng and Hu, Guosheng and Li, Xiaodong and Zhang, Zhihong},
  booktitle={Proceedings of the 63rd Annual Meeting of the Association for Computational Linguistics (Volume 1: Long Papers)},
  url={https://aclanthology.org/2025.acl-long.1142.pdf},
  pages={23433--23456},
  year={2025}
}

@inproceedings{yang-etal-2018-hotpotqa,
    title = "{H}otpot{QA}: A Dataset for Diverse, Explainable Multi-hop Question Answering",
    author = "Yang, Zhilin  and
      Qi, Peng  and
      Zhang, Saizheng  and
      Bengio, Yoshua  and
      Cohen, William  and
      Salakhutdinov, Ruslan  and
      Manning, Christopher D.",
    editor = "Riloff, Ellen  and
      Chiang, David  and
      Hockenmaier, Julia  and
      Tsujii, Jun{'}ichi",
    booktitle = "Proceedings of the 2018 Conference on Empirical Methods in Natural Language Processing",
    month = oct # "-" # nov,
    year = "2018",
    address = "Brussels, Belgium",
    publisher = "Association for Computational Linguistics",
    url = "https://aclanthology.org/D18-1259/",
    doi = "10.18653/v1/D18-1259",
    pages = "2369--2380"
}

@article{trivedi-etal-2022-musique,
    title = "{M}u{S}i{Q}ue: Multihop Questions via Single-hop Question Composition",
    author = "Trivedi, Harsh  and
      Balasubramanian, Niranjan  and
      Khot, Tushar  and
      Sabharwal, Ashish",
    editor = "Roark, Brian  and
      Nenkova, Ani",
    journal = "Transactions of the Association for Computational Linguistics",
    volume = "10",
    year = "2022",
    address = "Cambridge, MA",
    publisher = "MIT Press",
    url = "https://aclanthology.org/2022.tacl-1.31/",
    doi = "10.1162/tacl_a_00475",
    pages = "539--554"
}

@article{welbl-etal-2018-constructing,
    title = "Constructing Datasets for Multi-hop Reading Comprehension Across Documents",
    author = "Welbl, Johannes  and
      Stenetorp, Pontus  and
      Riedel, Sebastian",
    editor = "Lee, Lillian  and
      Johnson, Mark  and
      Toutanova, Kristina  and
      Roark, Brian",
    journal = "Transactions of the Association for Computational Linguistics",
    volume = "6",
    year = "2018",
    address = "Cambridge, MA",
    publisher = "MIT Press",
    url = "https://aclanthology.org/Q18-1021/",
    doi = "10.1162/tacl_a_00021",
    pages = "287--302"
}

@inproceedings{trivedi-etal-2023-interleaving,
    title = "Interleaving Retrieval with Chain-of-Thought Reasoning for Knowledge-Intensive Multi-Step Questions",
    author = "Trivedi, Harsh  and
      Balasubramanian, Niranjan  and
      Khot, Tushar  and
      Sabharwal, Ashish",
    editor = "Rogers, Anna  and
      Boyd-Graber, Jordan  and
      Okazaki, Naoaki",
    booktitle = "Proceedings of the 61st Annual Meeting of the Association for Computational Linguistics (Volume 1: Long Papers)",
    month = jul,
    year = "2023",
    address = "Toronto, Canada",
    publisher = "Association for Computational Linguistics",
    url = "https://aclanthology.org/2023.acl-long.557/",
    doi = "10.18653/v1/2023.acl-long.557",
    pages = "10014--10037"
}

@article{li2025search,
  title={Search-o1: Agentic search-enhanced large reasoning models},
  author={Li, Xiaoxi and Dong, Guanting and Jin, Jiajie and Zhang, Yuyao and Zhou, Yujia and Zhu, Yutao and Zhang, Peitian and Dou, Zhicheng},
  journal={arXiv preprint arXiv:2501.05366},
  year={2025}
}

@inproceedings{HippoRAG-2024,
author = {Guti\'{e}rrez, Bernal Jim\'{e}nez and Shu, Yiheng and Gu, Yu and Yasunaga, Michihiro and Su, Yu},
title = {HippoRAG: neurobiologically inspired long-term memory for large language models},
year = {2025},
isbn = {9798331314385},
publisher = {Curran Associates Inc.},
address = {Red Hook, NY, USA},
booktitle = {Proceedings of the 38th International Conference on Neural Information Processing Systems},
articleno = {1902},
numpages = {38},
location = {Vancouver, BC, Canada},
series = {NIPS '24}
}

@inproceedings{hippo2-gutierrez2025,
title={From {RAG} to Memory: Non-Parametric Continual Learning for Large Language Models},
author={Bernal Jim{\'e}nez Guti{\'e}rrez and Yiheng Shu and Weijian Qi and Sizhe Zhou and Yu Su},
booktitle={Forty-second International Conference on Machine Learning},
year={2025},
url={https://openreview.net/forum?id=LWH8yn4HS2}
}

@inproceedings{press-etal-2023-measuring-data,
    title = "Measuring and Narrowing the Compositionality Gap in Language Models",
    author = "Press, Ofir  and
      Zhang, Muru  and
      Min, Sewon  and
      Schmidt, Ludwig  and
      Smith, Noah  and
      Lewis, Mike",
    editor = "Bouamor, Houda  and
      Pino, Juan  and
      Bali, Kalika",
    booktitle = "Findings of the Association for Computational Linguistics: EMNLP 2023",
    month = dec,
    year = "2023",
    address = "Singapore",
    publisher = "Association for Computational Linguistics",
    url = "https://aclanthology.org/2023.findings-emnlp.378/",
    doi = "10.18653/v1/2023.findings-emnlp.378",
    pages = "5687--5711"
}

@inproceedings{ho-etal-2020-constructing-data,
    title = "Constructing A Multi-hop {QA} Dataset for Comprehensive Evaluation of Reasoning Steps",
    author = "Ho, Xanh  and
      Duong Nguyen, Anh-Khoa  and
      Sugawara, Saku  and
      Aizawa, Akiko",
    editor = "Scott, Donia  and
      Bel, Nuria  and
      Zong, Chengqing",
    booktitle = "Proceedings of the 28th International Conference on Computational Linguistics",
    month = dec,
    year = "2020",
    address = "Barcelona, Spain (Online)",
    publisher = "International Committee on Computational Linguistics",
    url = "https://aclanthology.org/2020.coling-main.580/",
    doi = "10.18653/v1/2020.coling-main.580",
    pages = "6609--6625"
}

@misc{anonymous2025stepsearch,
    title={StepSearch: Igniting LLMs Search Ability via Step-Wise Proximal Policy Optimization}, 
    author={Ziliang Wang and Xuhui Zheng and Kang An and Cijun Ouyang and Jialu Cai and Yuhang Wang and Yichao Wu},
    year={2025},
    eprint={2505.15107},
    archivePrefix={arXiv},
    primaryClass={cs.CL},
    url={https://arxiv.org/abs/2505.15107}, 
}

@misc{dong2025agentic,
    title={Agentic Reinforced Policy Optimization},
    author={Guanting Dong and Hangyu Mao and Kai Ma and Licheng Bao and Yifei Chen and Zhongyuan Wang and Zhongxia Chen and Jiazhen Du and Huiyang Wang and Fuzheng Zhang and Guorui Zhou and Yutao Zhu and Ji-Rong Wen and Zhicheng Dou},
    year={2025},
    eprint={2507.19849},
    archivePrefix={arXiv},
    primaryClass={cs.LG}
}

@misc{lai2025computerrl,
    title={ComputerRL: Scaling End-to-End Online Reinforcement Learning for Computer Use Agents},
    author={Hanyu Lai and Xiao Liu and Yanxiao Zhao and Han Xu and Hanchen Zhang and Bohao Jing and Yanyu Ren and Shuntian Yao and Yuxiao Dong and Jie Tang},
    year={2025},
    eprint={2508.14040},
    archivePrefix={arXiv},
    primaryClass={cs.AI}
}

@misc{shao2024deepseekmath,
      title={DeepSeekMath: Pushing the Limits of Mathematical Reasoning in Open Language Models}, 
      author={Zhihong Shao and Peiyi Wang and Qihao Zhu and Runxin Xu and Junxiao Song and Xiao Bi and Haowei Zhang and Mingchuan Zhang and Y. K. Li and Y. Wu and Daya Guo},
      year={2024},
      eprint={2402.03300},
      archivePrefix={arXiv},
      primaryClass={cs.CL},
      url={https://arxiv.org/abs/2402.03300}, 
}

@misc{zhang2025lets,
    title={LeTS: Learning to Think-and-Search via Process-and-Outcome Reward Hybridization},
    author={Qi Zhang and Shouqing Yang and Lirong Gao and Hao Chen and Xiaomeng Hu and Jinglei Chen and Jiexiang Wang and Sheng Guo and Bo Zheng and Haobo Wang and Junbo Zhao},
    year={2025},
    eprint={2505.17447},
    archivePrefix={arXiv},
    primaryClass={cs.CL}
}

@article{zhao2025parallelsearch,
  title={ParallelSearch: Train your LLMs to Decompose Query and Search Sub-queries in Parallel with Reinforcement Learning},
  author={Zhao, Shu and Yu, Tan and Xu, Anbang and Singh, Japinder and Shukla, Aaditya and Akkiraju, Rama},
  journal={arXiv preprint arXiv:2508.09303},
  year={2025}
}

@inproceedings{10.5555/3495724.3496517,
author = {Lewis, Patrick and Perez, Ethan and Piktus, Aleksandra and Petroni, Fabio and Karpukhin, Vladimir and Goyal, Naman and K\"{u}ttler, Heinrich and Lewis, Mike and Yih, Wen-tau and Rockt\"{a}schel, Tim and Riedel, Sebastian and Kiela, Douwe},
title = {Retrieval-augmented generation for knowledge-intensive NLP tasks},
year = {2020},
isbn = {9781713829546},
publisher = {Curran Associates Inc.},
address = {Red Hook, NY, USA},
booktitle = {Proceedings of the 34th International Conference on Neural Information Processing Systems},
articleno = {793},
numpages = {16},
location = {Vancouver, BC, Canada},
series = {NIPS '20}
}

@inproceedings{zhang-etal-2024-end,
    title = "End-to-End Beam Retrieval for Multi-Hop Question Answering",
    author = "Zhang, Jiahao  and
      Zhang, Haiyang  and
      Zhang, Dongmei  and
      Yong, Liu  and
      Huang, Shen",
    editor = "Duh, Kevin  and
      Gomez, Helena  and
      Bethard, Steven",
    booktitle = "Proceedings of the 2024 Conference of the North American Chapter of the Association for Computational Linguistics: Human Language Technologies (Volume 1: Long Papers)",
    month = jun,
    year = "2024",
    address = "Mexico City, Mexico",
    publisher = "Association for Computational Linguistics",
    url = "https://aclanthology.org/2024.naacl-long.96/",
    doi = "10.18653/v1/2024.naacl-long.96",
    pages = "1718--1731"
}

@inproceedings{li-du-2023-leveraging,
    title = "Leveraging Structured Information for Explainable Multi-hop Question Answering and Reasoning",
    author = "Li, Ruosen  and
      Du, Xinya",
    editor = "Bouamor, Houda  and
      Pino, Juan  and
      Bali, Kalika",
    booktitle = "Findings of the Association for Computational Linguistics: EMNLP 2023",
    month = dec,
    year = "2023",
    address = "Singapore",
    publisher = "Association for Computational Linguistics",
    url = "https://aclanthology.org/2023.findings-emnlp.452/",
    doi = "10.18653/v1/2023.findings-emnlp.452",
    pages = "6779--6789"
}

@inproceedings{shi-etal-2024-generate,
    title = "Generate-then-Ground in Retrieval-Augmented Generation for Multi-hop Question Answering",
    author = "Shi, Zhengliang  and
      Zhang, Shuo  and
      Sun, Weiwei  and
      Gao, Shen  and
      Ren, Pengjie  and
      Chen, Zhumin  and
      Ren, Zhaochun",
    editor = "Ku, Lun-Wei  and
      Martins, Andre  and
      Srikumar, Vivek",
    booktitle = "Proceedings of the 62nd Annual Meeting of the Association for Computational Linguistics (Volume 1: Long Papers)",
    month = aug,
    year = "2024",
    address = "Bangkok, Thailand",
    publisher = "Association for Computational Linguistics",
    url = "https://aclanthology.org/2024.acl-long.397/",
    doi = "10.18653/v1/2024.acl-long.397",
    pages = "7339--7353"
}

@article{survey_of_natural_language_generation,
author = {Ji, Ziwei and Lee, Nayeon and Frieske, Rita and Yu, Tiezheng and Su, Dan and Xu, Yan and Ishii, Etsuko and Bang, Ye Jin and Madotto, Andrea and Fung, Pascale},
title = {Survey of Hallucination in Natural Language Generation},
year = {2023},
issue_date = {December 2023},
publisher = {Association for Computing Machinery},
address = {New York, NY, USA},
volume = {55},
number = {12},
issn = {0360-0300},
url = {https://doi.org/10.1145/3571730},
doi = {10.1145/3571730},
journal = {ACM Comput. Surv.},
month = mar,
articleno = {248},
numpages = {38}
}

@INPROCEEDINGS{10448015,
  author={Feng, Zhangyin and Feng, Xiaocheng and Zhao, Dezhi and Yang, Maojin and Qin, Bing},
  booktitle={ICASSP 2024 - 2024 IEEE International Conference on Acoustics, Speech and Signal Processing (ICASSP)}, 
  title={Retrieval-Generation Synergy Augmented Large Language Models}, 
  year={2024},
  volume={},
  number={},
  pages={11661-11665},
  doi={10.1109/ICASSP48485.2024.10448015}}

@inproceedings{liu-etal-2024-ra,
    title = "{RA}-{ISF}: Learning to Answer and Understand from Retrieval Augmentation via Iterative Self-Feedback",
    author = "Liu, Yanming  and
      Peng, Xinyue  and
      Zhang, Xuhong  and
      Liu, Weihao  and
      Yin, Jianwei  and
      Cao, Jiannan  and
      Du, Tianyu",
    editor = "Ku, Lun-Wei  and
      Martins, Andre  and
      Srikumar, Vivek",
    booktitle = "Findings of the Association for Computational Linguistics: ACL 2024",
    month = aug,
    year = "2024",
    address = "Bangkok, Thailand",
    publisher = "Association for Computational Linguistics",
    url = "https://aclanthology.org/2024.findings-acl.281/",
    doi = "10.18653/v1/2024.findings-acl.281",
    pages = "4730--4749"
}

@inproceedings{shao-etal-2023-enhancing,
    title = "Enhancing Retrieval-Augmented Large Language Models with Iterative Retrieval-Generation Synergy",
    author = "Shao, Zhihong  and
      Gong, Yeyun  and
      Shen, Yelong  and
      Huang, Minlie  and
      Duan, Nan  and
      Chen, Weizhu",
    editor = "Bouamor, Houda  and
      Pino, Juan  and
      Bali, Kalika",
    booktitle = "Findings of the Association for Computational Linguistics: EMNLP 2023",
    month = dec,
    year = "2023",
    address = "Singapore",
    publisher = "Association for Computational Linguistics",
    url = "https://aclanthology.org/2023.findings-emnlp.620/",
    doi = "10.18653/v1/2023.findings-emnlp.620",
    pages = "9248--9274"
}

@inproceedings{retrieve_summarize_plan,
author = {Jiang, Zhouyu and Sun, Mengshu and Liang, Lei and Zhang, Zhiqiang},
title = {Retrieve, Summarize, Plan: Advancing Multi-hop Question Answering with an Iterative Approach},
year = {2025},
isbn = {9798400713316},
publisher = {Association for Computing Machinery},
address = {New York, NY, USA},
url = {https://doi.org/10.1145/3701716.3716889},
doi = {10.1145/3701716.3716889},
booktitle = {Companion Proceedings of the ACM on Web Conference 2025},
pages = {1677–1686},
numpages = {10},
location = {Sydney NSW, Australia},
series = {WWW '25}
}

@inproceedings{jin2025flashrag,
  title={Flashrag: A modular toolkit for efficient retrieval-augmented generation research},
  author={Jin, Jiajie and Zhu, Yutao and Dou, Zhicheng and Dong, Guanting and Yang, Xinyu and Zhang, Chenghao and Zhao, Tong and Yang, Zhao and Wen, Ji-Rong},
  booktitle={Companion Proceedings of the ACM on Web Conference 2025},
  pages={737--740},
  year={2025}
}

@article{chen2019understanding,
  title={Understanding dataset design choices for multi-hop reasoning},
  author={Chen, Jifan and Durrett, Greg},
  journal={arXiv preprint arXiv:1904.12106},
  year={2019}
}

% \documentclass[11pt]{article}

% % Change "review" to "final" to generate the final (sometimes called camera-ready) version.
% % Change to "preprint" to generate a non-anonymous version with page numbers.
% \usepackage[final]{acl}
% \usepackage{multirow}

% % Standard package includes
% \usepackage{times}
% \usepackage{latexsym}
% \usepackage{amsmath}
% \usepackage[most]{tcolorbox}
% % For proper rendering and hyphenation of words containing Latin characters (including in bib files)
% \usepackage[T1]{fontenc}

% % This assumes your files are encoded as UTF8
% \usepackage[utf8]{inputenc}
% \usepackage{amssymb}

% % This is not strictly necessary, and may be commented out,
% % but it will improve the layout of the manuscript,
% % and will typically save some space.
% \usepackage{microtype}

% % This is also not strictly necessary, and may be commented out.
% % However, it will improve the aesthetics of text in
% % the typewriter font.
% \usepackage{inconsolata}

% %Including images in your LaTeX document requires adding
% %additional package(s)
% \usepackage{graphicx}
% \usepackage{subcaption}
% % \usepackage{svg}  % 注释掉svg包，可能引起问题

% % define color
% \usepackage{array}
% \usepackage{xcolor}
% \usepackage{colortbl}
% \usepackage{longtable}
% \usepackage{booktabs}

% \usepackage{appendix}
% \newcommand{\ourMethod}{\textbf{GlobalRAG}\ }

% % additional
% \usepackage[breakable]{tcolorbox}

% \begin{document}

\clearpage
\appendix
\section{Prompt for Rollout}
\label{appendix:A}
{
The prompt we use for trajectory generation is shown in Figure~\ref{figure:prompt}.

\tcbset{
    mybox/.style={
        enhanced,
        colback=gray!10,           
        colframe=gray,           
        boxrule=1pt,               
        arc=0pt,                   
        left=8pt, right=8pt, top=8pt, bottom=8pt, 
        fonttitle=\bfseries,       
        coltitle=white,            
        attach boxed title to top center={
            yshift=-3pt,
        },
        boxed title style={
            colback=black,         
            colframe=black,        
            boxrule=1pt,
            sharp corners,         
            left=8pt, right=8pt, top=4pt, bottom=4pt 
        },
        before skip=0pt,          
        after skip=5pt            
    }
}

\begin{figure}[!h]
\centering
\begin{tcolorbox}[
    mybox,
    breakable,
    title=Prompt Used for Trajectory Generation,
]
Answer the given question by following the steps below. You must conduct all reasoning inside <think> and </think> before producing <plan>, <search>, <subAnswer> and <answer>.
\medskip

\textbf{Step 1:} Explicitly generate one or more sub-questions within the <plan> and </plan> block.
\begin{itemize}
    \item Each sub-question must contain both a question and a placeholder (\#1, \#2, etc.) that represents the answer to that question.
    \item Each sub-question should be as brief and precise as possible.
    \item If a sub-question depends on the answer to a previous one, use a placeholder (\#1, \#2, etc.) to represent that dependency.
    % 剩余内容...
\end{itemize}

\textbf{Step 2:} For each sub-question, create a block enclosed in <subPlan> and </subPlan>.
\begin{itemize}
    \item In sequential order, take one sub-question from <plan> and fill it between <search> and </search>.
    \item Conclude the block with a <subAnswer> that binds the answer to the current sub-question.
\end{itemize}
% 剩余内容...

\textbf{Step 3:} Provide the final result inside <answer> and </answer>, without detailed explanations.

\bigskip
\#\# Now, it's your turn! Please answer the following question!!!
\medskip

\textbf{Question:} \{Question\}
\end{tcolorbox}
\caption{The prompt used for trajectory generation}
\label{figure:prompt}
\end{figure}
% 使用 figure* 环境来实现跨双栏的效果
% \begin{figure}
% \centering

% \begin{tcolorbox}[mybox, title=Step 1: Question Decomposition, width=0.48\columnwidth]
% \textbf{Instruction:} Explicitly generate sub-questions within \textless plan\textgreater blocks.

% \textbf{Output Format:}
% \begin{verbatim}
% {
%     "Q1": ["First sub-question", "#1"],
%     "Q2": ["Second sub-question using #1", "#2"]
% }
% \end{verbatim}
% \end{tcolorbox}
% \hfill
% \begin{tcolorbox}[mybox, title=Step 2: Information Search, width=0.48\columnwidth]
% \textbf{Instruction:} For each sub-question, search using \textless search\textgreater query \textless/search\textgreater and provide \textless subAnswer\textgreater.
% \end{tcolorbox}

% \vspace{0.3cm}

% \begin{tcolorbox}[mybox, title=Step 3: Final Answer Generation, width=\columnwidth]
% \textbf{Instruction:} Provide the final result inside \textless answer\textgreater and \textless/answer\textgreater, combining all sub-answers without detailed explanations.

% \textbf{Question:} \{Question\}
% \end{tcolorbox}

% \caption{The prompts used in our multi-step reasoning framework}
% \label{fig:prompts}
% \end{figure}
% }

\section{Analysis of RL-based RAG Problems}
\label{appendix:B}
{
Following this analysis, we present a quantitative breakdown of Search-R1’s failure modes across datasets. From up to 300 sampled instances per dataset (1325 instances in total, including all 125 instances from Bamboogle), Search-R1 fails on 1072 cases. Among these errors, 44.96\% are attributed to the absence of global planning, while 28.45\% result from unfaithful execution.

Specifically, we sampled 300 instances from each of five public multi-hop reasoning benchmarks—HotpotQA, 2WikiMultiHopQA (2Wiki), MuSiQue, Bamboogle, and WikiHop. For Bamboogle, which consists of only 125 instances, we evaluated on the full set. To identify primary failure modes, we ran Search-R1 on each 300-item subset and isolated error cases: 219 (HotpotQA), 214 (2Wiki), 275 (MuSiQue), 90 (Bamboogle), and 274 (WikiHop). For each error, we then assessed whether the gold answer appeared in any document retrieved across multiple search rounds. If the answer never appeared, we labeled the case as a failure to retrieve answer-bearing evidence. The prevalence of this retrieval-failure criterion was 58.90\% (HotpotQA), 44.39\% (2Wiki), 85.45\% (MuSiQue), 76.67\% (Bamboogle), and 85.04\% (WikiHop).

We further analyzed the reasoning traces of the “no-answer-retrieved” subset using DeepSeek-R1 and identified two dominant error categories: (i) absence of global planning, where the model fails to construct an overall plan for decomposing and solving the question; and (ii) unfaithful execution, where retrieval drifts from the original target and results in missing the correct document. The distributions were as follows—HotpotQA: global-planning absence, 63.57\%; unfaithful execution, 36.43\%; other, 0.00\%. 2Wiki: 50.53\% vs. 47.37\% vs. 2.10\%; MuSiQue: 67.23\% vs. 30.64\% vs. 2.13\%; Bamboogle: 56.52\% vs. 37.68\% vs. 5.80\%; WikiHop: 56.57\% vs. 41.97\% vs. 1.46\%. These results indicate that, when answer-bearing contexts are not retrieved, the absence of global planning and unfaithful execution constitute the predominant sources of failure.
}

\section{Detailed GRPO Adaptation for \ourMethod{}}
\label{appendix:C}

\subsection{Token-level Policy Optimization with Masked GRPO}

To integrate GRPO with our \ourMethod{}, we adopt the token-masking strategy from Search-R1~\cite{jin-2025-searchr1}, where all tokens between \texttt{<information>} and \texttt{</information>} are masked. This ensures that the policy gradient is computed solely on tokens generated by the language model, excluding retrieved or externally provided content from optimization.

The GRPO training objective is defined as follows:

\begin{multline}
J_{\text{GRPO}}(\theta) = \mathbb{E}_{(q, a) \sim \mathcal{D}, \{o_i\}_{i=1}^G \sim \pi_{\theta_{\text{old}}}(\cdot|q)} \Bigg[ \\
\frac{1}{G} \sum_{i=1}^G \frac{1}{|o_i|} \sum_{t=1}^{|o_i|} \min \Bigg(
\frac{\pi_\theta(o_{i,t} \mid q, o_{i,<t})}{\pi_{\theta_{\text{old}}}(o_{i,t} \mid q, o_{i,<t})} \hat{A}_{i,t}, \\
\text{clip}\Big(\frac{\pi_\theta(o_{i,t} \mid q, o_{i,<t})}{\pi_{\theta_{\text{old}}}(o_{i,t} \mid q, o_{i,<t})}, 1 - \varepsilon, 1 + \varepsilon \Big) \hat{A}_{i,t}
\Bigg) \\
- \beta D_{\text{KL}}\big[\pi_\theta \parallel \pi_{\text{ref}}\big] \Bigg]
\end{multline}

Here, $\hat{A}_{i,t}$ denotes the normalized token-level advantage for the $t$-th token in the $i$-th trajectory within a group of size $G$, computed as
\[
\hat{A}_{i,t} = \frac{R_{i,t} - \mathrm{mean}(R_t)}{\mathrm{std}(R_t)}.
\]
The parameter $\varepsilon$ controls the clipping range for the policy ratio, and $\beta$ weights the KL divergence regularization term that constrains the updated policy $\pi_\theta$ toward the reference policy $\pi_{\text{ref}}$.

This formulation enables GRPO to optimize the actor model over multiple rollouts while stabilizing training and ensuring that only LLM-generated tokens contribute to the policy gradient.

\subsection{Single-model ReAct Instantiation}

Our method is instantiated under a ReAct-style single-model framework, where a single language model parameterized by $\theta$ jointly performs reasoning, query generation, and answer synthesis.
All intermediate reasoning steps, search queries, and final answers are generated autoregressively by the same policy $\pi_\theta$, conditioned on the evolving interaction history.

This unified formulation allows GRPO to assign token-level credit consistently across reasoning and action tokens within a shared policy space, avoiding credit fragmentation across multiple modules.

\subsection{RL with Search Engine for Multi-hop Question Answering}

We adopt the RL with Search Engine paradigm for multi-hop QA.
Given a question $q$ and corpus $\mathcal{C}$, the agent issues queries and receives retrieved sets over $K$ hops, and a reader uses the collected evidence to produce the final answer:
\begin{equation}
\label{eq:rl-se}
\begin{aligned}
p(a \mid q,\mathcal{C})
&\approx
\sum_{r_{1:K},\,D_{1:K}}
\Bigg\{
\prod_{i=1}^{K}
\Big[
\pi_{\theta}\!\big(r_i \mid q, D_{<i}\big)
\\[-2pt]
&
\cdot\, p_{\mathrm{se}}\!\big(D_i \mid r_i,\mathcal{C}\big)
\Big]
\Bigg\}\,
p_{\phi}\!\big(a \mid q, D_{1:K}\big).
\end{aligned}
\end{equation}

\noindent\textbf{Training objective.}
\begin{equation}
\label{eq:rl-obj}
\begin{aligned}
\max_{\theta,\phi}\;\;
&\mathbb{E}_{(q,a^\star)\sim\mathcal{D}}\;
\mathbb{E}_{\tau\sim\pi_{\theta}(\cdot \mid q,\mathcal{C})}
\big[\,R(q,\tau,a^\star;\phi)\,\big]
\\[-2pt]
&\;-\;\beta\,\mathrm{D_{KL}}\!\left[
\pi_{\theta}(\cdot \mid q,\mathcal{C})
\;\Big\|\;
\pi_{\mathrm{ref}}(\cdot \mid q,\mathcal{C})
\right].
\end{aligned}
\end{equation}

\noindent Here $p_{\mathrm{se}}(\cdot \mid r_i,\mathcal{C})$ denotes the search-engine return distribution for query $r_i$, and $p_{\phi}(a \mid q, D_{1:K})$ denotes the reader’s answer distribution given the gathered evidence. The reward $R(\cdot)$ is method-specific, and the KL term is optionally applied to stabilize policy updates.

\section{Golden Trajectory Generation with Global Planning}
\label{appendix:D}

To provide high-quality supervision for learning planning-aware reasoning policies, we construct \emph{golden reasoning trajectories} that explicitly encode both global task decomposition and step-by-step execution. These trajectories serve as expert demonstrations for evaluating and optimizing model rollouts, ensuring alignment between high-level planning decisions and local reasoning actions. Throughout the entire trajectory construction process, GPT-4o is used as the underlying model.

The trajectory construction process is performed offline and consists of three main stages.

\subsection{Global Planning}

Given an input multi-hop question, a fixed teacher model, operating in an instruction-following mode with frozen parameters, first performs global task analysis to derive an explicit execution plan.

Specifically, the teacher model conducts a semantic analysis of the input query to identify its logical structure, constraints, and underlying reasoning type (e.g., bridge, comparison, or compositional reasoning). Based on this analysis, the original question is systematically decomposed into a sequence of fine-grained subqueries, each formulated as a natural-language question that can be independently resolved via retrieval and reasoning.

Each subquery is associated with a placeholder identifier that represents its expected intermediate result. These placeholders are subsequently referenced by downstream subqueries to explicitly capture information dependencies. The resulting subqueries and their dependency relations are organized into a structured representation, such as a task dependency graph, which defines a partial order over subgoals and serves as a global execution plan.

The entire plan is generated in a single forward pass of the teacher model using a predefined planning prompt, and remains fixed throughout subsequent stages, ensuring consistency and reproducibility.

\subsection{Subgoal Execution}

Given the global plan, the teacher model resolves each subtask in topological order according to the dependency structure.

For each subquery, the model performs localized reasoning, including evidence retrieval, information synthesis, and intermediate answer generation. Retrieval is conducted using the same dense retriever employed during training and inference, without access to oracle documents, to ensure that the generated trajectories reflect realistic reasoning conditions.

Once a subtask is completed, its generated answer is substituted into the corresponding placeholder and propagated to all dependent subqueries. This explicit result propagation mechanism ensures that later reasoning steps operate on an updated context that incorporates previously resolved information, thereby maintaining global coherence.

Throughout this process, retrieval queries are conditioned on both the original question and the current intermediate context, enabling targeted evidence acquisition for each subgoal. The execution of each subtask is recorded as a sequence of state--action--result tuples, forming a fine-grained trace of the reasoning process.

\subsection{Final Answer Generation}

After all subtasks have been resolved, the intermediate answers are integrated according to the dependency graph to produce a globally consistent final answer.

Additionally, to ensure the reliability and internal consistency of the constructed golden trajectories, we apply a set of strict filtering criteria throughout the generation process. First, at the global planning stage, if the execution plan produced by the teacher model violates the predefined topological constraints of the task dependency graph—e.g., introducing cyclic dependencies or inconsistent subgoal ordering—the corresponding sample is discarded. Second, during subgoal execution, each subquery is allowed up to 15 retrieval attempts; if the teacher model fails to derive a valid intermediate answer within this budget, the entire trajectory is removed to avoid propagating incomplete or unstable reasoning traces. Finally, after final answer generation, trajectories whose predicted answers do not exactly match the ground-truth annotations are excluded. Together, these filtering rules prioritize structural correctness, execution feasibility, and answer-level accuracy, resulting in a high-quality set of expert demonstrations suitable for training and evaluation. Detailed prompt we use can be found in PER-PSE~\cite{liu2025beyond}.

\section{Experimental Details}
\label{appendix:E}
\subsection{Datasets}
\label{appendix:E.1}
% Our experiments utilize five multi-hop QA datasets: HotPotQA~\cite{yang-etal-2018-hotpotqa}, 2WikiMultiHopQA~\cite{ho-etal-2020-constructing-data}, MuSiQue~\cite{trivedi-etal-2022-musique}, Bamboogle~\cite{press-etal-2023-measuring-data} and WikiHop~\cite{welbl-etal-2018-constructing}. For each dataset, we construct the retrieval corpus following the standardized procedure established in prior ~\cite{jin2025flashrag}.
% The entire system is deployed as an online retrieval service using FastAPI for efficient query processing.

% The training set for \ourMethod{} is constructed from the train splits of HotpotQA 2WikiMultiHopQA and MuSiQue, totaling 8,394 examples. Moreover, we precompute golden plan and golden graph for all the data in the training datasets. 
% HotpotQA, 2WikiMultihopQA, MuSiQue and Bamboogle are sourced from the FlashRAG Datasets collection. Since the test or development sets of HotpotQA, 2WikiMultihopQA, and MuSiQue partially overlap with our training data, we perform deduplication. After processing, 5,187 samples are retained from the HotpotQA development set, 8,372 from the 2WikiMultihopQA test set, 444 from the MuSiQue development set. For Bamboogle, the entire datasets are directly used as test sets, including 125 samples, respectively. For WikiHop, the entire dev sets are used as test sets, including 5,129 samples.

% 12 25
Our experiments utilize five multi-hop QA datasets: HotPotQA~\cite{yang-etal-2018-hotpotqa}, 2WikiMultiHopQA~\cite{ho-etal-2020-constructing-data}, MuSiQue~\cite{trivedi-etal-2022-musique}, Bamboogle~\cite{press-etal-2023-measuring-data}, and WikiHop~\cite{welbl-etal-2018-constructing}. For each dataset, we construct the retrieval corpus following the standardized procedure established in prior work~\cite{jin2025flashrag}, and deploy the system as an online retrieval service using FastAPI for efficient query processing.

The training set for \ourMethod{} is built from the training splits of HotpotQA, 2WikiMultiHopQA, and MuSiQue, totaling 8,394 examples, with golden plans and golden graphs precomputed for all training data. HotpotQA, 2WikiMultiHopQA, MuSiQue, and Bamboogle are sourced from the FlashRAG Datasets collection. To prevent data leakage, we deduplicate instances where the test or development sets overlap with our training data. After processing, 5,187 samples are retained from the HotpotQA development set, 8,372 from the 2WikiMultiHopQA test set, 444 from the MuSiQue development set, 125 from Bamboogle, and 5,129 from WikiHop. We report results on the full test sets for Bamboogle and WikiHop, and on the sampled subsets for HotpotQA, 2WikiMultiHopQA, and MuSiQue. For datasets with non-public test sets (e.g., HotpotQA, 2WikiMultiHopQA, MuSiQue, WikiHop), we use their development sets as proxies for evaluation and report the corresponding results.

\subsection{Metrics}
\label{appendix:E.2}
Following the settings in DualRAG~\cite{cheng-2025-dualrag}, we adopt the following evaluation metrics:

\subsubsection{Exact Match (EM)}
The EM metric measures whether the predicted answer is exactly the same as the ground truth:
\begin{equation}
    \text{EM} = \frac{1}{N} \sum_{i=1}^{N} \mathbf{1}(\hat{y}_i = y_i)
\end{equation}

\subsubsection{F1 Score}
The F1 score is the harmonic mean of token-level precision and recall computed between the predicted and gold token sets, with values in the range \([0,1]\):

\begin{equation}
    F1(y, \hat{y}) = \frac{2 \times P \times R}{P + R}
\end{equation}

where Precision (\(P\)) and Recall (\(R\)) are defined as:
\begin{align}
P &= \frac{|\text{CommonTokens}(y, \hat{y})|}{|\text{Tokens}(\hat{y})|}, \\
R &= \frac{|\text{CommonTokens}(y, \hat{y})|}{|\text{Tokens}(y)|}
\end{align}
% \[
% P = \frac{|\text{CommonTokens}(y, \hat{y})|}{|\text{Tokens}(\hat{y})|}, \quad
% R = \frac{|\text{CommonTokens}(y, \hat{y})|}{|\text{Tokens}(y)|}
% \]

Precision and recall are also commonly used for evaluation in QA tasks.

\subsection{Training and retrieval Details}
\label{appendix:E.3}
Experiments use Qwen-2.5-3/7B (Base/Instruct) models. For retrieval, we build a \texttt{faiss-gpu} index over the 2018 Wikipedia snapshot and use \textsc{E5} embeddings as the dense retriever. Following Search-R1, we retrieve 3 passages for all methods.
Training merges HotpotQA, 2WikiMultiHopQA, and MuSiQue. Global-plan supervision for \ourMethod{} is derived via subgoal–dependency plans, verifying intermediate steps, yielding 8,394 supervised trajectories.

\subsection{Training Hyperparameters}
We trained our model on $4 \times$ A800 GPUs using a batch size of 256 and sequence length of 4,096 tokens, with FSDP and CPU offloading for memory optimization. The training employed GRPO with $\beta = 0.001$, $\epsilon = 0.2$, learning rate of $1 \times 10^{-6}$, and standard sampling parameters (temperature = 1.0, top-p = 1.0). We generated 5 responses per prompt over 200 steps, saving checkpoints every 10 steps, with an action budget of 4 and top-3 paragraph retrieval.

For the progressive reward scheduling described in the main text, we adopt a curriculum that gradually shifts emphasis from process-level supervision to final answer accuracy. The scheduling weight is defined as
\begin{equation}
    w_t = \frac{1}{1+\exp\left(\frac{t-0.9T}{10}\right)},
\end{equation}
where $t$ denotes the current training step, and $T$ is a hyperparameter that determines the transition point, set to $T=50$.
The process reward aggregates format, structural, semantic, and step-level components with coefficients $\alpha=0.1$ and $\lambda=\gamma=\delta=0.5$, respectively, while the answer reward is applied without additional scaling. These hyperparameters are fixed across all experiments unless otherwise specified.
}

\section{Performance–Reasoning Cost Trade-off Study}
\label{appendix:F}
\subsection{Inference-Time Performance--Cost Trade-off}

\begin{figure}[!h]
    \centering
    \includegraphics[width=0.8\linewidth]{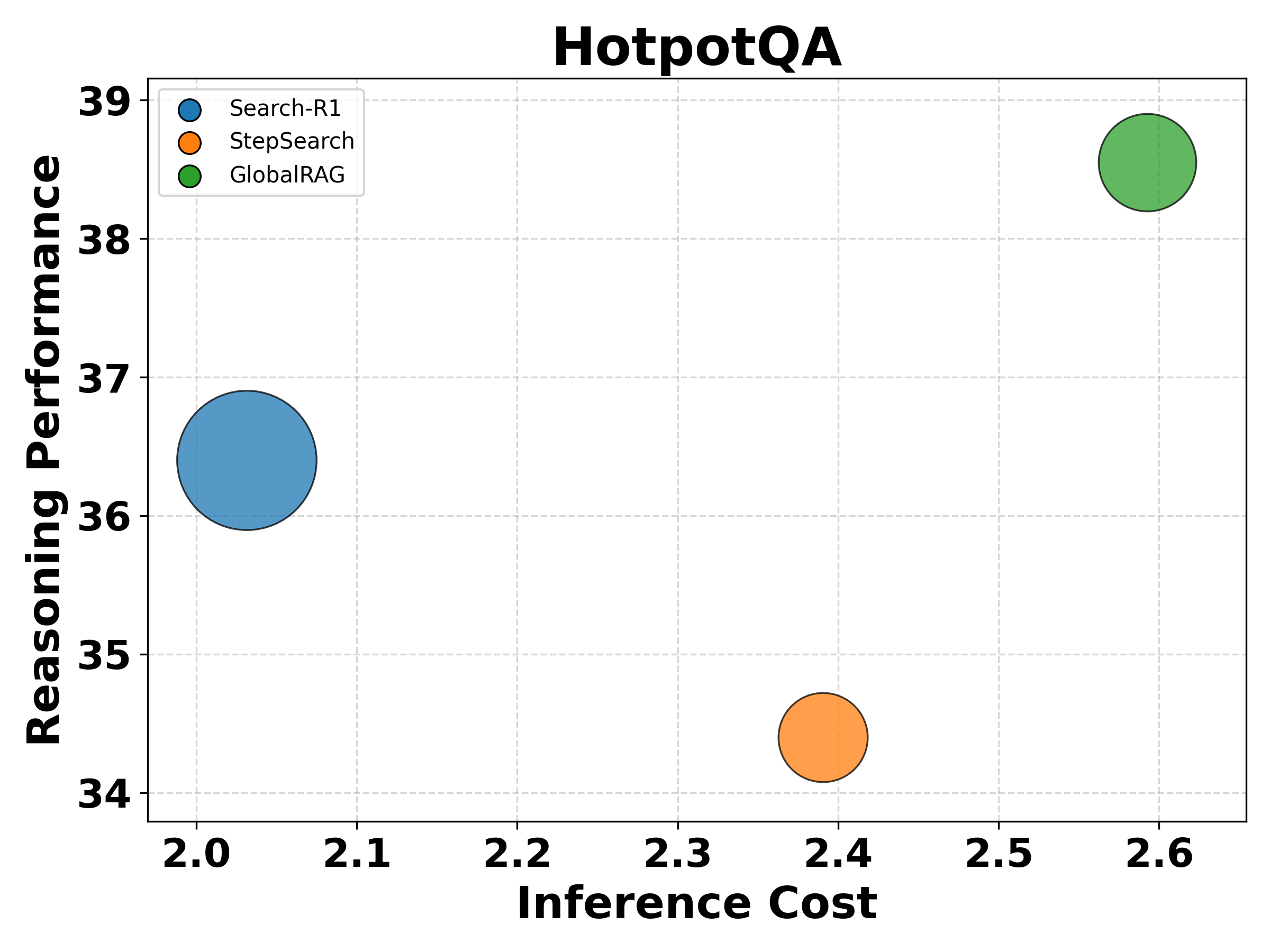}
    \caption{Inference-time performance versus token cost for three reinforcement learning based methods on \textbf{HotpotQA}.}
    \label{fig:hotpotqa}
\end{figure}

\begin{figure}[!h]
    \centering
    \includegraphics[width=0.8\linewidth]{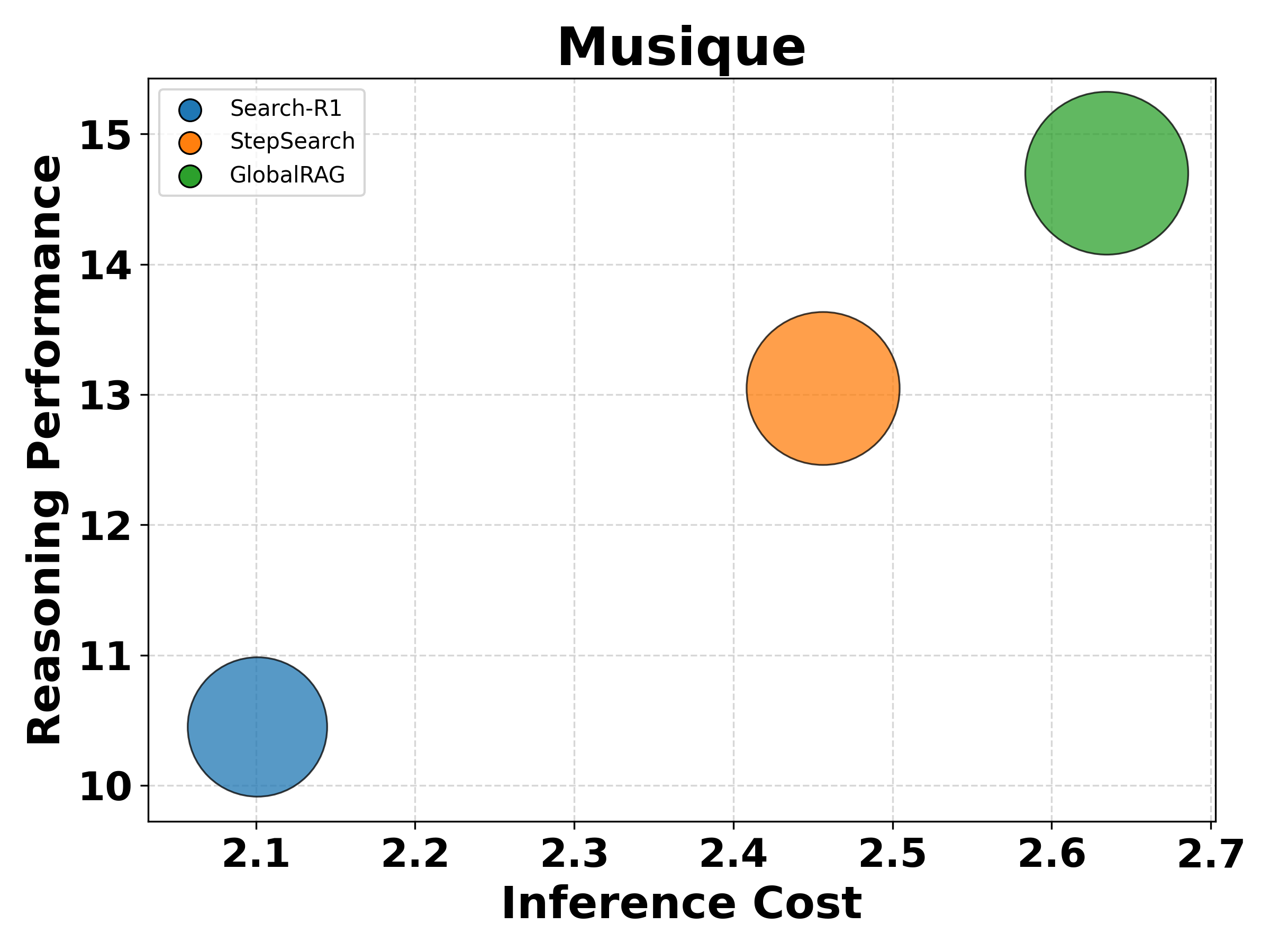}
    \caption{Inference-time performance versus token cost for three reinforcement learning based methods on \textbf{Musique}.}
    \label{fig:musique}
\end{figure}

\begin{figure}[!h]
    \centering
    \includegraphics[width=0.8\linewidth]{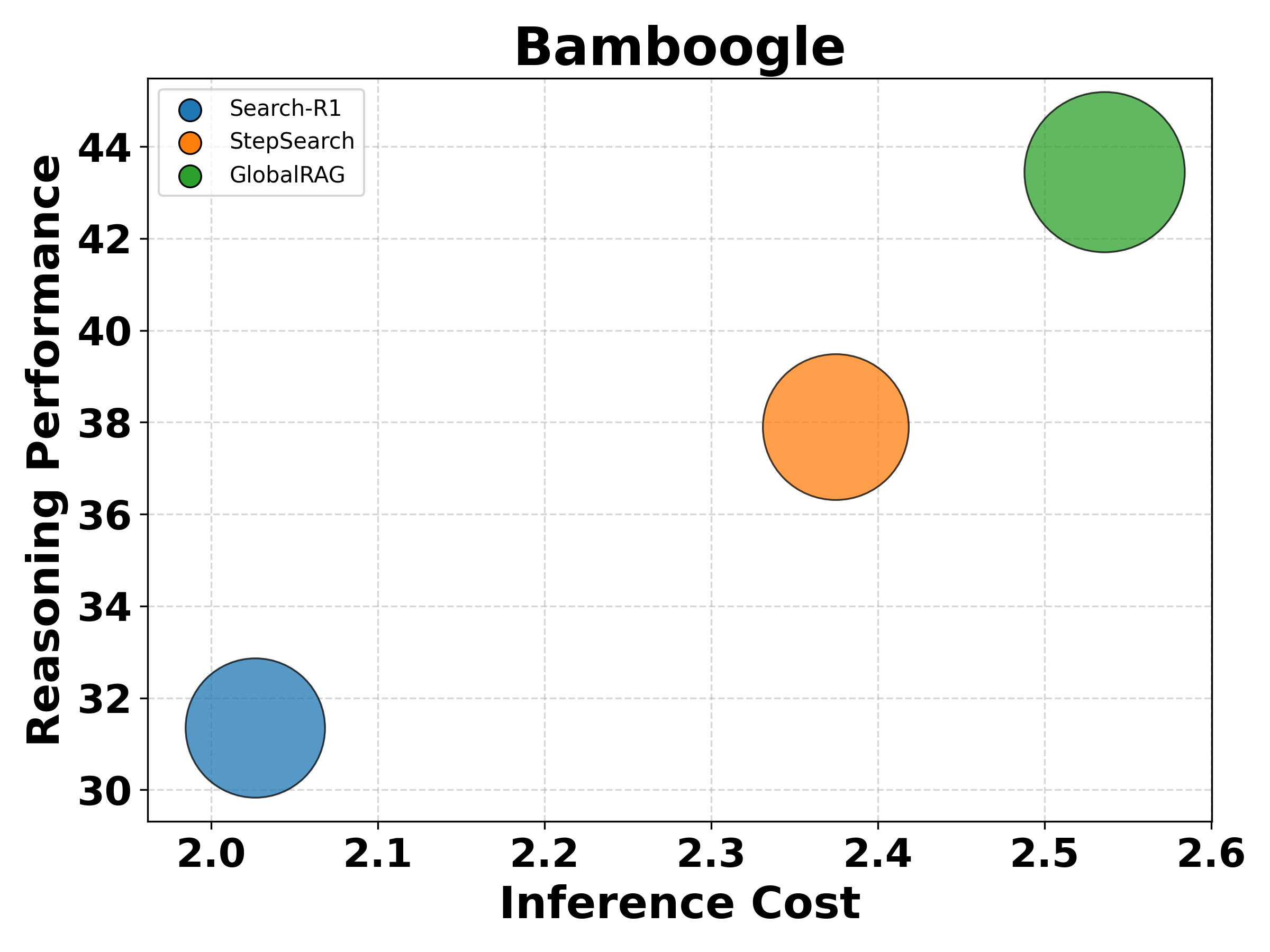}
    \caption{Inference-time performance versus token cost for three reinforcement learning based methods on \textbf{Bamboogle}.}
    \label{fig:bamboogle}
\end{figure}

\begin{figure}[!h]
    \centering
    \includegraphics[width=0.8\linewidth]{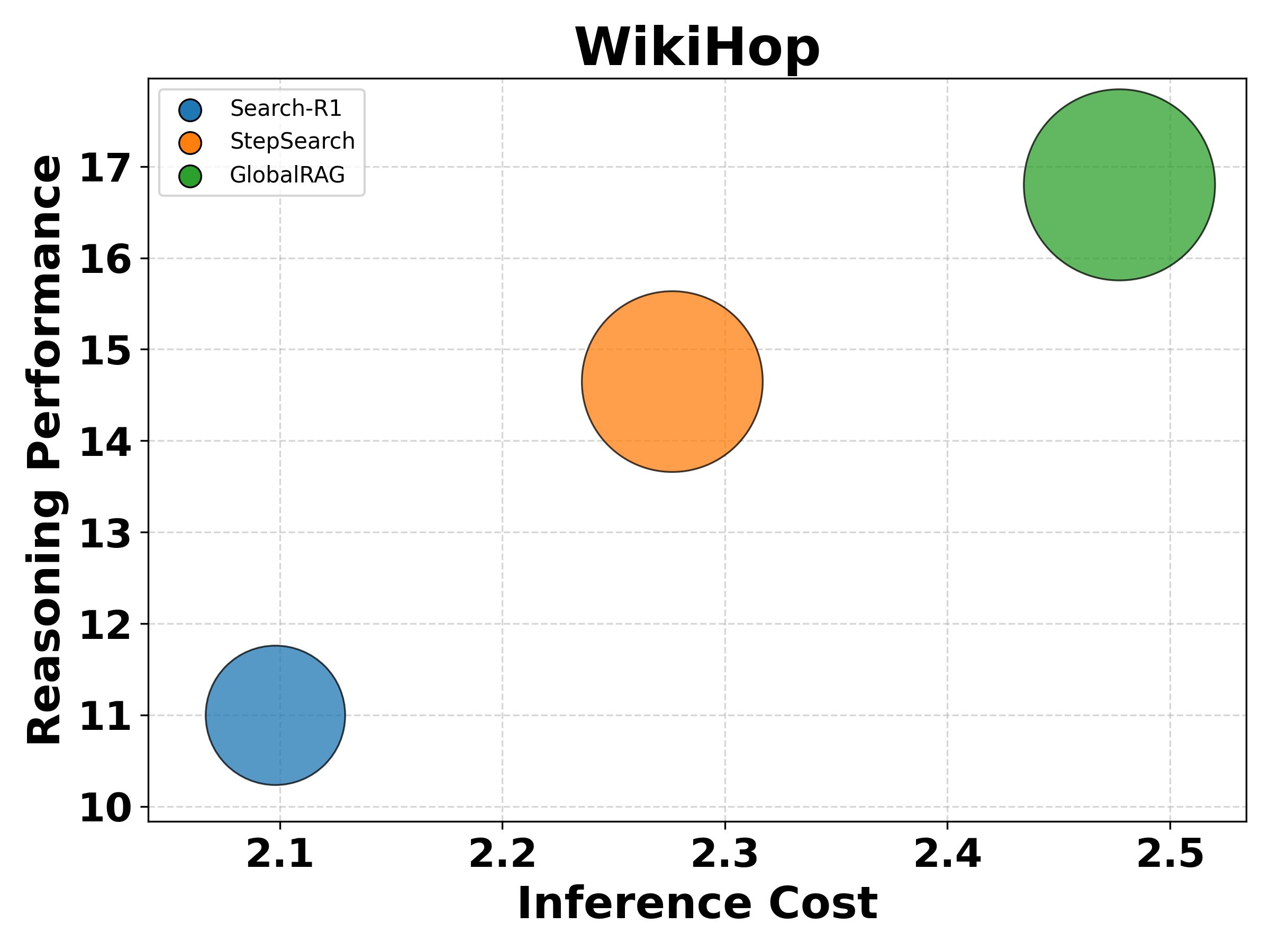}
    \caption{Inference-time performance versus token cost for three reinforcement learning based methods on \textbf{WikiHop}.}
    \label{fig:wikihop}
\end{figure}

To systematically evaluate the efficiency and scalability of \textbf{GlobalRAG}, we conduct a comparative study against two competitive baselines: \textbf{Search-R1} and \textbf{StepSearch}. All methods are implemented using the \textsc{Qwen2.5-3B-Instruct} model as the backbone, ensuring a fair assessment of reasoning strategies independent of model scale. We evaluate the methods across five representative multi-hop reasoning datasets: \textsc{HotpotQA}, \textsc{Bamboogle}, \textsc{Musique}, \textsc{2WikiMultihopQA}, and \textsc{WikiHop}. Reasoning performance (y-axis) is measured as the average of \textit{Exact Match (EM)} and \textit{F1} scores. Inference cost (x-axis) is represented by the logarithm of total token usage per query, $\log_{10}(\text{Tokens})$. Bubble size encodes the efficiency ratio, defined as:
\begin{equation}
    \frac{\text{Average(EM, F1)}}{\log_{10}(\text{Tokens})}
\end{equation}
where larger bubbles indicate higher performance per unit of token expenditure.

The figures here, alongside Figure~\ref{fig:2wikimultihopqa} in the main text, illustrate the trade-off between reasoning accuracy and computational cost across all datasets. Several key observations emerge. First, \textbf{GlobalRAG} (green bubbles) consistently occupies the top-right quadrant, demonstrating the highest reasoning performance. For instance, as shown in Figure~\ref{fig:2wikimultihopqa} for \textsc{2WikiMultihopQA}, GlobalRAG achieves a score above 45, whereas the baselines remain below 36, highlighting the effectiveness of global evidence integration and reasoning path optimization for complex multi-hop queries. Second, although GlobalRAG requires more tokens, its efficiency ratio remains comparable to or exceeds that of the baselines in most datasets, notably \textsc{Bamboogle} and \textsc{WikiHop}, suggesting that performance gains arise from high-quality reasoning rather than brute-force token usage. Finally, the baselines show distinct limitations: \textbf{Search-R1} (blue) maintains the lowest inference cost but suffers from a substantial performance gap, often plateauing due to limited capability in capturing inter-document dependencies. \textbf{StepSearch} (orange) lies between Search-R1 and GlobalRAG in cost, yet its performance improvements are not always proportional—for example, on \textsc{HotpotQA}, it requires more tokens but performs slightly worse than Search-R1.

Overall, this trade-off study demonstrates that \textbf{GlobalRAG} establishes a new Pareto frontier among retrieval-augmented generation systems. By strategically leveraging additional tokens for global context integration, it overcomes performance limitations inherent in step-wise or iterative search methods, offering the most robust scaling behavior for 3B-parameter models when reasoning accuracy is paramount.

\subsection{Training Efficiency under Reinforcement Learning}

\begin{table}[htbp]
\centering
\caption{Training Efficiency Comparison under Reinforcement Learning Framework}
\label{tab:training_efficiency}
\resizebox{\columnwidth}{!}{
\begin{tabular}{lcccc}
\toprule
Method & RL-based & Batch Size & Training Steps & Training Data Size \\
\midrule
Search-R1  & \checkmark & 512 & 500 & 170K \\
StepSearch & \checkmark & 256 & 500 & 19K  \\
GlobalRAG  & \checkmark & 256 & \textbf{200} & \textbf{8K}  \\
\bottomrule
\end{tabular}
}
\end{table}

Beyond inference-time efficiency, we further examine the training efficiency of the three methods under a unified reinforcement learning framework, as summarized in Table~\ref{tab:training_efficiency}. To ensure a fair comparison, all methods adopt the same batch size of 256. \textbf{Search-R1} is trained for 500 optimization steps with approximately 170K training samples, while \textbf{StepSearch} follows the same number of training steps but relies on a smaller dataset of 19K samples. In contrast, \textbf{GlobalRAG} converges substantially faster, requiring only 200 training steps and merely 8K training samples.

This comparison reveals a clear efficiency advantage of \textbf{GlobalRAG} during training. Despite significantly lower data and optimization budgets, GlobalRAG achieves superior or comparable reasoning performance, indicating more effective reward utilization and policy learning. When considered jointly with the inference-time trade-off illustrated in Figures~\ref{fig:hotpotqa}--\ref{fig:wikihop}, these results demonstrate that GlobalRAG consistently establishes a more favorable efficiency frontier than Search-R1 and StepSearch across both training and inference stages.

\subsection{Reward Computation Overhead Analysis}

\begin{table}[htbp]
\centering
\caption{Per-step Wall-clock Time Breakdown under Different Reward Designs (Qwen2.5-7B-Instruct)}
\label{tab:step_time_breakdown}
\resizebox{\columnwidth}{!}{
\begin{tabular}{lcccc}
\toprule
\multirow{2}{*}{Component} 
& \multicolumn{2}{c}{Search-R1 Reward} 
& \multicolumn{2}{c}{GlobalRAG Reward} \\
\cmidrule(lr){2-3} \cmidrule(lr){4-5}
& Time (s) & Ratio (\%) & Time (s) & Ratio (\%) \\
\midrule
Rollout Generation 
& 186.82 & 50.69 & 289.89 & 42.30 \\
Reward Computation 
& 1.07 & 0.29 & 112.12 & 16.36 \\
Policy Update 
& 94.28 & 25.58 & 145.24 & 21.19 \\
\midrule
Total per RL Step 
& 368.55 & 100.0 & 685.40 & 100.0 \\
\bottomrule
\end{tabular}
}
\end{table}

\begin{table}[htbp]
\centering
\caption{Per-token Normalized Computation Cost per RL Step}
\label{tab:per_token_cost}
\resizebox{\columnwidth}{!}{
\begin{tabular}{lccc}
\toprule
Reward Design 
& Rollout Time / Token (ms) 
& Reward Time / Token (ms) 
& Update Time / Token (ms) \\
\midrule
Search-R1 Reward 
& 23.61 & 0.12 & 10.54 \\
GlobalRAG Reward 
& 28.67 & 7.00 & 9.06 \\
\bottomrule
\end{tabular}
}
\end{table}

To better contextualize the training efficiency gains reported in the previous subsection, we further analyze the wall-clock overhead introduced by different reward designs. Tables~\ref{tab:step_time_breakdown} and~\ref{tab:per_token_cost} report a fine-grained runtime decomposition under controlled experimental settings: \textbf{Search-R1} and \textbf{GlobalRAG} share the same \textsc{Qwen2.5-7B-Instruct} backbone, identical reinforcement learning pipeline, and the same GlobalRAG-based training dataset. The sole difference lies in the reward formulation.

As shown in Table~\ref{tab:step_time_breakdown}, rollout generation dominates per-step runtime for both reward designs, accounting for over 40\% of total wall-clock time. Although GlobalRAG employs a more structured and computation-intensive reward, reward evaluation remains a secondary contributor compared to rollout sampling and policy optimization. Specifically, reward computation under GlobalRAG constitutes 16.36\% of the per-step cost, while rollout generation and policy updates together still account for more than 60\% of total runtime.

The per-token normalized analysis in Table~\ref{tab:per_token_cost} further clarifies this behavior. GlobalRAG incurs a higher reward computation cost per token (7.00 ms), which can be attributed to graph-aware matching and semantic similarity assessment. However, this overhead is amortized by the substantially larger cost of rollout generation. Notably, rollout and policy update costs per token remain comparable across reward designs, indicating that planning-aware rewards do not alter the fundamental computational profile of reinforcement learning optimization.

Taken together, these results demonstrate that the superior training efficiency of GlobalRAG, as shown in Table~\ref{tab:training_efficiency}, does not stem from a cheaper reward signal. Instead, GlobalRAG achieves faster convergence and lower data requirements despite a moderately higher reward computation cost. This confirms that its efficiency gains arise from more informative and effective supervision, rather than reduced computational complexity, reinforcing the practical scalability of GlobalRAG under standard reinforcement learning regimes.

\section{Quantitative Analysis of Plan Decomposition Diversity}
\label{appendix:G}

To further examine whether GlobalRAG relies on a single predefined reasoning trajectory or is able to generate diverse plan decompositions, we conduct a quantitative analysis on the diversity of planning behaviors at inference time. This study is performed on the \textsc{Bamboogle} dataset using a \textsc{Qwen2.5-3B-Instruct} model fine-tuned under the GlobalRAG framework.

For each question, we sample five independent responses with identical decoding configurations. In total, this results in 625 generated answers for 125 questions. Among these generations, 254 answers are judged to be correct according to the dataset evaluation protocol. To assess plan-level diversity, we further group correct answers by their corresponding plan decompositions, where a plan is defined as the sequence of intermediate reasoning steps and sub-question structures produced by the model. Duplicate plans for the same question are collapsed into a single instance.

After deduplication, we obtain 144 distinct correct plan decompositions. This indicates that a substantial portion of correct answers arises from multiple, non-identical reasoning plans rather than repeated execution of a single trajectory. Quantitatively, over 56\% of correct generations correspond to distinct plan decompositions, suggesting that the learned policy is capable of exploring alternative reasoning structures that still lead to valid answers.

These results provide empirical evidence that GlobalRAG is not strictly constrained to reproducing a single golden plan during inference. Instead, the model demonstrates a degree of flexibility in plan construction, generating multiple viable decompositions for the same question. While repeated plans remain common—reflecting stable high-reward reasoning patterns—the presence of a large number of distinct correct plans indicates that the planning policy learned by GlobalRAG is not limited to a fixed, deterministic trajectory.

We emphasize that this analysis is intended to establish the existence of plan-level diversity rather than to claim optimal or exhaustive exploration of the plan space. Nonetheless, the observed diversity supports the view that GlobalRAG can move beyond rigid adherence to predefined reasoning paths and instead learn a more flexible planning behavior under reinforcement learning.

\section{Number of Retrieved Passages Study}
\label{appendix:H}
{
\begin{figure}[h]
    \centering
    \includegraphics[width=0.95\linewidth]{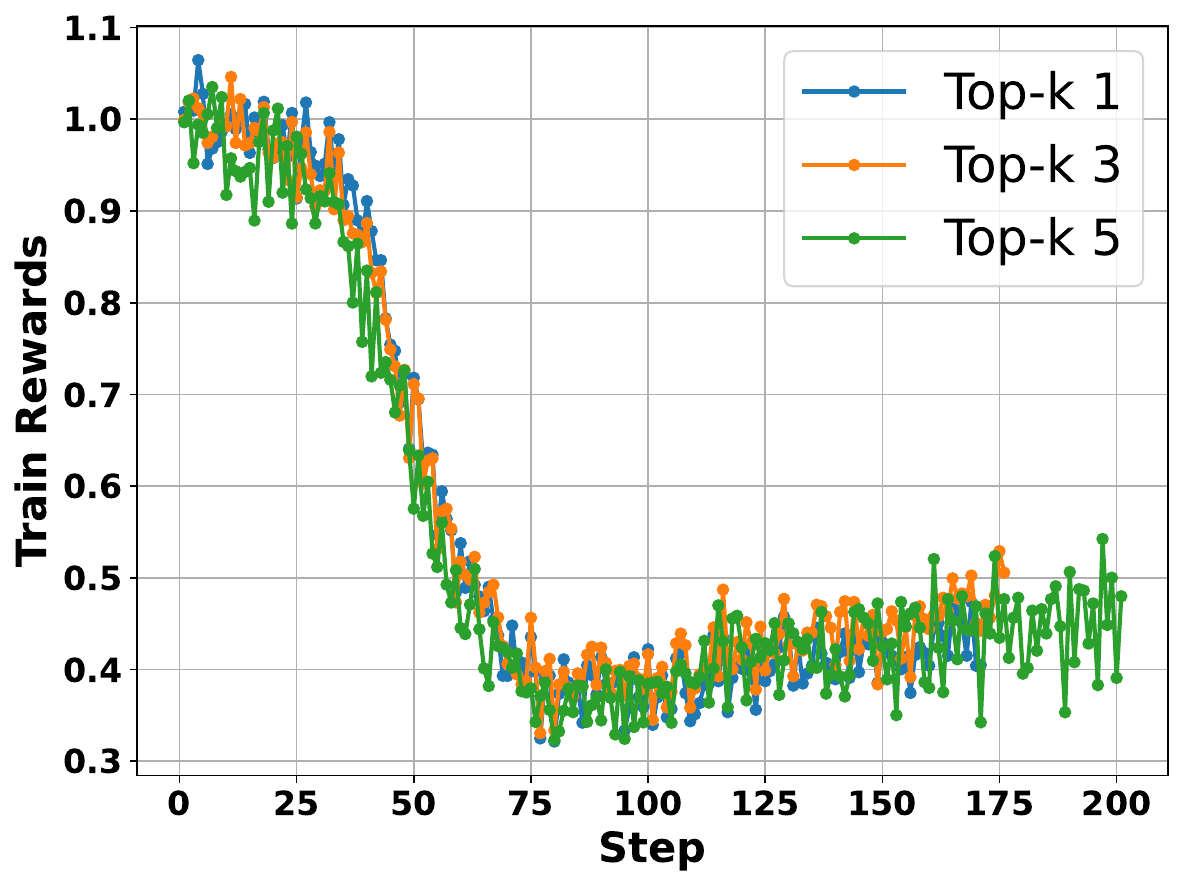}
    \caption{Comparison of training rewards under different numbers of retrieved passages ($k$). The figure reports the evolution of training rewards over steps when varying the top-$k$ retrieved passages ($k \in \{1, 3, 5\}$). (LLM: Qwen2.5-3B-instruct, RL: GRPO)}
    \label{fig:appendix-retrieval}
\end{figure}

\begin{table}[h]
\centering
\small
\setlength{\tabcolsep}{5pt}
\resizebox{\columnwidth}{!}{%
\begin{tabular}{lcccccccccc}
\toprule
\multirow{2}{*}{Methods} 
& \multicolumn{2}{c}{HotpotQA} 
& \multicolumn{2}{c}{2Wiki} 
& \multicolumn{2}{c}{Musique} 
& \multicolumn{2}{c}{Bamboogle} 
& \multicolumn{2}{c}{WikiHop} \\
\cmidrule(lr){2-3}
\cmidrule(lr){4-5}
\cmidrule(lr){6-7}
\cmidrule(lr){8-9}
\cmidrule(lr){10-11}
& EM & F1 & EM & F1 & EM & F1 & EM & F1 & EM & F1 \\
\midrule
$k=1$ & 28.9 & 40.1 & 38.4 & 43.7 & 8.0 & 14.2 & 30.4 & 37.7 & 12.6 & 20.6 \\
$k=3$ & 32.9 & 44.2 & 42.3 & 47.8 & 10.8 & 18.6 & 37.6 & 49.3 & 12.9 & 20.7 \\
$k=5$ & 32.4 & 44.0 & 43.2 & 48.6 & 9.9 & 16.9 & 39.2 & 52.8 & 13.4 & 20.9 \\
\bottomrule
\end{tabular}
}
\caption{Performance of Qwen2.5-3B-Instruct under different retrieval depths ($k$).}
\label{tab:retrieval}
\end{table}

We conduct a systematic analysis of GlobalRAG under different retrieval depths $k$ on both In-Domain and Out-of-Domain multi-hop question answering benchmarks. The results are summarized in Table \ref{tab:retrieval}. Overall, model performance exhibits a consistent upward trend as the retrieval depth increases.

On typical In-Domain multi-hop reasoning datasets such as HotpotQA and 2WikiMultihopQA, increasing the retrieval depth from $k=1$ to $k=3$ leads to substantial improvements in both EM and F1. For instance, on HotpotQA, EM improves from 28.9 to 32.9, while F1 increases from 40.1 to 44.2. When the retrieval depth is further expanded to $k=5$, performance remains at a relatively high level but the gains become marginal, suggesting that a moderate retrieval depth is sufficient to capture most of the relevant evidence required for multi-hop reasoning.

In contrast, on the more challenging Musique dataset, increasing $k$ results in only limited performance gains. This behavior may be attributed to the dataset’s higher intrinsic difficulty or to a more dispersed distribution of supporting evidence, where simply retrieving additional passages does not substantially alleviate the reasoning burden.

For Out-of-Domain evaluation, the model is generally less sensitive to retrieval depth than in In-Domain settings, yet still demonstrates consistent improvements as $k$ increases. On datasets such as Bamboogle and WikiHop, both EM and F1 improve steadily when $k$ is increased from 1 to 3 and further to 5, indicating that expanding the retrieval context helps compensate for domain shifts by providing additional external knowledge. Notably, on Bamboogle, F1 rises from 37.7 to 52.8, highlighting that deeper retrieval can partially mitigate information sparsity caused by distributional mismatch.

It is also worth noting that the performance gains exhibit a clear diminishing-returns pattern. Across both In-Domain and Out-of-Domain benchmarks, a retrieval depth of $k=3$ is often sufficient to cover the majority of useful evidence, while further increasing $k$ yields relatively limited benefits. This phenomenon may be attributed to the global planning mechanism employed by GlobalRAG: guided by reinforcement learning signals, the model learns to organize subgoals and utilize retrieved evidence more effectively under a smaller retrieval budget, thereby reducing reliance on excessive or redundant documents.

Overall, these results demonstrate that GlobalRAG achieves stable and consistent improvements across different retrieval depths. The study further confirms that, for multi-hop question answering, appropriately increasing the retrieval depth can enhance reasoning performance, while the reinforcement learning–driven global planning mechanism improves retrieval efficiency and achieves a favorable balance between performance and retrieval cost.

}

\section{Group Size Study in GRPO}
\label{appendix:I}
{
\begin{figure}[h]
    \centering
    \includegraphics[width=0.95\linewidth]{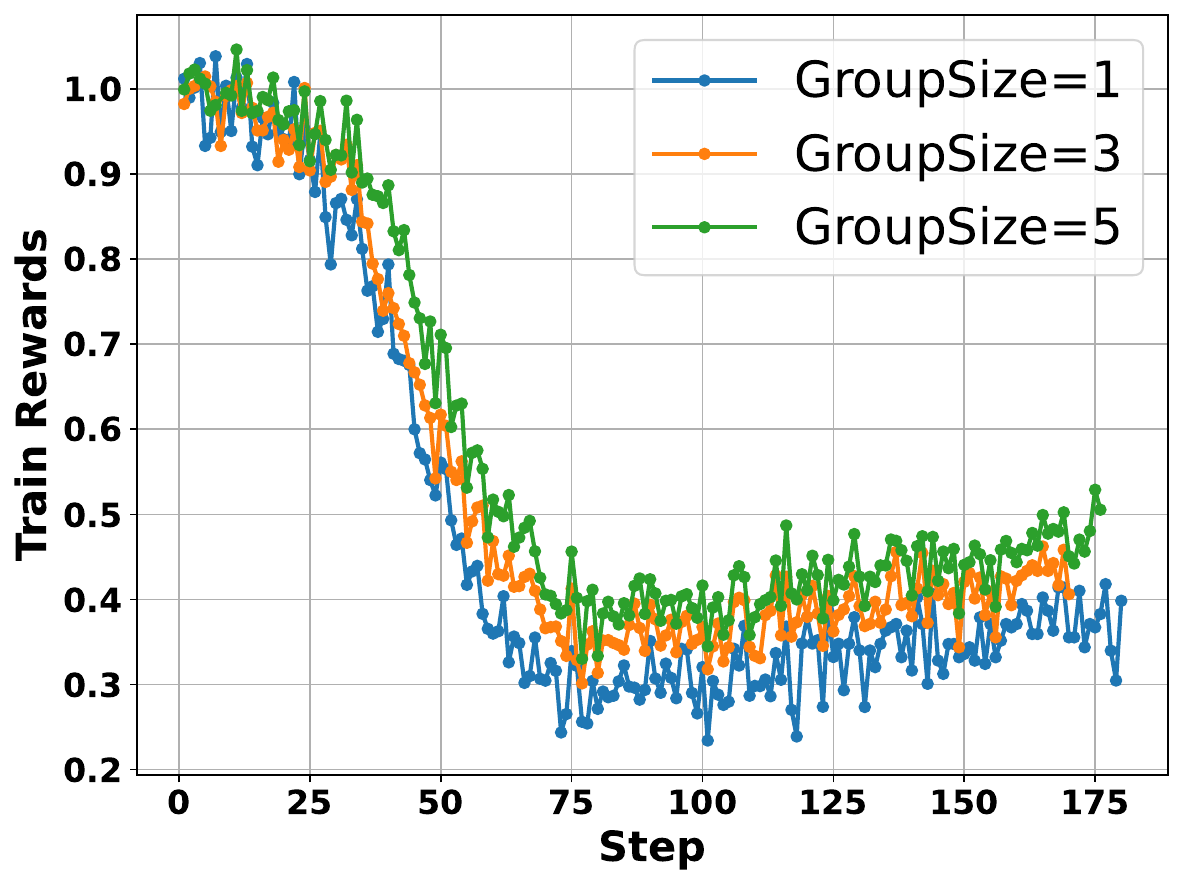}
    \caption{Comparison of GRPO training dynamics under different group sizes. (LLM: Qwen2.5-3B-instruct, RL: GRPO)}
    \label{fig:appendix-groupsize}
\end{figure}

\begin{table}[h]
\centering
\small
\setlength{\tabcolsep}{5pt}
\resizebox{\columnwidth}{!}{%
\begin{tabular}{lcccccccccc}
\toprule
\multirow{2}{*}{Methods} 
& \multicolumn{2}{c}{HotpotQA} 
& \multicolumn{2}{c}{2Wiki} 
& \multicolumn{2}{c}{Musique} 
& \multicolumn{2}{c}{Bamboogle} 
& \multicolumn{2}{c}{WikiHop} \\
\cmidrule(lr){2-3}
\cmidrule(lr){4-5}
\cmidrule(lr){6-7}
\cmidrule(lr){8-9}
\cmidrule(lr){10-11}
& EM & F1 & EM & F1 & EM & F1 & EM & F1 & EM & F1 \\
\midrule
$n=1$ & 27.2 & 38.0 & 37.3 & 42.4 & 8.9 & 16.5 & 31.2 & 42.4 & 8.5 & 16.1 \\
$n=3$ & 30.3 & 42.1 & 37.3 & 43.4 & 9.3 & 16.5 & 32.8 & 45.6 & 10.7 & 18.4 \\
$n=5$ & 32.9 & 44.2 & 42.3 & 47.8 & 10.8 & 18.6 & 37.6 & 49.3 & 12.9 & 20.7 \\
\bottomrule
\end{tabular}
}
\caption{Performance of Qwen2.5-3B-Instruct with GRPO under different group sizes ($n$).}
\label{tab:groupsize}
\end{table}

Table \ref{tab:groupsize} reports the performance of GlobalRAG trained with GRPO under different group size (n) settings across multiple multi-hop question answering benchmarks. As n increases from 1 to 5, we observe consistent improvements in both EM and F1 on all datasets.

From an algorithmic perspective, this trend can be attributed to the statistical properties of advantage estimation in GRPO. Unlike policy optimization methods that rely on an explicit value function, GRPO estimates advantages using relative rewards within a sampled group, where the group-wise average reward serves as an implicit baseline. When the group size is small, this baseline estimate is more susceptible to noise from individual samples, resulting in higher variance in the advantage signal and less stable policy updates. Increasing the group size yields a more reliable estimate of the expected reward under the current policy, thereby providing a more accurate reference point for computing relative advantages.

This effect is particularly pronounced on reasoning-intensive benchmarks such as HotpotQA, 2Wiki, and WikiHop. For instance, on HotpotQA, increasing n from 1 to 5 leads to gains of 5.7 EM and 6.2 F1 points, indicating that a more stable relative advantage signal helps the model better discriminate among retrieval–reasoning trajectories of varying quality. On smaller or noisier datasets such as Musique and Bamboogle, the improvements are more modest but remain consistent, suggesting that larger group sizes primarily contribute to variance reduction during training rather than exploiting dataset-specific artifacts.

At the same time, the gains exhibit diminishing returns as the group size increases, highlighting a trade-off between the accuracy of baseline estimation and computational cost. Based on this observation, we adopt n=5 as the default group size in the remaining experiments unless otherwise specified.
}

\section{Reward Weight Sensitivity Study}
\label{appendix:J}
{
\begin{figure}[!h]
    \centering
    \includegraphics[width=\columnwidth]{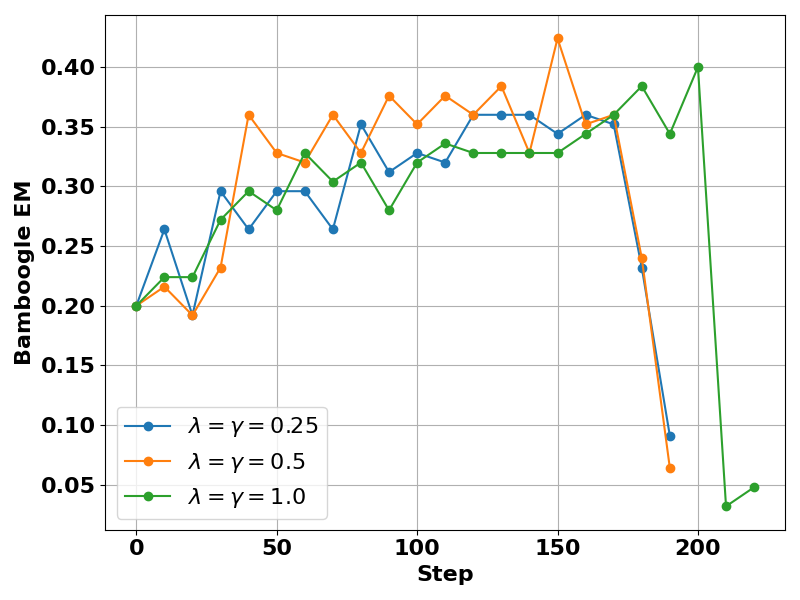}
    \caption{Validation EM on the Bamboogle dataset under different global-plan reward weight coefficients ($\lambda=\gamma$), evaluated every ten training steps. The plan reward jointly incorporates structural and semantic constraints.}
    \label{fig:reward_sensitivity}
\end{figure}

\begin{table}[t]
\centering
\small
\setlength{\tabcolsep}{4.5pt}
\resizebox{\columnwidth}{!}{%
\begin{tabular}{lcccccccccc}
\toprule
\multirow{2}{*}{Weight} 
& \multicolumn{2}{c}{HotpotQA} 
& \multicolumn{2}{c}{2Wiki} 
& \multicolumn{2}{c}{Musique}
& \multicolumn{2}{c}{Bamboogle} 
& \multicolumn{2}{c}{WikiHop} \\
\cmidrule(lr){2-3} \cmidrule(lr){4-5} \cmidrule(lr){6-7}
\cmidrule(lr){8-9} \cmidrule(lr){10-11}
& EM & F1 & EM & F1 & EM & F1 & EM & F1 & EM & F1 \\
\midrule
$\lambda=\gamma=0.25$ 
& 31.4 & 42.2 
& 40.9 & 46.9
& 10.6 & 18.2
& 36.0 & 48.7
& 11.7 & 19.3
\\
$\lambda=\gamma=0.5$
& \textbf{32.9} & \textbf{44.2} 
& \textbf{42.3} & \textbf{47.8} 
& 10.8 & \textbf{18.6} 
& \textbf{37.6} & 49.3 
& 12.9 & 20.7 \\
$\lambda=\gamma=1.0$
& 32.2 & 43.8 
& 41.4 & 47.2 
& \textbf{11.1} & 18.0 
& \textbf{37.6} & \textbf{49.6} 
& \textbf{13.1} & \textbf{21.5} \\
\bottomrule
\end{tabular}
}
\caption{Performance comparison under different global-plan reward weight coefficients across five QA benchmarks, evaluated using Qwen2.5-3B-Instruct.}
\label{tab:reward_sensitivity}
\end{table}
}

We investigate the effect of the global-plan reward weight by varying the coefficient $\lambda=\gamma \in \{0.25, 0.5, 1.0\}$, where the plan reward jointly accounts for structural validity and semantic alignment. Figure~\ref{fig:reward_sensitivity} reports the validation EM on Bamboogle evaluated every ten training steps. All settings exhibit steady improvements during the early stages of training, indicating that plan-level supervision provides effective guidance for multi-hop reasoning. Nevertheless, their later-stage behaviors differ noticeably. A smaller weight ($\lambda=\gamma=0.25$) results in smoother but relatively slower performance gains, suggesting that weak plan supervision limits the exploitation of global reasoning signals. In contrast, a larger weight ($\lambda=\gamma=1.0$) achieves competitive peak accuracy but shows clear instability toward the end of training, with abrupt performance degradation, implying that overly strong plan rewards may dominate optimization and reduce robustness.

The aggregated results across five benchmarks in Table~\ref{tab:reward_sensitivity} further support this observation. Overall, $\lambda=\gamma=0.5$ achieves the most balanced performance, delivering consistent improvements on HotpotQA, 2WikiMultihopQA, and Bamboogle, while remaining competitive on Musique and WikiHop. Although $\lambda=\gamma=1.0$ attains the best scores on some datasets, its advantages are less consistent and accompanied by unstable training dynamics. These results suggest that a moderate global-plan reward weight provides a favorable trade-off between effective plan guidance and stable optimization. Accordingly, we adopt $\lambda=\gamma=0.5$ as the default configuration in our main experiments.

\section{Transition Point Sensitivity Study}
\begin{figure}[h]
    \centering
    \includegraphics[width=0.95\linewidth]{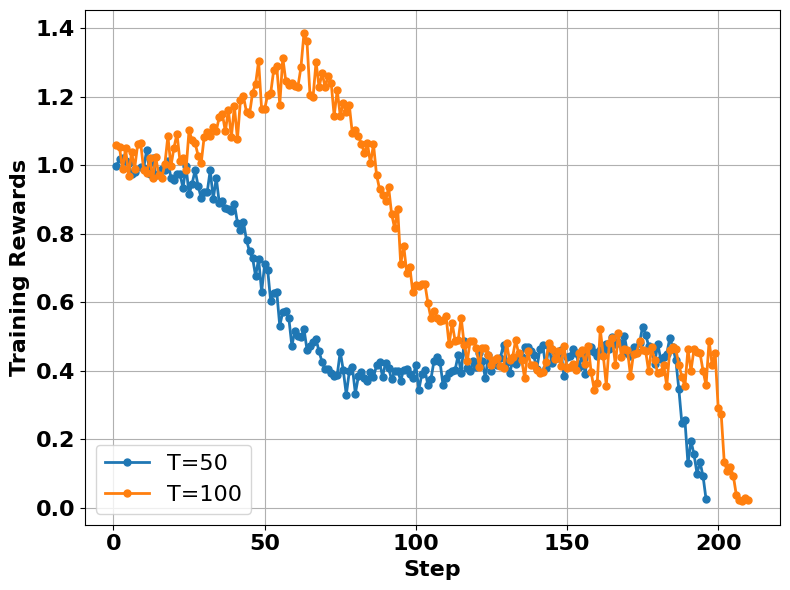}
    \caption{Training reward curves of the Qwen2.5-3B-Instruct model under different transition point settings (T=50 and T=100).}
    \label{fig:T-rewards}
\end{figure}

\begin{figure}[h]
    \centering
    \includegraphics[width=0.95\linewidth]{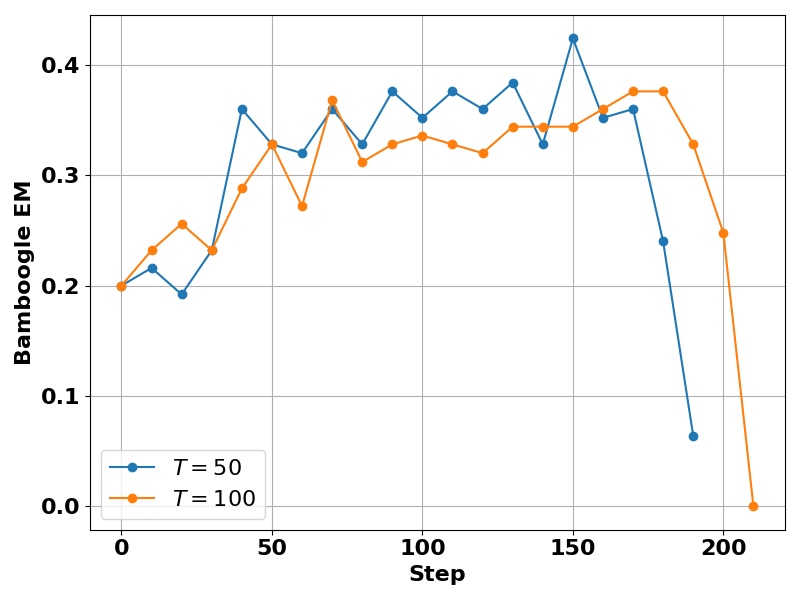}
    \caption{Performance comparison under transition point settings (T=50 and T=100) across five QA benchmarks, evaluated using Qwen2.5-3B-Instruct.}
    \label{fig:T-bamboogle}
\end{figure}

\label{appendix:K}
\begin{table}[h]
\centering
\small
\setlength{\tabcolsep}{4.5pt}
\resizebox{\columnwidth}{!}{%
\begin{tabular}{lcccccccccc}
\toprule
\multirow{2}{*}{Methods} 
& \multicolumn{2}{c}{HotpotQA} 
& \multicolumn{2}{c}{2Wiki} 
& \multicolumn{2}{c}{Musique}
& \multicolumn{2}{c}{Bamboogle} 
& \multicolumn{2}{c}{WikiHop} \\
\cmidrule(lr){2-3} \cmidrule(lr){4-5} \cmidrule(lr){6-7}
\cmidrule(lr){8-9} \cmidrule(lr){10-11}
& EM & F1 & EM & F1 & EM & F1 & EM & F1 & EM & F1 \\
\midrule
T=50
& 32.9 & 44.2 
& 42.3 & 47.8 
& 10.8 & 18.6 
& 37.6 & 49.3 
& 12.9 & 20.7 \\
T=100
&31.2  &43.1  
&40.6  &46.6      
&10.0  &17.0  
&39.2  &50.6      
&12.2  &19.8  \\
\bottomrule
\end{tabular}
}
\caption{Performance comparison of Qwen2.5-3B-Instruct under different Transition Point $T$ across five QA benchmarks.}
\label{tab:temperature_ablation}
\end{table}

\textbf{Training Dynamics:} As illustrated in Figure~\ref{fig:T-rewards}, the reward curve for $T=50$ exhibits a more rapid and stable ascent compared to $T=100$. When $T=100$, the model remains in the process-oriented learning phase for a longer duration, which, while helpful for initial planning, delays the optimization for final answer accuracy. In contrast, $T=50$ provides a more balanced transition, allowing the model to quickly capitalize on refined reasoning paths to improve final outcomes.

\textbf{Performance Comparison:} Table~\ref{tab:temperature_ablation} and Figure~\ref{fig:T-bamboogle} present the evaluation results. We observe that:
$T=50$ outperforms $T=100$ on four out of five benchmarks (HotpotQA, 2WikiMultiHop, Musique, and WikiHop). The improvements in EM and F1 scores suggest that an earlier shift towards outcome-based optimization helps the model better internalize the relationship between global planning and correct answer generation.

Based on the empirical evidence, we conclude that $T=50$ is the optimal transition point for GlobalRAG. It achieves the best trade-off between guiding the model's reasoning process and ensuring high-fidelity final answers. Setting the transition point too late (e.g., $T=100$) may lead to over-optimization of intermediate steps at the expense of final task performance. Consequently, $T=50$ is adopted as the default setting for all other experiments in this study.

\section{Detailed Results for Algorithmic Attribution Study}
\label{appendix:L}
{

%\begin{table}[!h]
%\centering
%\footnotesize
%\setlength{\tabcolsep}{3.8pt}
%\renewcommand{\arraystretch}{1.12}
%\resizebox{\columnwidth}{!}{%
%\begin{tabular}{lcccccccccc}
%\toprule
%\multirow{2}{*}{Methods} &
%\multicolumn{2}{c}{\textbf{HotpotQA}} &
%\multicolumn{2}{c}{\textbf{2Wiki}} &
%\multicolumn{2}{c}{\textbf{MuSiQue}} &
%\multicolumn{2}{c}{\textbf{Bamboogle}} &
%\multicolumn{2}{c}{\textbf{WikiHop}} \\
%\cmidrule(lr){2-3}\cmidrule(lr){4-5}\cmidrule(lr){6-7}\cmidrule(lr){8-9}\cmidrule(lr){10-11}
%& EM & F1 & EM & F1 & EM & F1 & EM & F1 & EM & F1 \\
%\midrule
%\textbf{Qwen2.5-3b-Instruct} \\
%\text{Search-R1} &30.1 &36.1 &36.5 &40.8 &8.3 &12.7 &32.0 &33.5 &10.6 &13.3 \\
%\text{GlobalRAG} &32.9 &44.2 &42.3 &47.8 &10.8 &18.6 &37.6 &49.3 &12.9 &20.7 \\

%\midrule
%\textbf{Qwen2.5-7b-Instruct} \\
%\text{Search-R1} &34.4 &45.6 &42.4 &47.6 &9.0 &16.7 &41.1 &49.3 &15.1 &21.9 \\

%\text{GlobalRAG} &38.2 &50.7 &47.8 &53.2 &14.9 &21.2 &48.8 &61.6 &19.0 &28.0 \\

%\midrule
%\textbf{Llama3.1-8b-Instruct} \\
%Search-R1 & 35.7 & 46.9 & 44.6 & 50.1 & 12.9 & 20.0 & 44.8 & 56.5 & 16.5 & 22.0 \\
%GlobalRAG & 36.9 & 48.6 & 46.6 & 51.9 & 17.0 & 25.7 & 49.6 & 61.9 & 17.8 & 25.1 \\

%\bottomrule
%\end{tabular}%
%}
%\caption{Algorithmic attribution (controlled data): %GlobalRAG vs. Search-R1.}
%\label{tb:algorithmic}
%\end{table}

This appendix provides the complete quantitative results for the algorithmic attribution study summarized in the main text. All experiments reported here are conducted under strictly controlled training data settings, ensuring that both Search-R1 and GlobalRAG are trained on identical datasets. Additional experimental details and analysis can be found in the corresponding section of the main paper.

Table~\ref{tb:algorithmic} reports EM and F1 scores across five multi-hop question answering benchmarks using three backbone models of different scales: Qwen2.5-3B-Instruct, Qwen2.5-7B-Instruct, and Llama3.1-8B-Instruct. These results complement the main-text discussion by presenting the full benchmark-level breakdown.

Under all model configurations, GlobalRAG consistently achieves higher performance than Search-R1 across datasets. The improvements are especially evident on benchmarks that require complex multi-hop reasoning and evidence aggregation, such as HotpotQA, MuSiQue, and Bamboogle. Notably, the performance gap generally increases with model scale, indicating that GlobalRAG is able to more effectively exploit larger backbone capacity.

On MuSiQue, which emphasizes long-horizon reasoning and structured knowledge integration, GlobalRAG demonstrates relatively larger gains compared to Search-R1 across all backbones. This observation further supports the main-text claim that the proposed algorithmic design better captures global reasoning structure under challenging conditions.

Overall, the detailed results reported in this appendix provide comprehensive empirical support for the conclusion drawn in the main text: when trained on identical data, the observed performance improvements of GlobalRAG over Search-R1 are attributable to algorithmic differences rather than data-related factors.
}

\section{Case Study}
\label{appendix:M}
To better understand the performance characteristics of our proposed method, we conduct a comparative analysis between \ourMethod{} and Search-R1 through case studies. Our analysis reveals that \ourMethod{} demonstrates significant advantages over the baseline approach in two critical aspects: enhanced retrieval precision through global planning and improved reasoning accuracy via structured sub-question decomposition.

Specifically, our global planning strategy enables more precise retrieval of relevant evidence. As shown in Table~\ref{tab:case_study1} and Table~\ref{tab:case_study2}, \ourMethod{} consistently retrieves the key information needed for complex queries, whereas Search-R1 often returns incomplete or less relevant content. Moreover, explicit \texttt{<subPlan>} steps and sub-answer validation rewards help filter distracting evidence and ensure the correctness of intermediate reasoning and sub-answers. This advantage is especially clear in Table~\ref{tab:case_study3} and Table~\ref{tab:case_study4}, where \ourMethod{} remains coherent and accurate even under misleading information, demonstrating improved reliability on complex QA tasks.

% Specifically, the global planning strategy employed by our method enables more precise information retrieval and acquisition of accurate relevant information. As demonstrated in Table~\ref{tab:case_study1} and Table~\ref{tab:case_study2}, \ourMethod{} successfully identifies and retrieves the most pertinent information needed for answering complex queries, whereas Search-R1 often retrieves incomplete or less relevant content. Furthermore, the explicit incorporation of \texttt{<subPlan>} steps for structured reasoning, combined with a reward mechanism for validating sub-question answers, allows our method to effectively filter out distracting information while ensuring the correctness of both intermediate reasoning processes and sub-answers. This advantage is particularly evident in Table~\ref{tab:case_study3} and Table~\ref{tab:case_study4}, where \ourMethod{} maintains reasoning coherence and accuracy despite the presence of potentially misleading information. These findings demonstrate that our method's architectural innovations contribute to enhanced accuracy and reliability in complex question-answering tasks.

\begin{table*}[t]
\centering
\resizebox{0.95\textwidth}{!}{
\begin{tabular}{p{0.95\textwidth}}
\toprule
\toprule
\textbf{Case 1:} Into what does the river for which Toongabbie Creek serves as the mouth flow? \\
\textbf{Ground Truth:} \textbf{Sydney Harbour} \\
\midrule

\textbf{\ourMethod:} \\
\colorbox{gray!15}{\parbox{\dimexpr\linewidth-2\fboxsep}
{\texttt{<think>} To answer this question, I need to break it down into sub-questions. First, I need to identify the river that Toongabbie Creek flows into. Then, I need to determine what that river flows into. I can proceed step-by-step to find the answer. \texttt{</think>}}}

\colorbox{orange!20}{\parbox{\dimexpr\linewidth-2\fboxsep}{\texttt{<plan>} \{ "Q1": ["For what river does Toongabbie Creek serve as the mouth?", "\#1"], "Q2": ["Into what does \#1 flow?", "\#2"] \} \texttt{</plan>}}}

\colorbox{cyan!15}{\parbox{\dimexpr\linewidth-2\fboxsep}{
\texttt{<subPlan>}\\
\hspace*{1.5em}%
\texttt{<think>} First, I need to identify the river that Toongabbie Creek serves as the mouth of. \texttt{</think>}\\
\hspace*{1.5em}%
\texttt{<search>} river for which Toongabbie Creek serves as the mouth \texttt{</search>}\\
\hspace*{1.5em}%
\texttt{<information>}
Doc 1: ... The creek flows generally south-west, then south by east, then south by west, then east before reaching its \textcolor{blue}{\textbf{confluence with the Darling Mills Creek to form the Parramatta River}}, in the suburb of ...
\texttt{</information>}\\
\hspace*{1.5em}%
\texttt{<think>} The information provided indicates that Toongabbie Creek flows into the Parramatta River. \texttt{</think>}\\
\hspace*{1.5em}%
\texttt{<subAnswer>} \textcolor{green!60!black}{\textbf{\#1 = Parramatta River}} \texttt{</subAnswer>}\\
\texttt{</subPlan>}}}

\colorbox{cyan!15}{\parbox{\dimexpr\linewidth-2\fboxsep}{%
\texttt{<subPlan>}\\
\hspace*{1.5em}%
\texttt{<think>} Next, I need to determine what the Parramatta River flows into. \texttt{</think>}\\
\hspace*{1.5em}%
\texttt{<search>} Parramatta River flows into \texttt{</search>}\\
\hspace*{1.5em}%
\texttt{<information>}
Doc 1: Parramatta River The Parramatta River is an intermediate tide dominated, drowned valley estuary located in Sydney, New South Wales, Australia. \textcolor{blue}{\textbf{The Parramatta River is the main tributary of Sydney Harbour}}, a branch of Port Jackson ... the river flows in an easterly direction to a line between Yurulbin, Birchgrove and Manns Point, Greenwich. Here it \textcolor{blue}{\textbf{flows into Port Jackson}} ...
\texttt{</information>}\\
\hspace*{1.5em}%
\texttt{<think>} The information provided indicates that the Parramatta River flows into Sydney Harbour. \texttt{</think>}\\
\hspace*{1.5em}%
\texttt{<subAnswer>} \textcolor{green!60!black}{\textbf{\#2 = Sydney Harbour}} \texttt{</subAnswer>}\\
\texttt{</subPlan>}}}

\colorbox{gray!15}{\parbox{\dimexpr\linewidth-2\fboxsep}{\texttt{<think>} To answer the original question, I need to combine the answers from the sub-questions. With this information, I can determine into what river Toongabbie Creek flows. With this, I can provide the final answer to the question. \texttt{</think>}}}

\colorbox{yellow!25}{\parbox{\dimexpr\linewidth-2\fboxsep}{\texttt{<answer>} \textcolor{green!60!black}{\textbf{Sydney Harbour}} \texttt{</answer>}}}\\

\hdashline

\textbf{Search-R1:} \\
\colorbox{gray!15}{\parbox{\dimexpr\linewidth-2\fboxsep}{\texttt{<think>} To find out what river the Toongabbie Creek serves as the mouth to, I will conduct a search for this information. \texttt{</think>}}}\\
\colorbox{green!10}{\parbox{\dimexpr\linewidth-2\fboxsep}{\texttt{<search>} Into what does the river for which Toongabbie Creek serves as the mouth flow \texttt{</search>}}}\\
\colorbox{blue!10}{\parbox{\dimexpr\linewidth-2\fboxsep}{\texttt{<information>} Doc 1: Toongabbie Creek, an urban watercourse that is part of the Parramatta River catchment, is located in Greater Western Sydney, New South Wales, Australia ... The creek flows generally south-west, then south by east, then south by west, then east before reaching its \textcolor{blue}{\textbf{confluence with the Darling Mills Creek to form the Parramatta River}}, in the suburb of ...
\texttt{</information>}}}
\colorbox{gray!15}{\parbox{\dimexpr\linewidth-2\fboxsep}{\texttt{<think>} Based on the information retrieved: \texttt{</think>}}}
\colorbox{green!10}{\parbox{\dimexpr\linewidth-2\fboxsep}{\texttt{<search>} what river does the mouth of Toongabbie Creek flow into \texttt{</search>}}}
\colorbox{blue!10}{\parbox{\dimexpr\linewidth-2\fboxsep}
{\texttt{<information>} 
...
\texttt{</information>}}}
\colorbox{gray!15}{\parbox{\dimexpr\linewidth-2\fboxsep}{\texttt{<think>} From the provided documents: \texttt{</think>}}}
\colorbox{yellow!25}{\parbox{\dimexpr\linewidth-2\fboxsep}{\texttt{<answer>} \textcolor{red}{\textbf{Parramatta River}} \texttt{</answer>}}}\\

% \midrule

% \textbf{Analysis:} 
% GlobalRAG explicitly decompositions and guides the model to orderly execute inference steps according to the preset plan. In contrast, search-R1 does not perform problem decomposition. The Search uses the original problem instead of sub-problems, resulting in inaccurate search and misjudging the answers of sub-problems as the final answers.
\bottomrule
\end{tabular}
}
\caption{Case Study 1. The predictions are colored \textcolor{red}{\textbf{Red}} if the model answers incorrectly and \textcolor{green!60!black}{\textbf{Green}} if it answers correctly. The core evidences are colored \textcolor{blue}{\textbf{Blue}}.}
\label{tab:case_study1}
\end{table*}

\begin{table*}[t]
\centering
\resizebox{0.95\textwidth}{!}{
\begin{tabular}{p{0.95\textwidth}}
\toprule
\toprule
\textbf{Case 2:} Who was the first African American mayor of the most populous city in the United States?  \\
\textbf{Ground Truth:} \textbf{David Dinkins} \\
\midrule

\textbf{\ourMethod:} \\
\colorbox{gray!15}{\parbox{\dimexpr\linewidth-2\fboxsep}
{\texttt{<think>} To answer this question, I need to break it down into sub-questions. First, I need to identify the most populous city in the United States. Then, I need to find the first African American mayor of that city. I can reason through the sub-questions to get to the final answer. \texttt{</think>}}}

\colorbox{orange!20}{\parbox{\dimexpr\linewidth-2\fboxsep}{\texttt{<plan>} \{ "Q1": ["What is the most populous city in the United States?", "\#1"], "Q2": ["Who was the first African American mayor of \#1?", "\#2"]  \} \texttt{</plan>}}}

\colorbox{cyan!15}{\parbox{\dimexpr\linewidth-2\fboxsep}{
\texttt{<subPlan>}\\
\hspace*{1.5em}%
\texttt{<think>} First, I need to determine the most populous city in the United States. \texttt{</think>}\\
\hspace*{1.5em}%
\texttt{<search>} most populous city in the United States \texttt{</search>}\\
\hspace*{1.5em}%
\texttt{<information>}
Doc 1: New York City The City of New York, often called New York City (NYC) or simply New York (NY), is \textcolor{blue}{\textbf{the most populous city in the United States}} ...
\texttt{</information>}\\
\hspace*{1.5em}%
\texttt{<think>} The information provided indicates that the most populous city in the United States is New York City. \texttt{</think>}\\
\hspace*{1.5em}%
\texttt{<subAnswer>} \textcolor{green!60!black}{\textbf{\#1 = New York City}} \texttt{</subAnswer>}\\
\texttt{</subPlan>}}}

\colorbox{cyan!15}{\parbox{\dimexpr\linewidth-2\fboxsep}{%
\texttt{<subPlan>}\\
\hspace*{1.5em}%
\texttt{<think>} Next, I need to find the first African American mayor of New York City. \texttt{</think>}\\
\hspace*{1.5em}%
\texttt{<search>} first African American mayor of New York City \texttt{</search>}\\
\hspace*{1.5em}%
\texttt{<information>}
... David Dinkins David Norman Dinkins (born July 10, 1927) is an American politician, lawyer, and author who served as the 106th Mayor of New York City, from 1990 to 1993. He was \textcolor{blue}{\textbf{the first and, to date, the only African American to hold that office}} ...
\texttt{</information>}\\
\hspace*{1.5em}%
\texttt{<think>} The information provided indicates that the first African American mayor of New York City was David Dinkins. \texttt{</think>}\\
\hspace*{1.5em}%
\texttt{<subAnswer>} \textcolor{green!60!black}{\textbf{\#2 = David Dinkins }} \texttt{</subAnswer>}\\
\texttt{</subPlan>}}}

\colorbox{gray!15}{\parbox{\dimexpr\linewidth-2\fboxsep}{\texttt{<think>} Now that I have the necessary subAnswers, I can determine the first African American mayor of the most populous city in the United States. With this information, I will be able to provide the final answer to the question. \texttt{</think>}}}

\colorbox{yellow!25}{\parbox{\dimexpr\linewidth-2\fboxsep}{\texttt{<answer>} \textcolor{green!60!black}{\textbf{David Dinkins}} \texttt{</answer>}}}\\

\hdashline

\textbf{Search-R1:} \\
\colorbox{gray!15}{\parbox{\dimexpr\linewidth-2\fboxsep}{\texttt{<think>} To find out who the first African American mayor of the most populous city in the United States was, I will conduct some research. \texttt{</think>}}}\\
\colorbox{green!10}{\parbox{\dimexpr\linewidth-2\fboxsep}{\texttt{<search>} first African American mayor of the most populous city in the United States \texttt{</search>}}}\\
\colorbox{blue!10}{\parbox{\dimexpr\linewidth-2\fboxsep}{\texttt{<information>} Doc 1: ... she became the \textcolor{blue}{\textbf{first female African American Mayor of a major metropolitan city}} [Compton, California] ...
Doc 2: Robert Clayton Henry ... was Mayor of Springfield, Ohio from 1966 to 1968. He was the \textcolor{blue}{\textbf{first African-American mayor of an American city of any size}} ...
Doc 3: Theodore M. Berry ... was the \textcolor{blue}{\textbf{first African-American mayor of Cincinnati, Ohio}} ...
\texttt{</information>}}}
\colorbox{gray!15}{\parbox{\dimexpr\linewidth-2\fboxsep}{\texttt{<think>} Based on the research: \texttt{</think>}}}
\colorbox{green!10}{\parbox{\dimexpr\linewidth-2\fboxsep}{\texttt{<search>} who was the first African American mayor of the most populous city in the United States \texttt{</search>}}}
\colorbox{blue!10}{\parbox{\dimexpr\linewidth-2\fboxsep}{\texttt{<information>} ...
\texttt{</information>}}}
\colorbox{gray!15}{\parbox{\dimexpr\linewidth-2\fboxsep}{\texttt{<think>} From the data, I realize Robert C. Henry could be wrongly aligned, Ted Berry indeed form Document important positions. \texttt{</think>}}}
\colorbox{yellow!25}{\parbox{\dimexpr\linewidth-2\fboxsep}{\texttt{<answer>} \textcolor{red}{\textbf{Ted Berry}} \texttt{</answer>}}}\\

% \midrule

% \textbf{Analysis:} 
% GlobalRAG explicitly decompositions and guides the model to orderly execute inference steps according to the preset plan. In contrast, search-R1 does not perform problem decomposition. The Search uses the original problem instead of sub-problems, resulting in generalization of the search results and deviation of the answers, thereby misjudging the answers of the sub-problems as the final answers.
\bottomrule
\end{tabular}
}
\caption{Case Study 2. The predictions are colored \textcolor{red}{\textbf{Red}} if the model answers incorrectly and \textcolor{green!60!black}{\textbf{Green}} if it answers correctly. The core evidences are colored \textcolor{blue}{\textbf{Blue}}.}
\label{tab:case_study2}
\end{table*}

\begin{table*}[t]
\centering
\resizebox{0.95\textwidth}{!}{
\begin{tabular}{p{0.95\textwidth}}
\toprule
\toprule
\textbf{Case 3:} Who is the spouse of the producer of \textit{Dolores Claiborne}? \\
\textbf{Ground Truth:} \textbf{Helen Mirren} \\
\midrule

\textbf{\ourMethod:} \\
\colorbox{gray!15}{\parbox{\dimexpr\linewidth-2\fboxsep}
{\texttt{<think>} To answer this question, I need to break it down into sub-questions. First, I need to identify the producer of the film Dolores Claiborne. Then, I need to find who that producer is married to. I can proceed step-by-step to find the answer. \texttt{</think>}}}

\colorbox{orange!20}{\parbox{\dimexpr\linewidth-2\fboxsep}{\texttt{<plan>} \{ "Q1": ["Who is the producer of Dolores Claiborne?", "\#1"], "Q2": ["Who is the spouse of \#1?", "\#2"] \} \texttt{</plan>}}}

\colorbox{cyan!15}{\parbox{\dimexpr\linewidth-2\fboxsep}{
\texttt{<subPlan>}\\
\hspace*{1.5em}%
\texttt{<think>} First, I need to identify the producer of the film Dolores Claiborne. \texttt{</think>}\\
\hspace*{1.5em}%
\texttt{<search>} producer of Dolores Claiborne film \texttt{</search>}\\
\hspace*{1.5em}%
\texttt{<information>}
Doc 1: \textcolor{blue}{\textbf{Dolores Claiborne is a 1995 American
psychological thriller drama film directed by Taylor Hackford}} and starring Kathy Bates, Jennifer Jason Leigh, and David Strathairn ...
\texttt{</information>}\\
\hspace*{1.5em}%
\texttt{<think>} The information clearly states that the film was directed by Taylor Hackford. \texttt{</think>}\\
\hspace*{1.5em}%
\texttt{<subAnswer>} \textcolor{green!60!black}{\textbf{\#1 = Taylor Hackford}} \texttt{</subAnswer>}\\
\texttt{</subPlan>}}}

\colorbox{cyan!15}{\parbox{\dimexpr\linewidth-2\fboxsep}{%
\texttt{<subPlan>}\\
\hspace*{1.5em}%
\texttt{<think>} Next, I need to find who Taylor Hackford is married to. \texttt{</think>}\\
\hspace*{1.5em}%
\texttt{<search>} Taylor Hackford spouse \texttt{</search>}\\
\hspace*{1.5em}%
\texttt{<information>}
Doc 1: ... Hackford has been married three times ...
Doc 2: ... \textcolor{blue}{\textbf{Hackford has been married to Academy Award-winning actress Helen Mirren since 1997}} ...
\texttt{</information>}\\
\hspace*{1.5em}%
\texttt{<think>} The information indicates that Taylor Hackford is married to Helen Mirren. \texttt{</think>}\\
\hspace*{1.5em}%
\texttt{<subAnswer>} \textcolor{green!60!black}{\textbf{\#2 = Helen Mirren}} \texttt{</subAnswer>}\\
\texttt{</subPlan>}}}

\colorbox{gray!15}{\parbox{\dimexpr\linewidth-2\fboxsep}{\texttt{<think>} Now that I have the necessary sub Answers, I can determine the spouse of the producer of Dolores Claiborne. \texttt{</think>}}}

\colorbox{yellow!25}{\parbox{\dimexpr\linewidth-2\fboxsep}{\texttt{<answer>} \textcolor{green!60!black}{\textbf{Helen Mirren}} \texttt{</answer>}}}\\

\hdashline

\textbf{Search-R1:} \\
\colorbox{gray!15}{\parbox{\dimexpr\linewidth-2\fboxsep}{\texttt{<think>} To find the spouse of the producer of "Dolores Claiborne", I need to gather details on the producer of the film. I will conduct a search to locate that information. \texttt{</think>}}}\\
\colorbox{green!10}{\parbox{\dimexpr\linewidth-2\fboxsep}{\texttt{<search>} producer of Dolores Claiborne \texttt{</search>}}}\\
\colorbox{blue!10}{\parbox{\dimexpr\linewidth-2\fboxsep}{\texttt{<information>} Doc 1: Dolores Claiborne is a 1995 American psychological thriller drama film directed by Taylor Hackford ... \textcolor{blue}{\textbf{The screenplay for ``Dolores Claiborne'' was adapted by Tony Gilroy}} ...
\texttt{</information>}}}
\colorbox{gray!15}{\parbox{\dimexpr\linewidth-2\fboxsep}{\texttt{<think>} From the findings, I confirm that the producer of "Dolores Claiborne" is Tony Gilroy. \texttt{</think>}}}
\colorbox{green!10}{\parbox{\dimexpr\linewidth-2\fboxsep}{\texttt{<search>} who is the spouse of Tony Gilroy \texttt{</search>}}}
\colorbox{blue!10}{\parbox{\dimexpr\linewidth-2\fboxsep}{\texttt{<information>} Doc 1: Tony Gilroy ... He has two children, Sam and Kathryn, and is married to \textcolor{blue}{\textbf{Susan Gilroy}} ...
\texttt{</information>}}}
\colorbox{gray!15}{\parbox{\dimexpr\linewidth-2\fboxsep}{\texttt{<think>} From this information: \texttt{</think>}}}
\colorbox{yellow!25}{\parbox{\dimexpr\linewidth-2\fboxsep}{\texttt{<answer>} \textcolor{red}{\textbf{Susan Gilroy}} \texttt{</answer>}}}\\

% \midrule

% \textbf{Analysis:} 
% GlobalRAG, through explicit decomposition, enables the model to systematically execute problem-solving steps according to a predefined plan. Moreover, for first sub-problem, the model's response demonstrates explicit reasoning through the <think> mechanism, leading to the correct answer. In contrast, while Search-R1 exhibits a certain degree of planning capability in its implicit reasoning, the absence of explicit plan guidance results in deviations in addressing first sub-problem, ultimately leading to an incorrect final answer.

\bottomrule
\end{tabular}
}
\caption{Case Study 3. The predictions are colored \textcolor{red}{\textbf{Red}} if the model answers incorrectly and \textcolor{green!60!black}{\textbf{Green}} if it answers correctly. The core evidences are colored \textcolor{blue}{\textbf{Blue}}.}
\label{tab:case_study3}
\end{table*}

\begin{table*}[t]
\centering
\resizebox{0.85\textwidth}{!}{
\begin{tabular}{p{0.95\textwidth}}
\toprule
\toprule
\textbf{Case 4:} When was the Western Germanic language spoken from which the small settlement situated on the river Leda opposite Leer derives its name?  \\
\textbf{Ground Truth:} \textbf{between the 8th and 16th centuries } \\
\midrule

\textbf{\ourMethod:} \\
\colorbox{gray!15}{\parbox{\dimexpr\linewidth-2\fboxsep}
{\texttt{<think>} To answer this question, I need to break it down into sub-questions. First, ... Then, ... Finally, ... I will follow the steps to generate sub-questions and answer them sequentially.  \texttt{</think>}}}

\colorbox{orange!20}{\parbox{\dimexpr\linewidth-2\fboxsep}{\texttt{<plan>} \{ "Q1": ["Which small settlement derives its name from the river Leda?", "\#1"], "Q2": ["From which Western Germanic language does \#1 derive its name?", "\#2"], "Q3": ["When was the Western Germanic language spoken from which \#1 derives its name?", "\#3"]   \} \texttt{</plan>}}}

\colorbox{cyan!15}{\parbox{\dimexpr\linewidth-2\fboxsep}{
\texttt{<subPlan>}\\
\hspace*{1.5em}%
\texttt{<think>} First, I need to identify the small settlement that derives its name from the river Leda. \texttt{</think>}\\
\hspace*{1.5em}%
\texttt{<search>} small settlement named after river Leda opposite Leer \texttt{</search>}\\
\hspace*{1.5em}%
\texttt{<information>}
Doc 3: ... On the southern bank of the Leda, in the "Overledingen Land" (Overledingen="country over the Leda"), opposite Leer, lies the \textcolor{blue}{\textbf{small settlement of Kloster Muhde}} ("Muhde" from the \textcolor{blue}{\textbf{Old Frisian "mutha"}} meaning "(river) mouth") ...
\texttt{</information>}\\
\hspace*{1.5em}%
\texttt{<think>} The information provided indicates that the small settlement situated on the river Leda opposite Leer is Kloster Muhde. \texttt{</think>}\\
\hspace*{1.5em}%
\texttt{<subAnswer>} \textcolor{green!60!black}{\textbf{\#1 = Kloster Muhde}} \texttt{</subAnswer>}\\
\texttt{</subPlan>}}}

\colorbox{cyan!15}{\parbox{\dimexpr\linewidth-2\fboxsep}{
\texttt{<subPlan>}\\
\hspace*{1.5em}%
\texttt{<think>} Next, I need to determine the Western Germanic language from which the name of Kloster Muhde derives. \texttt{</think>}\\
\hspace*{1.5em}%
\texttt{<search>} origin of name of Kloster Muhde in Lower Saxony \texttt{</search>}\\
\hspace*{1.5em}%
\texttt{<information>}
...
\texttt{</information>}\\
\hspace*{1.5em}%
\texttt{<think>} ... Western Germanic language from which the name derives is Old Frisian. \texttt{</think>}\\
\hspace*{1.5em}%
\texttt{<subAnswer>} \textcolor{green!60!black}{\textbf{\#2 = Old Frisian}} \texttt{</subAnswer>}\\
\texttt{</subPlan>}}}

\colorbox{cyan!15}{\parbox{\dimexpr\linewidth-2\fboxsep}{%
\texttt{<subPlan>}\\
\hspace*{1.5em}%
\texttt{<think>} Finally, I need to find out when the Western Germanic language from which Kloster Muhde derives its name was spoken. \texttt{</think>}\\
\hspace*{1.5em}%
\texttt{<search>} when Old Frisian language was spoken \texttt{</search>}\\
\hspace*{1.5em}%
\texttt{<information>}
Doc 1: Old Frisian Old Frisian is a West Germanic language \textcolor{blue}{\textbf{spoken between the 8th and 16th centuries}} in the area ...
\texttt{</information>}\\
\hspace*{1.5em}%
\texttt{<think>} The information indicates that the Old Frisian language, from which the name of Kloster Muhde derives, was spoken between the 8th and 16th centuries. \texttt{</think>}\\
\hspace*{1.5em}%
\texttt{<subAnswer>} \textcolor{green!60!black}{\textbf{\#3 = between the 8th and 16th centuries }} \texttt{</subAnswer>}\\
\texttt{</subPlan>}}}

\colorbox{gray!15}{\parbox{\dimexpr\linewidth-2\fboxsep}{\texttt{<think>} ... With this information, I can provide the final answer to the question. \texttt{</think>}}}

\colorbox{yellow!25}{\parbox{\dimexpr\linewidth-2\fboxsep}{\texttt{<answer>} \textcolor{green!60!black}{\textbf{between the 8th and 16th centuries}} \texttt{</answer>}}}\\

\hdashline

\textbf{Search-R1:} \\
\colorbox{gray!15}{\parbox{\dimexpr\linewidth-2\fboxsep}{\texttt{<think>} To find out ... , I need to first identify the settlement's name and the historical context surrounding its name. \texttt{</think>}}}\\
\colorbox{green!10}{\parbox{\dimexpr\linewidth-2\fboxsep}{\texttt{<search>} name of settlement on river Leda opposite Leer Western Germanic language \texttt{</search>}}}\\
\colorbox{blue!10}{\parbox{\dimexpr\linewidth-2\fboxsep}{\texttt{<information>} ... (Overledingen="country over the Leda"), opposite Leer, lies the \textcolor{blue}{\textbf{small settlement of Kloster Muhde}} ("Muhde" from the \textcolor{blue}{\textbf{Old Frisian "mutha"}} meaning ...
\texttt{</information>}}}
\colorbox{gray!15}{\parbox{\dimexpr\linewidth-2\fboxsep}{\texttt{<think>} From the information retrieved: The small settlement on the river Leda opposite Leer is named Kloster Muhde. \texttt{</think>}}}
\colorbox{green!10}{\parbox{\dimexpr\linewidth-2\fboxsep}{\texttt{<search>} when was Western Germanic language spoken \texttt{</search>}}}
\colorbox{blue!10}{\parbox{\dimexpr\linewidth-2\fboxsep}{\texttt{<information>}  
... Germanic dialects were closely enough related to have been \textcolor{blue}{\textbf{mutually intelligible up to the 7th century}}... successively. The High German consonant shift that occurred mostly during the \textcolor{blue}{\textbf{7th century AD}} in what is now southern Germany, Austria, and Switzerland can be considered \textcolor{blue}{\textbf{the end of the linguistic unity among the West Germanic dialects}} ...
\texttt{</information>}}}
\colorbox{gray!15}{\parbox{\dimexpr\linewidth-2\fboxsep}{\texttt{<think>} Based on these documents: \texttt{</think>}}}
\colorbox{yellow!25}{\parbox{\dimexpr\linewidth-2\fboxsep}{\texttt{<answer>} \textcolor{red}{\textbf{7th century}} \texttt{</answer>}}}\\

% \midrule

% \textbf{Analysis:} 
% GlobalRAG's multi-step explicit decomposition and targeted retrieval-reasoning process enable it to precisely locate the specific West Germanic language and its usage time in the problem, thereby providing the correct answer. However, search-R1 fails to distinguish between specific languages and the entire language family, resulting in generalized Search results and biased answers.
\bottomrule
\end{tabular}
}
\caption{Case Study 4. The predictions are colored \textcolor{red}{\textbf{Red}} if the model answers incorrectly and \textcolor{green!60!black}{\textbf{Green}} if it answers correctly. The core evidences are colored \textcolor{blue}{\textbf{Blue}}.}
\label{tab:case_study4}
\end{table*}

\end{document}